\documentclass[twocolumn]{article}

\usepackage{titletoc}  %
\usepackage{placeins}  %

\usepackage{maira}

\usepackage{pdfpages}

\hypersetup{colorlinks=true, linkcolor=blue, citecolor=blue, urlcolor=blue}

\usepackage[capitalize]{cleveref}

\title{MAIRA-2: Grounded Radiology Report Generation}

\author[1]{Shruthi Bannur$^\star$}
\author[1]{Kenza Bouzid$^\star$}
\author[1]{Daniel C. Castro}
\author[1]{Anton Schwaighofer}
\author[1]{Anja Thieme}
\author[1]{Sam Bond-Taylor}
\author[1]{Maximilian Ilse}
\author[1]{Fernando Pérez-García}
\author[1]{Valentina Salvatelli}
\author[1]{Harshita Sharma}
\author[1]{Felix Meissen}
\author[2]{Mercy Ranjit}
\author[2]{Shaury Srivastav}
\author[3]{Julia Gong}
\author[4]{Noel C. F. Codella}
\author[1]{Fabian Falck}
\author[1]{Ozan Oktay}
\author[4]{Matthew P. Lungren}
\author[1,5]{Maria Teodora Wetscherek}
\author[1]{Javier Alvarez-Valle$^\circ$}
\author[1]{Stephanie L. Hyland$^\circ$}
\affil[1]{Microsoft Research Health Futures}
\affil[2]{Microsoft Research India}
\affil[3]{Microsoft Azure AI}
\affil[4]{Microsoft Health and Life Sciences\authorcr}
\affil[5]{Department of Radiology, Addenbrooke’s Hospital, Cambridge University Hospitals}

\newcommand\blfootnote[1]{%
  \begingroup
  \renewcommand\thefootnote{}\footnote{#1}%
  \addtocounter{footnote}{-1}%
  \endgroup
}

\date{}  %

\begin{document}
\begin{acronym}
    \acro{VQA}{visual question answering}
    \acro{LLM}{large language model}
    \acro{SOTA}{state of the art}
    \acro{NPV}{negative predictive value}
    \acro{PPV}{positive predictive value}
    \acro{RL}{reinforcement learning}
    \acro{PA}{posteroanterior}
    \acro{AP}{anteroposterior}
    \acro{CXR}{chest X-ray}
    \acro{mIoU}{mean intersection over union}
    \acro{MLP}{multilayer perceptron}
    \acro{AI}{artificial intelligence}
    \acro{ROI}{region of interest}
    \acro{MPG}{medical phrase grounding}
    \acro{NLG}{natural language generation}
\end{acronym}

\newcommand{\raddinoorig}{R\textsc{ad}-DINO\xspace}
\newcommand{\raddino}{R\textsc{ad}-DINO-MAIRA-2\xspace}
\newcommand{\mairatwo}{MAIRA-2\xspace}
\newcommand{\mairatwothirteenb}{MAIRA-2~13B\xspace}
\newcommand{\chexagent}{CheXagent\xspace}
\newcommand{\maira}{MAIRA-1\xspace}
\newcommand{\modelname}[1]{\texttt{#1}}
\newcommand{\llama}{Llama3\xspace}
\newcommand{\llamaseventy}{Llama3-70B\xspace}
\newcommand{\medgemini}{Med-Gemini\xspace}

\newcommand{\eightk}{\texttt{GR-Bench}\xspace}  %
\newcommand{\hundredk}{\texttt{GR-1}\xspace} %
\newcommand{\padchest}{PadChest\xspace}
\newcommand{\padchesttwo}{PadChest-GR\xspace}
\newcommand{\privatedata}{USMix\xspace} %
\newcommand{\mscxr}{MS-CXR\xspace}
\newcommand{\mimiccxr}{MIMIC-CXR\xspace}
\newcommand{\openi}{IU-Xray\xspace} %

\newcommand{\evmetric}{\texttt{RadFact}\xspace} %
\newcommand{\evmetricgpt}{\texttt{\evmetric-GPT4}\xspace} %
\newcommand{\evmetricllama}{\texttt{\evmetric-Llama3}\xspace} %
\newcommand{\radfactcode}{\url{https://github.com/microsoft/RadFact}}
\newcommand{\lateralmentions}{\textit{\%Lateral mentions}\xspace}
\newcommand{\compmentions}{\textit{\%Comparison mentions}\xspace}

\newcommand{\macrofourteen}{Macro F$_1$-14\xspace}
\newcommand{\microfourteen}{Micro F$_1$-14\xspace}
\newcommand{\macrofive}{Macro F$_1$-5\xspace}
\newcommand{\rger}{RG\textsubscript{ER}\xspace}

\newcommand{\tabspace}{\vspace{0.15cm}}

\newcommand{\commondescription}{We report median and 95\% confidence intervals based on 500 bootstrap samples. `$\downarrow$' indicates that lower is better. CheXpert F$_1$ metrics are computed based on CheXbert labeller outputs. \evmetric uses \evmetricllama.\xspace}
\newcommand{\plotsdescription}{We report median and 95\% confidence intervals based on 500 bootstrap samples. `$\downarrow$' indicates that lower is better.\xspace}
\newcommand{\maintablesdescription}{We report median and 95\% confidence intervals based on 500 bootstrap samples. \textbf{Bold} indicates best performance for that metric, or overlapping CIs with best. `$\downarrow$' indicates that lower is better. CheXpert F$_1$ metrics are computed based on CheXbert labeller outputs. \evmetric uses \evmetricllama.\xspace}

\newcommand{\ci}[1]{{\scriptsize\textcolor{gray}{#1}}}
\newcommand{\mr}[1]{{\multirow{2}*{#1}}}

\newcommand{\reportsection}[1]{\textit{#1}}
\newcommand{\findings}{\reportsection{Findings}\xspace}
\newcommand{\impression}{\reportsection{Impression}\xspace}
\newcommand{\indication}{\reportsection{Indication}\xspace}
\newcommand{\technique}{\reportsection{Technique}\xspace}
\newcommand{\comparison}{\reportsection{Comparison}\xspace}
\newcommand{\frontal}{\reportsection{Frontal}\xspace}
\newcommand{\lateral}{\reportsection{Lateral}\xspace}
\newcommand{\prior}{\reportsection{Prior}\xspace}

\newcommand{\cmark}{\ding{51}}%
\newcommand{\xmark}{\ding{55}}%

\newcommand{\findgen}{\texttt{FindGen}\xspace}
\newcommand{\groundrep}{\texttt{GroundRep}\xspace}
\newcommand{\phraseground}{\texttt{PhraseGround}\xspace}
\newcommand{\ctext}[2]{{\sethlcolor{#1}\hl{#2}}}
\newcommand{\reporterror}[1]{\ctext{red!20}{#1}}
\newcommand{\reportomission}[1]{\ctext{lightgray!50}{[\textit{#1}]}}
\newcommand{\reporthighlight}[1]{\ctext{ProcessBlue!20}{#1}}

\NewDocumentCommand{\token}{mo}{$\langle$\texttt{#1}\IfValueT{#2}{$_{#2}$}$\rangle$}
\newcommand{\placeholder}[1]{\texttt{\{#1\}}}

\newcommand{\drop}{No}
\newcommand{\abTrain}{Train:}
\newcommand{\abInf}{Infer:}

\newcommand{\lateralshort}{\reportsection{Lat}}
\newcommand{\priorshort}{\reportsection{Prior}}
\newcommand{\techniqueshort}{\reportsection{Tech}}
\newcommand{\comparsionshort}{\reportsection{Comp}}

\newcommand{\abNoSections}{\drop~\techniqueshort~\drop~\comparsionshort\xspace}
\newcommand{\abDropLat}{\drop~\lateralshort}
\newcommand{\abDropTech}{\drop~\techniqueshort}
\newcommand{\abDropLatTech}{\drop~\lateralshort~\drop~\techniqueshort}
\newcommand{\abDropPrior}{\drop~\priorshort}
\newcommand{\abDropComp}{\drop~\comparsionshort}
\newcommand{\abDropPriorComp}{\drop~\priorshort~\drop~\comparsionshort}
\newcommand{\abFrontal}{Frontal only\xspace}
\newcommand{\abFrontalAllSec}{\drop~\priorshort~\drop~\lateralshort\xspace}
\newcommand{\abFrontalFG}{\abFrontal\drop~\findgen\xspace}
\newcommand{\abDrophundredk}{\drop\hundredk\xspace}
\newcommand{\abDropMSCXR}{\drop\mscxr\xspace}
\newcommand{\abFGOnly}{\drop\groundrep\xspace}
\newcommand{\abGROnly}{\drop\findgen\xspace}

\clearpage

\maketitle

\vspace*{-15mm}
\begin{abstract}
\blfootnote{$^\star$ Joint first authors. $^\circ$ Joint senior authors.}
Radiology reporting is a complex task requiring detailed medical image understanding and precise language generation, for which generative multimodal models offer a promising solution.
However, to impact clinical practice, models must achieve a high level of both verifiable performance and utility.
We augment the utility of automated report generation by incorporating localisation of individual findings on the image -- a task we call grounded report generation -- and enhance performance by incorporating realistic reporting context as inputs. We design a novel evaluation framework (\evmetric) leveraging the logical inference capabilities of \acp{LLM} to quantify report correctness and completeness at the level of individual sentences, while supporting the new task of grounded reporting.
We develop \mairatwo, 
a large radiology-specific multimodal model designed to generate chest X-ray reports with and without grounding.
\mairatwo achieves state of the art on existing report generation benchmarks and establishes the novel task of grounded report generation.
\end{abstract}

\section*{Introduction}
\label{sec:intro}
Medical imaging is central to the safe and effective delivery of modern medicine \citep{UKHealthSecurityAgency2022}.
Nonetheless, the increasing demand for imaging services is surpassing the capacity of radiologists to maintain a high quality standard in image reporting~\citep{fischetti2022evolving, Kalidindi2023}. 
The worsening shortage of radiology professionals is leading to increasing levels of stress and burnout among staff~\citep{RCR2022} and causing delays and disparities in the delivery of critical care~\citep{rimmer2017radiologist}.

Systems leveraging \ac{AI} could support radiologists by generating a first draft of the report, potentially enhancing operational efficiency, reducing radiologist workloads, and improving the quality and standardisation of patient care~\citep{huang2023generative, liu2019clinically, yildirim2024multimodal, yu2023evaluating}. Consequently, the generation of narrative-style reports from radiology images has become subject to increasing research interest as a challenging task for multimodal medical \ac{AI}~\citep{zhou2024generalist,yang2024advancing,tu2024medpalmm,chen2024chexagent,wang2023metransformer,li2023dynamic}.
However, for an \ac{AI}-generated draft report to be useful, it must: (i)~replicate or exceed what the radiologist would have written, without hallucinations or omissions, and (ii)~be easy to verify, shortcomings which remain unsolved to date.

Here, we propose modifications to the automated report generation task to bring \ac{AI} research closer to clinical utility. We advocate for
(i)~incorporating additional \emph{context}, bringing the inputs of the model closer to the information used by the radiologist \citep{nguyen2023pragmatic,bannur2023biovilt}, and (ii)~extending the task to require the spatial \emph{grounding} of each described finding in the image through image-level annotations, such as bounding boxes. We hypothesise that additional context will improve report quality, while grounding will support verification~\citep{bernstein2023can}, image comprehension~\citep{yildirim2024multimodal}, and potentially enable new use-cases as a key capability of `generalist medical \ac{AI}'~\citep{Moor2023}.

We propose \mairatwo, a first-of-its-kind model for the task of grounded radiology report generation. \mairatwo is a \ac{CXR}-specialised multimodal model capable of generating both grounded and non-grounded reports while integrating more comprehensive inputs -- namely the lateral view, prior frontal image, prior report, 
\reportsection{Indication}, \reportsection{Technique}, and \reportsection{Comparison} sections.

To evaluate the quality of draft reports with and without grounding, we propose a novel evaluation framework named \evmetric. 
Inspired by factuality-based approaches ~\citep{min2023factscore,schumacher2023extrinsically}, and building on the observation that GPT-4 exhibits strong logical reasoning capabilities in radiology~\citep{liu2023exploring}, \evmetric leverages \acp{LLM} to ascertain the factuality of \emph{each} sentence in a generated report, given sentences from the reference ground truth. This provides for an interpretable sentence-level view of errors, while %
also enabling evaluation of grounding annotations between matched sentences.

To support further research on grounded radiology report generation, we release the \mairatwo model, an open-source implementation of \evmetric at \radfactcode, and the annotation protocol for creating grounded reports in \cref{app:annotation_protocol}.

\begin{figure*}[p]
    \centering
    \includegraphics[width=\linewidth]{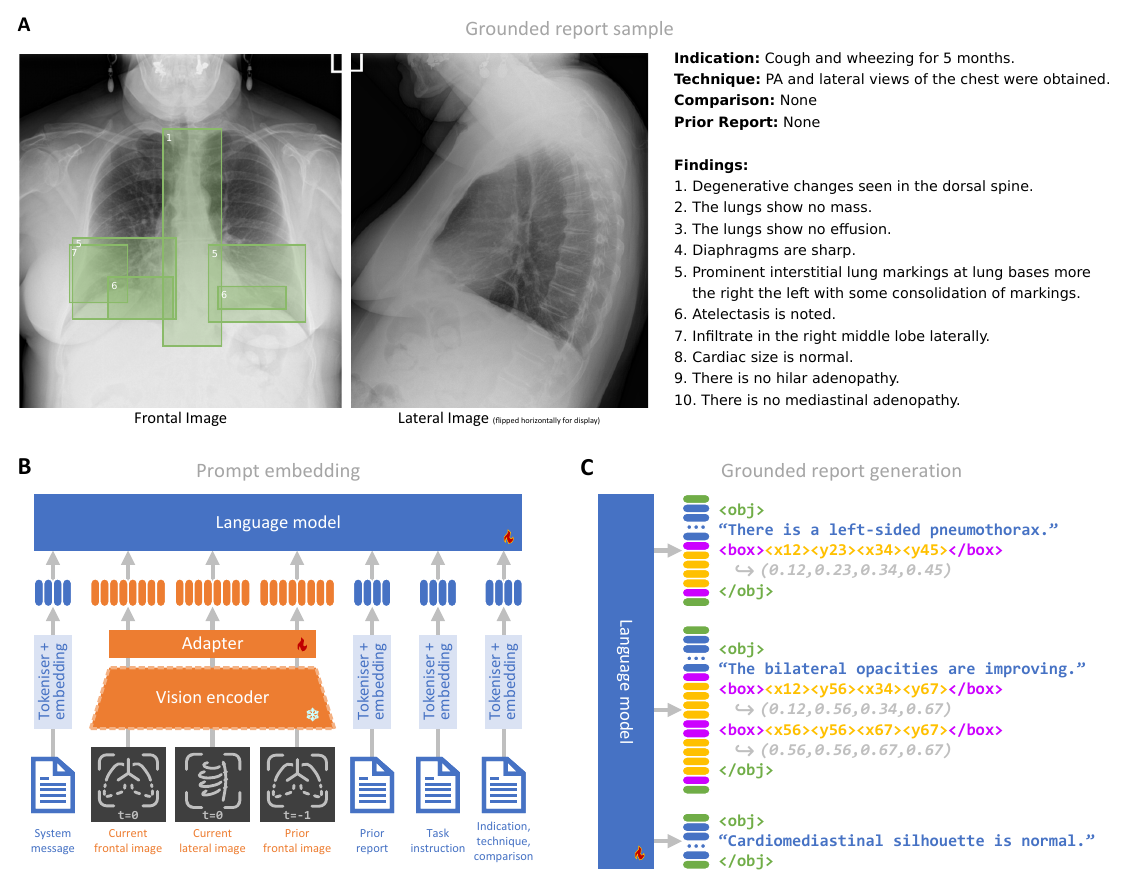}
    \caption{
    \textbf{Grounded report generation with \mairatwo.}
    (Panel~A)~An illustrative example of the grounded reporting task.
    A grounded report is a list of sentences potentially linked to spatial annotations (bounding boxes, in this work). Normal anatomy or non-findings, as well as non-localisable observations, do not require spatial annotations.
    To generate a grounded report, the model can be presented with all or some of the following: the current study's frontal and lateral X-ray images; indication, technique, and comparison; prior study's frontal image and report; along with a task-specific instruction.
    The \indication provides clinical context on the patient and influences interpretation and reporting.
    The \reportsection{Technique} describes acquired views and sometimes patient positioning (e.g.~supine, lateral), while \reportsection{Comparison} indicates whether the radiologist consulted prior studies.
    This example does not have a prior study so the model receives no prior frontal image or prior report.
    (Panel~B)~The \mairatwo model ingests interleaved text and images, using a frozen vision encoder (\raddino) and training an adapter and an autoregressive language model.
    Each 518\texttimes518 image is processed into patches of size 14\texttimes14 and encoded by \raddino into a sequence of 1369 visual tokens.
    We do not use the \token{CLS} token. %
    (Panel~C)~We equip the language model with coordinate tokens enabling it to describe locations on a grid over the image. Bounding boxes are represented using the top-left and bottom-right coordinates of the box. Each grounded finding is then a single sentence followed by one or more boxes, as illustrated. A non-grounded finding is simply described by a single sentence.
    }
    \label{fig:arch_diagram}
\end{figure*}

\section*{Methods}
\subsection*{Grounded radiology reporting -- a new task}
\label{sec:taskdef}

We define a grounded report as a list of sentences from the \findings section, each describing at most a single observation from the image(s), and associated with zero or more spatial annotations indicating the location of that observation if appropriate. An example is shown in \Cref{fig:arch_diagram}A.

These spatial annotations
should be as specific as possible while containing the finding. %
Non-findings (`No pneumothorax'), regions of normality (`Lungs are clear'), or abnormal findings without specific location (`Diffuse opacity') do not require spatial annotations. In this work, we use bounding boxes as spatial annotations, as they are commonly used to localise findings on \acp{CXR} \citep{nguyen2022vindr,wang2017chestx,boecking2022mscxr,müller2024weakly} and are easier to annotate than full segmentation masks. We provide a detailed annotation protocol for creating grounded reporting datasets in \cref{app:annotation_protocol}.

\begin{table*}[h]
    \centering
    \caption{Datasets used in the training and evaluation of \mairatwo. For report generation tasks (findings generation and grounded reporting), a sample consists of at least one image, a findings section, and other report sections. For phrase grounding, a sample is an image with a corresponding single phrase and one or more bounding boxes. \findgen = findings generation, \groundrep = grounded reporting, \phraseground = phrase grounding. `All' means all studies with a \findings section. Statistics on laterals and priors are percentages of samples. Having a prior means having a prior study, including a report and a frontal image. \mimiccxr: \cite{johnson2019mimic-cxr-dataset}. \mscxr: \cite{boecking2022mscxr}. \padchest: \cite{bustos2020padchest}. \privatedata is private, with a mix of in-patient and out-patient facilities in the US. \openi: \cite{demner2016preparing}. Datasets not used in evaluation have `--' for test set numbers. * \openi has no patient information so we report study information.}
    \sisetup{table-format=2.1}  %
    \small
    \setlength{\tabcolsep}{3pt}
    \tabspace
    \begin{tabular}{
        @{} lll
        S[table-format=6.0] S[table-format=4.0{*}]
        S[table-format=6.0] @{ } l S[table-format=4.0]
        SS
        SS @{}
    }
    \toprule
    \bfseries Data source & \textbf{Subset} & \textbf{Task} & \multicolumn{2}{c}{\# \textbf{Patients}} & \multicolumn{3}{c}{\# \textbf{Samples}} & \multicolumn{2}{c}{\% \textbf{Has Lateral}} & \multicolumn{2}{c}{\% \textbf{Has Prior}}\\
    \cmidrule(lr){4-5} \cmidrule(lr){6-8} \cmidrule(lr){9-10} \cmidrule(lr){11-12}
    & & & {Train}  & {Test} & \multicolumn{2}{c}{Train (\%)} & {Test} & {Train}  & {Test} & {Train} & {Test} \\
    \midrule
        \mimiccxr  & 
        All & %
            \findgen &
55218 &  %
285 &  %
158555 & (31\%) &  %
2461 &  %
60.6 &  %
45.3 &  %
64.2 &  %
88.6  %
\\
        &
        MS-CXR & %
            \phraseground &
            595 &  %
128 &  %
817 & (0.2\%) &  %
176 &  %
0 &  %
0 &  %
0 &  %
0  %
\\\midrule
        \padchest  &
        All & %
            \findgen &
52828 &  %
1559 &  %
85598 & (17\%)&  %
2925 &  %
46.0 &  %
50.4 &  %
38.3 &  %
48.1 \\  %
        \padchest  &
        \padchesttwo & %
            \groundrep &
3122 &  %
893 &  %
3183 & (0.6\%)&  %
915 &  %
44.7 &  %
45.7 &  %
32.3 &  %
31.7   %
\\\midrule
        \privatedata &
        All & %
            \findgen &
118031 &  %
{--} & %
193652 & (38\%) &  %
{--} & %

51.7 &  %
{--} & %

0 &  %
{--} %
\\
         &
            \hundredk & %
            \groundrep & %
45155 &  %
{--} & %
60463 & (12\%) &  %
{--} & %
48.0 &  %
{--} & %
0 &  %
{--}   %
\\
         &
            \eightk & %
            \groundrep & %
            8458 &  %
1199 &  %
8580 & (1.7\%) &  %
1231 &  %
81.2 &  %
79.8 &  %
0 &  %
0   %
            \\\midrule
        \openi &
        All &
        \findgen &
        {--} & %
        {3198*} & %
        {--} & %
        {--} & %
        {3306} & %
        {--} & %
        {92.1} & %
        {--} & %
        0  %
        \\\midrule
        \textbf{Total} & 
        & %
        Multi-task &
        226077  & %
        {--} & %
        510848 & (100\%) & %
        {--} & %
        53.4 & %
        {--} & %
        26.5 & %
        {--} 
        
    \\\bottomrule
    \end{tabular}%
    \label{tab:datasets}
\end{table*}

\subsection*{Data}
\label{sec:data}
We develop and evaluate \mairatwo on a set of public and private \ac{CXR} report generation datasets: \mimiccxr~\citep{johnson2019mimic-cxr-dataset}, \padchest~\citep{bustos2020padchest}, and \privatedata, a private dataset derived from a mix of US hospitals (described further in \Cref{sec:app_data}). \openi~\citep{demner2016preparing} is used as a fully held-out external evaluation set. Statistics are provided in \Cref{tab:datasets}. These datasets span in- and out-patient reporting scenarios. For each study we extract the \findings section, the current frontal (posteroanterior or anteroposterior) and lateral views, the prior study (for \mimiccxr and \padchest), and the \indication, \technique, and \comparison sections when available. 

To enable grounded reporting, we employed the proposed annotation protocol on a subset of \privatedata processed as described in \Cref{app:phrasification} (henceforth referred to as \eightk), and make use of the concurrently developed \padchesttwo grounded reporting dataset.

In total, \mairatwo is trained on 510,848 report generation or grounded reporting examples from 226,077 adult patients, including 72,226 (14\%) examples of grounded report generation. We split all datasets into training and evaluation subsets by patient.
Further data processing details provided in \Cref{sec:app_data}

\subsection*{\mairatwo architecture}
\label{sec:arch}
As depicted in \Cref{fig:arch_diagram}, \mairatwo uses a similar architecture to \maira~\citep{hyland2024maira1}, based on LLaVA~\citep{liu2023llava,liu2023llava15}.
We use a re-trained \raddinoorig~\citep{perez2024raddino} (denoted as \raddino) as the frozen image encoder, which is an 87M-parameter ViT-B~\citep{dosovitskiy2020vit}; the language model is initialised to the weights of Vicuna~7B~v1.5~\citep{vicuna2023};
 and the adapter is a randomly initialised \ac{MLP} with four layers. 
\mairatwo is trained in a multitask manner on both grounded and non-grounded reporting examples. Further training details are provided in \Cref{app:maira_details}.

\subsection*{Incorporating additional context}

Context beyond a single image plays a significant role in the contents of a radiology report, influencing both the interpretation of the image and communicative choices in the reporting itself. %
Prior work has demonstrated that using the \indication~\citep{nguyen2023pragmatic,dalla2022multimodal,hyland2024maira1}, lateral view~\citep{lee2023unixgen,mondal2023transformers,yang2020automatic,yuan2019automatic}, or prior study~\citep{bannur2023biovilt,dalla2023controllable,zhu2023utilizing} can improve generated report quality.

Hence, \mairatwo generates \ac{CXR} reports using: the current frontal image, the current lateral image, the prior frontal image and prior report, and the \reportsection{Indication}, \reportsection{Technique}, and \reportsection{Comparison} sections of the current study. These sections are interleaved with image tokens in a prompt provided to the \ac{LLM}.
Input images other than the current frontal \ac{CXR} are optional for \mairatwo.
When they are available, we likewise present their image tokens to the \ac{LLM} in a modified prompt.
Input sections are also optional and represented by the string `N/A' when missing. The full prompt is provided in \Cref{tab:maira_prompt}.

\subsection*{Supporting grounded reporting}
To enable \mairatwo to generate image annotations, we follow prior work~\citep{chen2022pix2seq,yang2022unitab,peng2023kosmos2} in adding specialised box tokens to the vocabulary of the \ac{LLM}. Each token represents a coordinate on a discretised grid of the image. Hence, to generate a bounding box, \mairatwo outputs tokens representing its top-left and bottom-right corners. As shown in \cref{fig:arch_diagram}, the box coordinates are surrounded by \token{box} delimiters, and full grounded and non-grounded sentences surrounded by \token{obj} delimiters.

Unlike prior work, we separately encode horizontal and vertical coordinates as disjoint sets of $N+N$ tokens, e.g. ``\token{x12}\token{y34}\token{x56}\token{y78}'', to help the model learn true 2D representations.
The grid size $N$ is set to 100 in all our experiments.

\subsection*{\evmetric: An evaluation suite for (grounded) reports}
\label{sec:radfact}
\begin{figure*}
    \centerline{\includegraphics[scale=0.45]{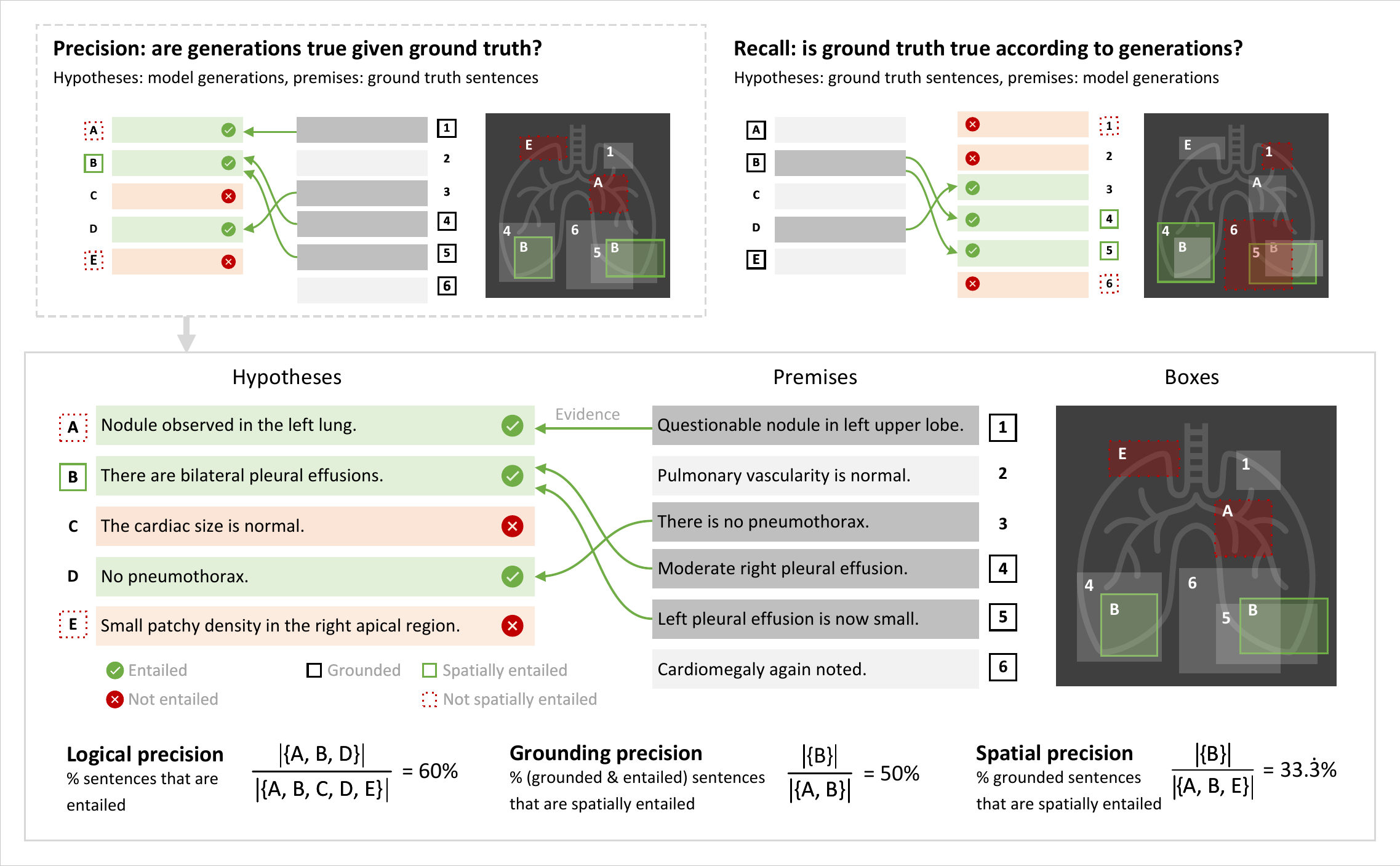}}
    
    \caption{
    Illustration of \evmetric.
    The proposed suite of \evmetric metrics enables evaluating both text reports and grounding annotations.
    It is based on logical inference, using an \ac{LLM} with task-specific prompting to classify hypotheses as entailed or not, given premises.
    The generated report is evaluated against a ground-truth report to compute precision metrics (top left), and conversely for recall metrics (top right).
    Detailed panel (bottom) shows a single direction of evaluation, taking the model generations as logical hypotheses and the original report as premises. 
    Here, logical precision measures the fraction of generated sentences that are entailed by sentences from the original report.
    Grounding precision is the fraction of \emph{logically entailed}, grounded sentences whose spatial annotations are also entailed.
    Spatial precision is the fraction of \emph{all} grounded sentences whose spatial annotations are also entailed, hence it is upper-bounded by grounding precision. 
    Here, spatial annotations of a sentence are one or more boxes (see sentence B). Spatial entailment requires that at least 50\% of the pixels associated with the sentence fall into the union of matched evidence boxes. In the above, sentence B's evidence comes from premises 4 and 5, hence its boxes are compared with the boxes from 4 and 5.
    }
    \label{fig:metric_diagram}
\end{figure*}

Traditional \ac{NLG} metrics are insufficient for radiology report generation evaluation as they treat all words equally without accounting for clinical significance. This has led to the development of radiology-specific metrics leveraging specialised models such as CheXbert~\citep{smit2020chexbert,irvin2019chexpert} or RadGraph~\citep{jain2021radgraph,yu2023evaluating,delbrouck2022rewards}, and more recently \acp{LLM}~\citep{chaves2024towards,wang2024llm}. However, existing approaches are limited in (i)~relying on pre-specified findings classes~\citep{smit2020chexbert}, specialised models~\citep{yu2023evaluating} or error types~\citep{chaves2024towards,wang2024llm}, and (ii)~not supporting the evaluation of \emph{grounded} reports.

To this end, we developed a framework called \evmetric for the evaluation of model-generated radiology reports given a ground-truth report, which enables evaluation of grounding annotations if present, and does not rely on pre-specified error categories or radiology-specialised models. Instead, \evmetric relies on the \emph{logical inference} capabilities of \acp{LLM}~\citep{liu2023radiologygpt,min2023factscore} to directly evaluate the correctness and completeness of generated reports, as illustrated in \Cref{fig:metric_diagram}. \evmetric provides a fine-grained \emph{suite} of metrics, capturing aspects of precision and recall at both text-only and text-and-grounding levels.

For report generation without grounding, \evmetric provides the following metrics:
\begin{itemize}
    \item \evmetric logical precision: the fraction of generated sentences that are entailed by the ground-truth report. This measures how truthful the model generations are, as it penalises hallucinations.
    \item \evmetric logical recall: the fraction of ground-truth sentences that are entailed by the generated report. This measures how complete the generated report is, as it penalises omissions.
\end{itemize}
When spatial annotation (grounding) is available, \evmetric further provides:
\begin{itemize}
    \item \evmetric grounding \{precision, recall\}: the fraction of \emph{logically entailed} grounded sentences that are \emph{also} spatially entailed. This tells us: which of the correctly \emph{described} findings were also \emph{correctly grounded}?
    \item \evmetric spatial \{precision, recall\}: the fraction of \emph{all} grounded sentences that are \emph{logically and spatially} entailed. This metric additionally penalises grounding incorrect sentences.
\end{itemize}

In \evmetric, we use \texttt{Llama3-70B-Instruct} \citep{llama3modelcard} (\url{https://huggingface.co/meta-llama/Meta-Llama-3-70B-Instruct}) for entailment verification with ten in-context examples -- we refer to this version as \evmetricllama. %
More details about \evmetric are available in \Cref{app:evmetric}.

\subsection*{Evaluation and metrics}
\label{sec:metrics}
We supplement \evmetric and enable comparison with prior work in report generation by additionally reporting the conventional `lexical' metric BLEU-4~\citep{papineni2002bleu}, and the radiology-specific RadCliQ version~0~\citep{yu2022evaluating-preprint}, RadGraph-F1~\citep{jain2021radgraph}, and macro-averaged CheXbert F1 score~\citep{irvin2019chexpert,smit2020chexbert}. We report a more comprehensive set of metrics in \Cref{app:extended_results}.
To quantify variance in the model's test set performance, we report median and 95\% confidence intervals over 500 bootstrapping replicates for all metrics.

We performed certain ablation experiments dropping different components of the input to \mairatwo to quantify the impact of additional report sections and images used by the model. We report two types of ablations: (i)~inference-time ablations, omitting the input at \emph{test time} only, to measure how much the model trained with that input has learned to indeed rely on it; and (ii)~training-time ablations, removing the input during both training and evaluation, to measure the overall impact of having the input available. We perform these analyses on the \mimiccxr findings generation task, as this is a public benchmark containing linkable prior images and reports, lateral images, and all the relevant report sections.

To complement our quantitative analyses, we also conducted a systematic, in-depth qualitative review of twenty random \mairatwo outputs with a thoracic radiologist (detailed in \Cref{app:qual_review}), as well as providing illustrations of \mairatwo outputs on grounded and non-grounded reporting, demonstrating success and failure cases, and enabling comparison to Med-Gemini~\citep{yang2024advancing} (\Cref{app:qualitative}).

\section*{Results}

\begin{figure*}[tp]
    \includegraphics[scale=0.55]{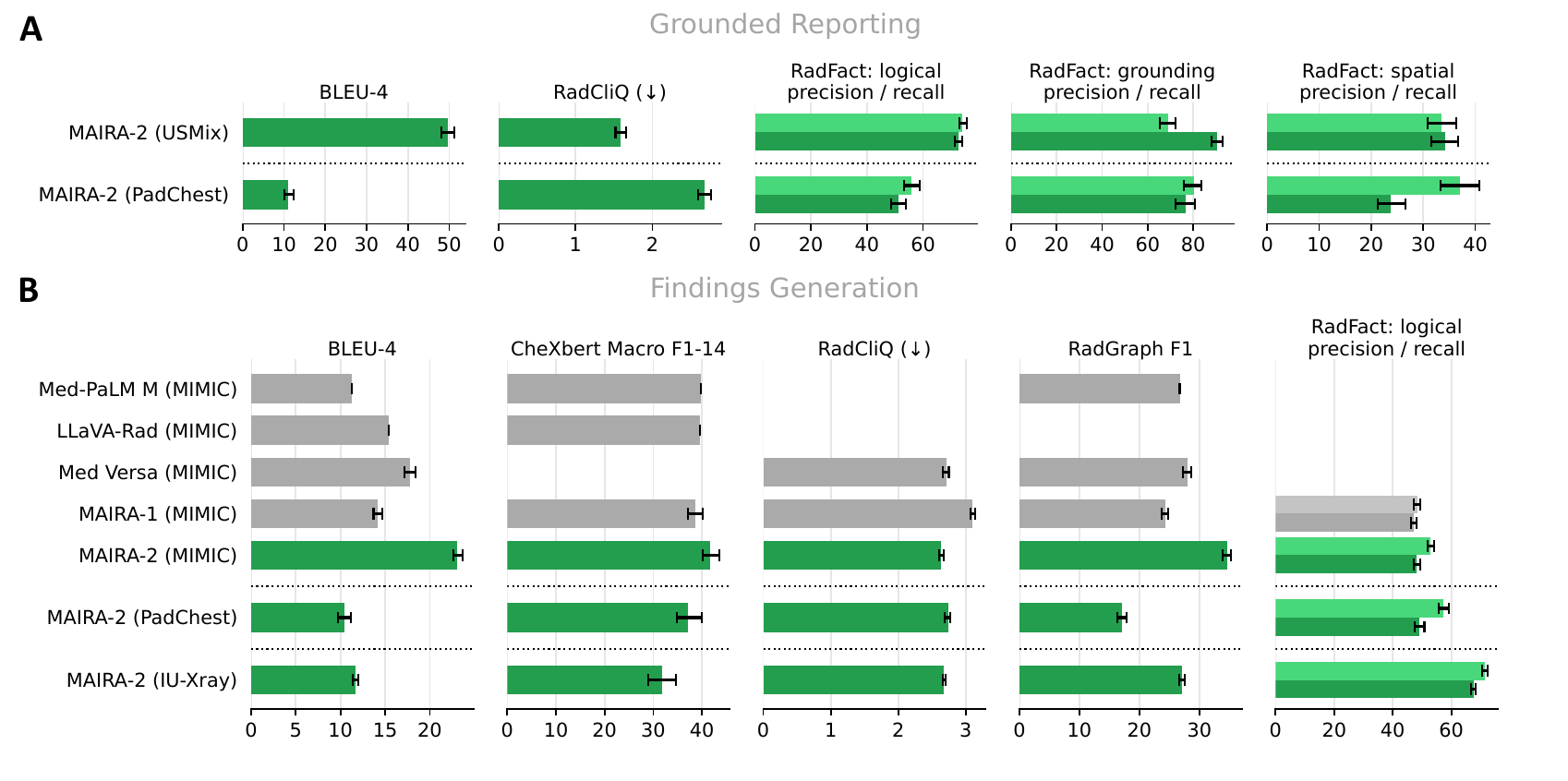}
    \caption{\textbf{\mairatwo can generate grounded reports, and establishes new state-of-the-art in non-grounded report generation.}
    (Panel~A) Performance on the grounded reporting task on \eightk (\privatedata) and \padchesttwo. \mairatwo achieves \evmetric logical precision above 50\% with high grounding precision (68.8\%, 80.2\% respectively) and moderate spatial precision (33.5\%, 37.1\%).
    (Panel~B) On \mimiccxr we compare to the closest prior state of the art, restricted to models evaluated for \findings generation, namely Med-PaLM M~\citep{tu2024medpalmm} (with a different test set, counting the laterals as individual samples), LLaVA-Rad~\citep{chaves2024towards}, MedVersa~\citep{zhou2024generalist}, and MAIRA-1~\citep{hyland2024maira1}. Since many of these models are not publicly available, we present their evaluation results as originally reported, for available metrics. For MAIRA-1, we obtained the model generations on the MIMIC-CXR test set in order to run \evmetric. There is no prior work evaluating on \padchest, hence we report \mairatwo performance to establish a benchmark. \openi is used as a fully held-out evaluation dataset. High \evmetric logical precision and recall on \openi demonstrate that \mairatwo generalises well to an unseen dataset.
    \commondescription\\
    }
    \label{fig:_figure3_report_generation}
\end{figure*}

\subsection*{MAIRA-2 establishes the new task of grounded reporting}
To the best of our knowledge, \mairatwo is the first \ac{CXR} model that both generates the full \findings section and grounds each detected finding in the image, and thus serves as a baseline for future work on this task. 
\Cref{fig:_figure3_report_generation}A shows the performance of \mairatwo on grounded report generation for \eightk and \padchesttwo.

On \eightk, \evmetric logical scores are consistently above 70\%, indicating a low rate of both omissions and hallucinations. On \padchesttwo, \evmetric logical precision and recall are 56\% and 51\%. The lower precision for \padchesttwo may be due to shorter reports in the dataset and the lower recall due to missing \indication sections in \padchesttwo, making it harder to report negatives. On \eightk, the \evmetric grounding precision indicates 69\% of the generated sentences that are logically correct are also correctly grounded, consistent with our observation that \mairatwo can also perform the related task of phrase grounding (\Cref{app:mscxr}). Conversely, the remarkable grounding recall above 90\% indicates that the model reliably covers the ground-truth boxes of correctly predicted findings.
However, the lower \evmetric spatial metrics demonstrate that the model often generates boxes that associated with incorrect sentences.
 On \padchesttwo \evmetric grounding precision and recall are more balanced, with scores of 80\% and 77\%.

\subsection*{MAIRA-2 is state-of-the-art on findings generation}
\Cref{fig:_figure3_report_generation}B shows the performance of \mairatwo on \emph{non-grounded} report generation on the \mimiccxr test set. We see that \mairatwo outperforms or matches all prior approaches across all metrics.
The impact on lexical metrics is most significant, where \mairatwo improves on prior scores by 17\% to 30\%. On existing clinical metrics, significant improvement is observed on the RadGraph-F$_1$ and on CheXbert \macrofourteen. 
For RadCliQ, \mairatwo and MedVersa have overlapping confidence intervals.
In the following sections, we explore the features of \mairatwo which result in these improvements.

With \evmetric, we see again an improvement from MAIRA-1 to \mairatwo, 
in agreement with other metrics. What \evmetric additionally reveals is that in \emph{absolute} terms, models continue to make errors, with only 52.9\% of sentences generated by \mairatwo confirmed true according to the reference report (i.e.~logical precision). We show qualitative examples of \mairatwo generations on \mimiccxr in \cref{app:qualitative_mimic}.

Although there is no prior work demonstrating findings generation performance on \padchest in English, in \Cref{fig:_figure3_report_generation}B  we show results from \mairatwo to enable future comparison. \mairatwo achieves \evmetric logical precision and recall of 57\% and 49\% on the \padchest dataset, however lexical scores are lower (ROUGE 28\%, BLEU-4 10\%). We speculate the drop in lexical metrics is due to the absence of section information (\indication, \technique, \comparison) in \padchest. In addition, the reporting style differs significantly between \padchest and \mimiccxr, which may impact the reliability of model-based metrics such as RadGraph-F$_1$ that were developed for \mimiccxr.
\Cref{fig:_figure3_report_generation}B further demonstrates that \mairatwo can generalise to the unseen dataset of \openi, achieving \evmetric logical precision and recall of 71\% and 68\% respectively.

\subsection*{Expert review reveals areas of strength and weakness}
\begin{figure*}
    \includegraphics[scale=0.37]{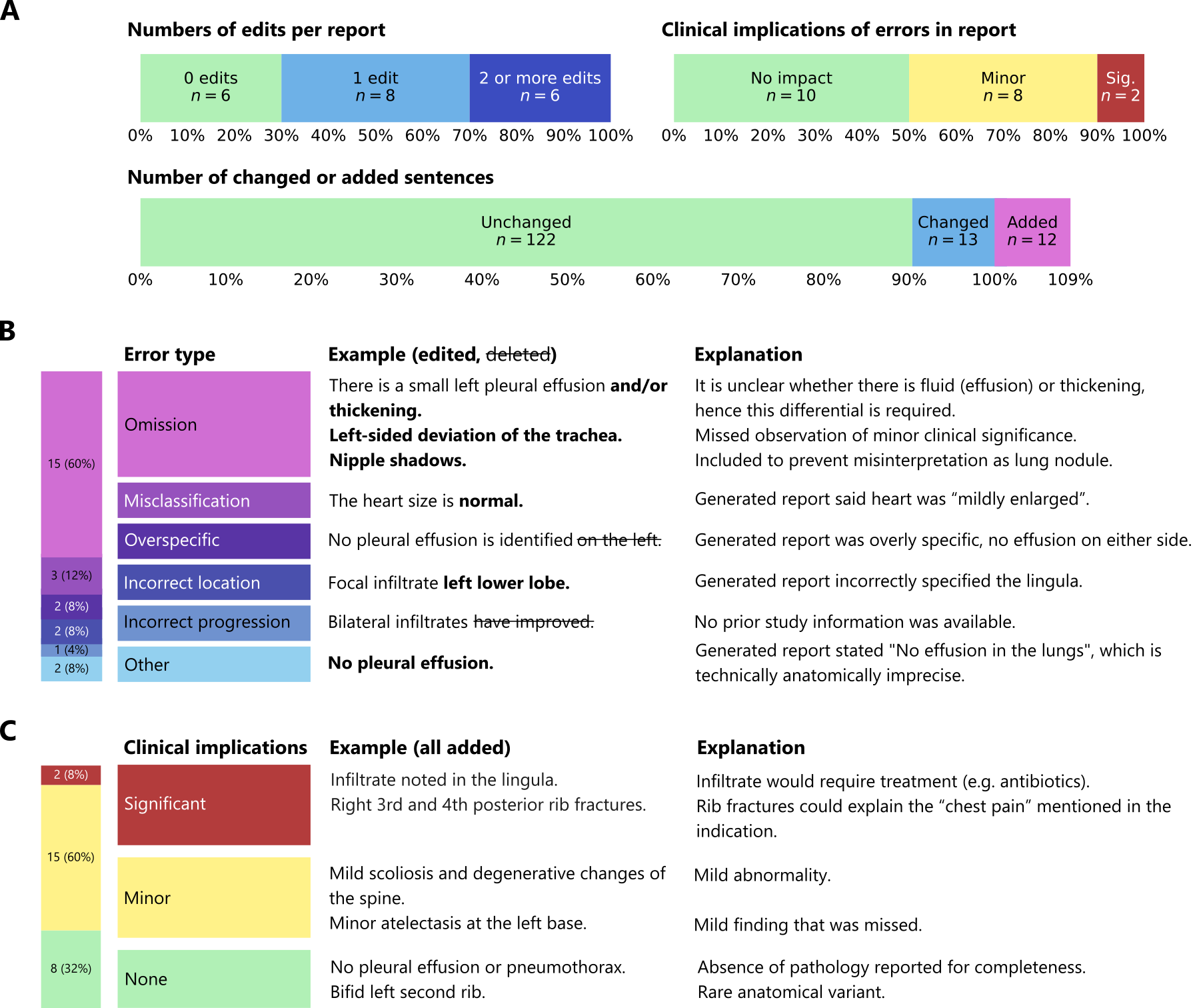}
    \caption{\textbf{In-depth qualitative review on the performance of \mairatwo on twenty randomly-selected examples from \eightk}. A thoracic radiologist was asked to assess every generated sentence and accept as-is, edit, delete, or add additional sentences. (Panel A) Of the 135 generated sentences, the majority (90\%, n=123) did not require any edits, amounting to six (30\%) fully-correct generated reports. Few edits related to clinically significant findings, with the majority of studies (90\%, n=18) having errors of no or minor clinical implications. (Panel B) Of the 25 errors (edits to sentences or additions), the majority (60\%, n=15) were omissions where \mairatwo failed to generate a finding. (Panel C) Most errors were deemed to have minor or no clinical implications (92\%, n=23). The full set of errors with explanation are provided in \Cref{tab:corrections_add,tab:corrections_del,tab:corrections_edits}.}
    \label{fig:qual_review}
\end{figure*}

Qualitative review by a thoracic radiologist of the text generated by \mairatwo on twenty random cases from \eightk (\Cref{fig:qual_review}) indicate that 14/20 reports (70\%) required  fewer than two corrections, and 123/135 generated sentences (91\%) were acceptable as-is.
With omissions being the most common error category (15 of 25 corrections), this analysis indicates model limitations include lower sensitivity on minor findings, occasional lack of internal consistency in reports, and lesser knowledge of device characteristics. The `clinical implications' of most errors were minor to none, with only two significant omissions observed.
Overall, these findings led the radiologist to conclude that the \mairatwo outputs were `acceptable as a draft', alike `the performance of a junior-to-mid level resident' that needs to receive additional human expert review before signing-off on any one report.

\subsection*{Prior studies reduce temporal hallucinations}

\begin{figure*}
\centering
    \includegraphics[scale=0.55]{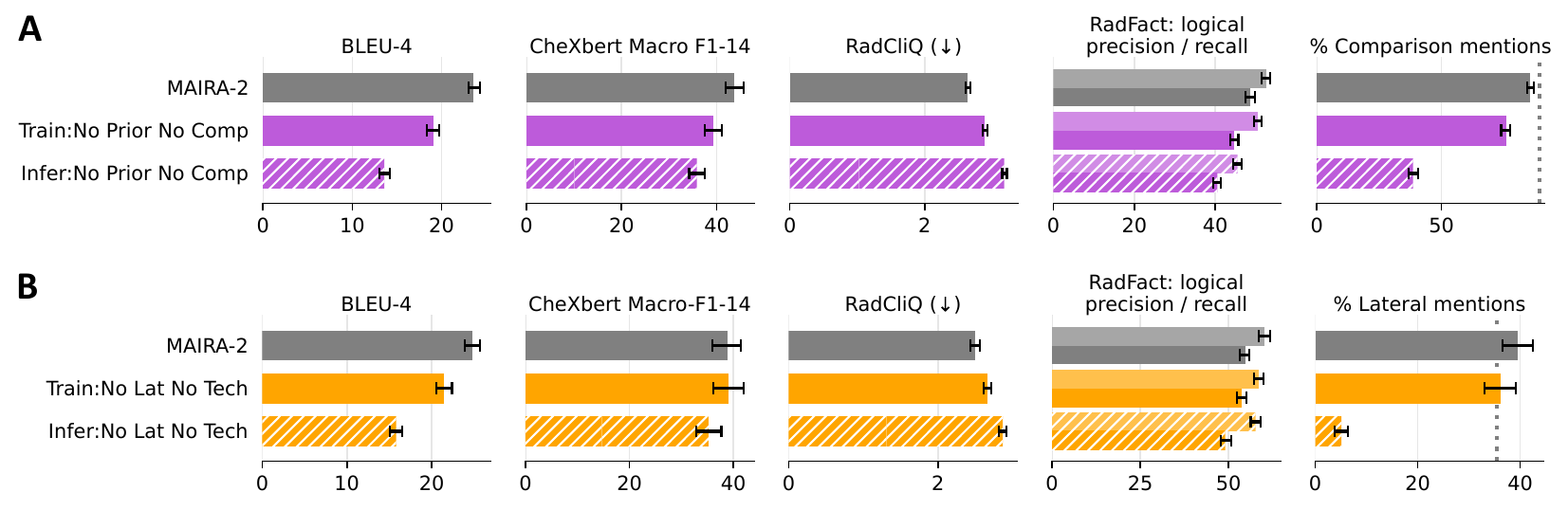}
    \caption{
    \textbf{Impact of dropping the model inputs during both training and inference (`\abTrain') and during inference only (`\abInf') on MIMIC \findings generation.}
    (Panel~A)~Dropping the prior study and comparison for the 88.6\% test subset that have a \prior (n=2181). \compmentions is estimated using Llama3-70B. The dashed line indicates the frequency of comparison mentions (91.84\%) in the ground-truth reports in the same data subset, for reference.
    (Panel~B)~Impact of dropping the lateral view and the technique section for the 30.6\% test subset that have a \lateral view (n=1,116). The dashed line indicates the frequency of lateral mentions (35.57\%) in the ground-truth reports in the same data subset, for reference.
    \plotsdescription
    Tabular representations of these results are available in \cref{tab:fg_prior_and_comparison_ablation,tab:fg_lateral_and_technique_ablation}, respectively. Note that for these ablations, we used a slightly earlier variant of \mairatwo trained without \padchesttwo.
    }
    \label{fig:_figure4_additional_inputs}
\end{figure*}

We measure the impact of prior study information through training and inference-time ablations on \mimiccxr presented in \Cref{fig:_figure4_additional_inputs}A.
As an additional metric, we use \texttt{Llama3-70B-Instruct} to determine whether a given report mentions temporal comparisons (see details in \cref{app:input_ablations}), referred to as \compmentions.
In the absence of a prior study, \compmentions should be close to zero.

Not using the prior study and \reportsection{Comparison} during training produces a significant drop across all metrics compared to the \mairatwo baseline as shown in \Cref{fig:_figure4_additional_inputs}A, and results in hallucinatory \compmentions close to the background rate of 75\% in this dataset.
Conversely, training with prior study and \reportsection{Comparison} means that when these inputs are not available for inference, the model produces significantly fewer \compmentions.
The significant drop in clinical and lexical metrics from the inference-time ablation further indicates that \mairatwo is effectively learning to use these inputs.

Additional piece-wise ablations (\cref{app:input_ablations}) show that dropping the prior study alone has a larger effect on clinical metrics such as CheXbert \macrofourteen while dropping the \reportsection{Comparison} predominantly impacts lexical metrics.

\subsection*{Multi-view inputs reduce spurious lateral mentions}
We analyse the impact of inputs related to multi-view studies, namely the lateral view and \technique section, through training and inference-time ablations on \mimiccxr in \Cref{fig:_figure4_additional_inputs}B.
 Analogously to temporal information, we quantify mentions of the lateral view in the \findings section (\lateralmentions) using regular expressions (\Cref{lst:lateral_regex}), to measure whether the model is effectively using the additional inputs. 

Not using the lateral image and the \reportsection{Technique} during training significantly decreases lexical metrics (BLEU-4 and RadCliQ), with clinical metrics (\evmetric and \macrofourteen) largely unchanged.
However, this ablated model generates hallucinatory lateral mentions close to the background rate of 36.1\% in this dataset.
Conversely, having trained with the lateral image and \reportsection{Technique} means a significant drop in hallucinatory \lateralmentions to 5.1\%.

Inference-time ablation of \mairatwo further demonstrates a marked drop in both clinical and lexical metrics in the absence of the lateral view and \reportsection{Technique}, indicating the model is learning to rely on these inputs, especially for certain pathologies. For example, the F$_1$ score for pleural effusion drops from 71.4~\ci{[66.6, 75.0]} to 64.7~\ci{[59.9, 69.5]} in the absence of the lateral view and \reportsection{Technique}.
We further analyse the impact of the lateral and the technique section separately in \Cref{app:input_ablations}.

\section*{Discussion}
Grounded radiology report generation is a novel task that requires a model to generate image-level localisations for each finding that can be localised within the image. This enables novel uses of automatically generated reports, such as potentially more rapid review of generated findings and use by non-radiologist clinicians, or even patients. In this work we have focused on the technical aspects of this new task to demonstrate its feasibility, leading to the development of \evmetric metric and construction of \mairatwo model.

\mairatwo is a large multimodal model making use of the radiology-specialised \raddino image encoder and the open Vicuna~7B~v1.5 \acl{LLM}. 
\mairatwo improves significantly upon the state of the art in findings generation on MIMIC-CXR owing to its more comprehensive set of inputs. Tailored to the \ac{CXR} setting, \mairatwo leverages the current frontal and lateral views, the prior study (frontal image and full report), the \reportsection{Indication} for the current study, as well as the \reportsection{Technique} and \reportsection{Comparison} sections. Through ablations, we have demonstrated the roles of these additional inputs in reducing hallucinations and improving clinical accuracy.
Extensive qualitative review with a radiologist, indicates that \mairatwo produces reports which may be acceptable as a `first draft' subject to consultant review, with the majority of generated sentences acceptable as-is. However, with the most commonly-observed error being that of missed finding, work to improve recall is required.

Our proposed evaluation framework, \evmetric, allows for a more nuanced assessment of automated reporting.
\evmetric targets the core objective of evaluation in report generation: to pinpoint the errors made by the model.
Using the generalisation capabilities and reasoning faculties of \acp{LLM}, \evmetric does not rely on a fixed set of finding categories or a model which is specialised to a certain reporting style, instead operating via more flexible logical inference.
Further, \evmetric provides for sentence-level granularity on model errors, and naturally supports both grounded and non-grounded reporting.
We share code for \evmetric at \radfactcode.

\evmetric however has limitations. For example, it does not distinguish between the \emph{nature} of errors beyond factuality, relying on strict logical entailment. This means some errors may be more or less clinically significant, and `partial errors' are penalised (for example, correctly describing the presence of a pneumothorax, but not that it has improved). 
By analysing a sentence at a time, it is also unable to detect internal inconsistencies in either generated or ground-truth reports, as uncovered by qualitative review. By open-sourcing \evmetric, we support further improvements to enable better evaluation standards on the task of radiology report generation including grounding.

Another limitation of this work is that neither of the grounded reporting datasets have all of the desirable inputs -- \eightk does not have priors, and \padchesttwo does not have sections other than \reportsection{Findings}. This limits our ability to probe the interaction between additional inputs and performance on grounding specifically. Further, although we conducted extensive qualitative analyses, these were predominantly with a single radiologist, limiting generalisability, especially as reporting styles can differ with geography.

Our ablations also indicate that the model may not be using additional imaging information to the fullest extent, instead exploiting shortcuts available in the report sections used as inputs. Other methods to incorporate additional imaging information may prove superior to our token concatenation approach. 

Overall we have demonstrated that grounded radiology reporting is possible with \mairatwo. Although performance in automated report generation continues to improve -- and we establish a new state-of-the-art on \mimiccxr with this work -- metrics to date, including \evmetric, indicate a gap between model performance and that which will be required to realise such systems in practice. The addition of grounding is a step towards real clinical impact in automated radiology report generation.

\section*{Acknowledgements}
We would like to acknowledge valuable inputs from (in alphabetical order): Tong	Bai, Neeltje Berger, Aurelia	Bustos, Alexandra	Eikenbary, Mary	Ellen Burt, Joaquin	Galant Herrero, Min	Gao, Will	Guyman, Houdong	Hu, Meng	Jia, Xinyang	Jiang, Gunter	Loch, Xufang	Luo, Addison	Mayberry, Flaviu	Negrean, Antonio	Pertusa, Hannah	Richardson, Abhishek	Rohatgi, José María	Salinas Serrano, Naiteek	Sangani, Manpreet	Singh, Kenji	Takeda, Ivan	Tarapov, Naoto	Usuyama, Zilong	Wang, Rui	Xia, Nishant	Yadav, and Zhengyuan	Yang.

\printbibliography

\begin{refsection}  %
\appendix
\numberwithin{figure}{section}
\numberwithin{table}{section}
\onecolumn

\startcontents[sections]
\printcontents[sections]{l}{1}{\setcounter{tocdepth}{2}}

\clearpage
\section{Extended background and related work}
\subsection{Why is grounded reporting a useful task?}
The ability to ground report findings or phrases within the relevant region in medical images 
has been described to play a significant role: (i) in assisting image understanding and radiological diagnosis \citep{chen2023grounding, yildirim2024multimodal, zou2024medrg}; and (ii) for verifying the correctness of \ac{AI} text outputs \citep{bernstein2023can} -- a key property to support the integration of automated report drafting systems in radiology workflows.
  
User research with radiologists and clinicians \citep{yildirim2024multimodal} demonstrates that although radiologists are capable of identifying relevant findings on an image via text location description alone (e.g., left lung consolidation), this can be more difficult when findings are small or overlapping (e.g., small pneumothorax, mass behind the heart); with more complex imaging; and when assessing images outside the reporter's core area of expertise. 
Grounded reporting may also have utility for non-radiology clinicians, where image grounding can support comprehension and a deeper engagement with the image beyond the text report \citep{yildirim2024multimodal}; and to improve communication with patients when reviewing image findings \citep{zou2024medrg}. %

Grounded reporting differs from the existing task of medical phrase grounding~\citep{muller2024chex, ichinose2023visual,zou2024medrg,boecking2022mscxr} in that phrase grounding aims to ground a \emph{specified} finding or phrase, typically assumed present within the image. Instead, a grounded report is a description of \emph{all} findings in an image with accompanying localisation, and does not require the phrases or findings to be provided. A variant of this task was explored in \cite{tanida2023rgrg}, where the model first located \emph{anatomical} regions before generating region-level descriptions. To overcome the many-to-many challenge faced by \cite{tanida2023rgrg}, where a single sentence in a report can describe multiple findings and hence several regions, we design a dataset such that each sentence describes at most a single finding, enabling precise localisation.

\subsection{Why do we expect additional inputs to help?}
\paragraph{Indication section:} Selective reporting of findings is mediated by the \reportsection{Indication}~\citep{nguyen2023pragmatic} for the study -- a report should `answer' any question it poses -- which further provides health context on the patient~\citep{yapp2022effect}. Empirically, providing the \reportsection{Indication} to the model improves the quality of generated reports~\citep{dalla2022multimodal,hyland2024maira1} and has become more commonplace~\citep{tu2024medpalmm,chaves2024towards,yang2024advancing}. 
\paragraph{Prior studies:} Comparison to previous imaging studies is crucial for tracking the development of disease or impact of treatment, and references to prior studies are frequent in radiology reporting~\citep{aideyan1995influence,bannur2023biovilt}. Such references can be removed to reduce hallucinations when prior studies are not available~\citep{ramesh2022improving,chaves2024towards,nguyen2023pragmatic}, or used in conjunction with prior images to enable descriptions of change~\citep{bannur2023biovilt,dalla2023controllable,zhu2023utilizing}.

\paragraph{Lateral view:}
The lateral view in a \ac{CXR} study provides complementary information to frontal (AP/PA) views. 
It is required to identify findings like vertebral compression fractures or small pleural effusions behind the diaphragm, and can assist in the detection and differentiation of conditions such as lung nodules, masses, and certain types of pneumonia. Incorporating the lateral view has been demonstrated to improve automated report generation~\citep{liu2024multi,lee2023unixgen,mondal2023transformers,yang2020automatic,yuan2019automatic}.

\paragraph{Comparison section:} The \reportsection{Comparison} section of the report indicates not simply the existence of a prior study, but whether the radiologist had access to it while writing their report.  Empirically, in the \mimiccxr dataset, when the \reportsection{Comparison} section is equivalent to `No comparison available', references to prior studies are rarely observed, in contrast to the background rate of 40\% in the full \mimiccxr dataset~\citep{bannur2023biovilt}.

\paragraph{Technique section:} The \reportsection{Technique} section of a report provides information on the view(s) available to the reporting radiologist. Further, it may disambiguate frontal views and provide information on patient positioning. Patient positioning in particular can influence the appearance of pathology such as effusions and pneumothorax.

\clearpage
\section{Extended methods}
\subsection{Datasets used to train and evaluate \mairatwo}

\label{sec:app_data}

Here we provide more details on the datasets used to train and evaluate \mairatwo. Statistics on the number of samples, number of patients, and prevalence of lateral and prior studies for each dataset are provided in \Cref{tab:datasets}.

For all datasets, we drop studies missing the \findings section. Each frontal view in a study is treated independently. If there are multiple laterals available, we select one randomly. At training time, for \mimiccxr if there are multiple frontal images in the prior study, all pairings of current and prior frontal images are used as individual samples. For \padchest we select a prior frontal randomly.

\paragraph{\mimiccxr \citep{johnson2019mimic-cxr-dataset}} 
For MIMIC-CXR we extract each report's \findings, \indication, \technique, and \comparison sections following \citet{johnson2019mimic-cxr-dataset}. We also use the MIMIC-CXR-derived phrase grounding dataset MS-CXR~\citep{boecking2022mscxr}, which contains individual phrases from reports and associated bounding boxes for a fixed set of pathologies. We follow the official \mimiccxr split~\citep{johnson2019mimic-cxr-jpg}, with the exception of studies in MS-CXR, which are not well-distributed across the official \mimiccxr splits. For MS-CXR, we create and share a patient-level split stratified by pathology, age, and sex: MS-CXR v1.1.0, \url{https://physionet.org/content/ms-cxr/}. Studies in the MS-CXR test and validation folds are not used in training -- otherwise we follow the official \mimiccxr split. We note that the official MIMIC-CXR test split is highly enriched for abnormal cases~\citep{johnson2019mimic-cxr-dataset}, hence prior studies are more common (\cref{tab:datasets}).

\paragraph{\padchest \citep{bustos2020padchest}}
The reports in the \padchest
dataset are originally in abbreviated Spanish. For the task of findings generation, we use the GPT-4-translated English version from the Interpret-CXR collection used in the RRG24 competition~\citep{xu-etal-2024-overview} (\url{https://huggingface.co/datasets/StanfordAIMI/rrg24-shared-task-bionlp}), which included only the \findings and \impression sections. %
For grounded reporting, we make use of the concurrently developed \padchesttwo dataset (unpublished; under submission).
Briefly, a subset of the original Spanish reports were processed by GPT-4 to extract individual finding sentences and translate them to English.
Radiologists then manually annotated bounding boxes for each positive finding in each study to produce a grounded reporting dataset.
We use the English version of the grounded reports in this study.
For both findings generation and grounded reporting tasks, we use the new official splits for \padchest, released as part of \padchesttwo.

\paragraph{\privatedata}
Our private dataset, \privatedata, is sourced from a set of US hospitals with a mix of in- and outpatient studies. We extract section text using GPT-4. No temporal study linkage is possible for this data source, so while we do not use prior study information, reports can contain references to prior studies. Two subsets of this dataset have been additionally annotated for grounded reporting: \eightk follows the protocol we release here (\cref{app:annotation_protocol}), whereas \hundredk followed slightly different guidelines. Protocol differences produced, for example, fewer but larger boxes per finding in \hundredk compared to \eightk, especially for bilateral findings. We consider \eightk our benchmark for grounded reporting on \privatedata and report test results on a held-out portion of it.

\paragraph{\openi \citep{demner2016preparing}}We use the entire \openi dataset for external validation for the task of \findings generation. Reports in this dataset are stored in XML format with sections pre-extracted. The \technique section was taken from each image caption. We also process the dataset to use the same indicator for deidentified information as used in MIMIC-CXR (``\_'').

\subsection{Additional \mairatwo model and training details}
\label{app:maira_details}

\paragraph{Training}
We train \mairatwo with a conventional autoregressive cross-entropy loss in a multitask setting on the dataset mix shown in \cref{tab:datasets}. Each sample in a batch has a task and input-specific prompt as outlined in \cref{tab:maira_prompt}. Following \citet{hyland2024maira1}, we do a single stage of training with a frozen image encoder, training the adapter and all the parameters of the \ac{LLM}. We train for three epochs and use the final checkpoint in evaluations. We use the AdamW optimiser~\citep{loshchilov2018adamw} with a global batch size of 128 across 16 NVIDIA A100 GPUs, a cosine scheduler with a warm-up of 0.03, and a learning rate of \num{2e-5}. In addition, we use a linear RoPE scaling factor of 1.5 in order to extend the context length of the LLM to handle up to 3 view images and additional inputs. 
\Cref{tab:maira_prompt} shows the full prompt provided to \mairatwo for each task.

\begin{table}[h]
\centering
\renewcommand{\arraystretch}{1.2}
\caption{\textbf{Prompt structure}. As shown in \cref{fig:arch_diagram}, the language model receives a sequence of tokens obtained by concatenating the following messages, replacing placeholders indicated by  \placeholder{brackets}. Each image placeholder is replaced with 1369 image tokens encoded by \raddino. Report section placeholders are replaced by the corresponding section from the sample, if available, otherwise `N/A'. For samples missing the lateral view or prior study, we entirely remove that part of the prompt, avoiding references to nonexistent image views. We show here the instruction for \groundrep. For \findgen, the instruction is simply ``provide a description of the findings in the radiology study.'' For phrase grounding, the instruction is simply ``Repeat the following as a grounded phrase with bounding boxes indicating all locations where it can be seen in the given chest X-ray image. Finding: \placeholder{phrase}``. For phrase grounding, only the current frontal view is used, without prior study information or report sections.}
\small
\tabspace
\centerline{
\begin{tabularx}{1.1\linewidth}{@{}rX@{}}
\toprule
    \textbf{Message type} & \textbf{Message} \\
    \midrule
    System & You are an expert radiology assistant tasked with interpreting a chest X-ray study. \\
    Current frontal & Given the current frontal image \placeholder{frontal\_image\_tokens} \\
    Current lateral & the current lateral image \placeholder{lateral\_image\_tokens} \\
    Prior frontal & and the prior frontal image \placeholder{prior\_image\_tokens} \\
    Prior report & PRIOR\_REPORT: \placeholder{prior\_report} \\
    Instruction & provide a description of the findings in the radiology study. Each finding should be described as a self-contained plain-text sentence. If the finding is groundable, locate the finding in the current frontal chest X-ray image, with bounding boxes indicating all locations where it can be seen in the current frontal image. Otherwise, generate just the ungrounded finding without bounding boxes \\
    \indication & INDICATION: \placeholder{indication} or `N/A' \\
    \technique & TECHNIQUE: \placeholder{technique} or `N/A' \\
    \comparison & COMPARISON: \placeholder{comparison} or `N/A'\\
    \bottomrule
\end{tabularx}}
\label{tab:maira_prompt}
\end{table}

\paragraph{More about token embeddings}
Inspired by Pix2Seq~\citep{chen2022pix2seq}, UniTAB~\citep{yang2022unitab}, and Kosmos-2~\citep{peng2023kosmos2}, \mairatwo represents a bounding box in terms of discretised coordinates representing the top-left and bottom-right corners on a uniform $N\times N$ grid ($N$ is set to 100 in all our experiments).
Kosmos-2 encodes each corner using a flat vocabulary with $N^2$ unique tokens for every possible grid location (e.g.\ ``\token{loc1234}\token{loc5678}'' for a box with corners $(0.12,0.34)$ and $(0.56,0.78)$), and UniTAB uses a shared vocabulary of $N$ tokens for both horizontal and vertical coordinates (e.g.\ ``\token{coord12}\token{coord34}\token{coord56}\token{coord78}'' for the same example box).
Because these encoding schemes offer no inductive bias for the model to learn true 2D representations, we instead choose to separately encode horizontal and vertical coordinates as disjoint sets of $N+N$ tokens, as e.g. ``\token{x12}\token{y34}\token{x56}\token{y78}''.
 The grid size $N$ is set to 100 in all our experiments.
All non-text tokens are appended to the pretrained language model's vocabulary, with corresponding embeddings initialised to the mean embedding of the existing tokens, following LLaVA~\citep{liu2023llava}.

\paragraph{Variant of \mairatwo for ablation experiments}
\label{app:maira_variant}
We conducted the ablations described in \Cref{fig:_figure4_additional_inputs,app:synergy,app:input_ablations} using a slightly earlier version of \mairatwo. This version was trained without the \padchesttwo grounded reporting dataset, and using a slightly smaller training dataset for \padchest findings generation. Hence, in these experiments we focus on \findings generation in \mimiccxr.

\subsection{Re-training \raddino}
\begin{table}[h]
    \centering
    \caption{Datasets used to train \raddino, our image encoder. There is no leakage between the training, validation and test patients in these datasets and those in \Cref{tab:datasets}.}
    \small
    \tabspace
    \begin{tabular}{@{}lS[table-format=6.0]@{}}
        \toprule
        \textbf{Data source}                            & \textbf{Num. images} \\
        \midrule
        BRAX~\citep{reis2022brax}                       & 41260 \\
        ChestX-ray8~\citep{wang2017chestx}              & 112120 \\
        CheXpert~\citep{irvin2019chexpert}              & 223648 \\
        MIMIC-CXR~\citep{johnson2019mimic-cxr-dataset}  & 368960 \\
        PadChest~\citep{bustos2020padchest}             & 136787 \\
        \privatedata (private)                          & 521608 \\
        \midrule
        \textbf{Total}                                  & 1404383 \\
        \bottomrule
    \end{tabular}
    \label{tab:datasets_rad_dino}
\end{table}

We retrained the image encoder, \raddinoorig~\citep{perez2024raddino}, for \SI{106000} iterations starting from the public ViT-B weights \citep{oquab2023dinov2}, using a global batch size of 1280 across 32 A100 GPUs.
The source datasets are the same as in~\cite{perez2024raddino}, though we excluded from the training set all images used for evaluation in this manuscript.
\Cref{tab:datasets_rad_dino} provides the number of images from each dataset to train \raddinoorig for \mairatwo, a version we call \raddino. There is no leakage between the training, validation, and test patients across the datasets in \cref{tab:datasets_rad_dino,tab:datasets}.

\subsection{Image processing}
We resized the original DICOM files isotropically with B-spline interpolation so that their shorter side was 518, min-max scaled intensities to [0, 255], and stored them as PNG files.
At training time, we centre-crop images to 518\texttimes518 pixels before applying
z-score normalisation with statistics (mean and variance) derived from MIMIC-CXR.
We used SimpleITK~\citep{itk} for all image preprocessing operations.

\subsection{Preparation of grounded reporting data}
Deriving a grounded report generation dataset from an existing narrative-style report generation dataset requires (i)~extracting sentences describing individual findings, and (ii)~acquiring spatial annotations for each finding. For this second step, we prepared a detailed annotation protocol for experts to follow, which is provided in \cref{app:annotation_protocol}. In the next section, we describe the process of extracting the sentences.

\subsubsection{Extraction of sentences from reports}
\label{app:phrasification}

Using \acp{LLM} we convert narrative reports (specifically the \reportsection{Findings} section) into lists of sentences, wherein each sentence should mention at most one finding. We do this in two places: (i)~construction of grounded reports, as described in Methods, and (ii)~to enable the use of \evmetric on narrative reports, since it operates on lists of sentences.

In \cref{lst:phrasification_system,lst:phrasification_fewshot} we show the system message and one of the few-shot examples used for this task. Due to space limitations, the complete set of few-shots will be shared alongside the metric implementation here: \radfactcode.
We use GPT-4 for this task, through a private Azure OpenAI deployment.

\begin{lstlisting}[float,caption={Instruction to GPT-4 for extracting single-finding sentences from narrative reports.},label={lst:phrasification_system}]
System: You are an AI radiology assistant. You are helping process reports from chest X-rays.
  
Please extract phrases from the radiology report which refer to objects, findings, or anatomies visible in a chest X-ray, or the absence of such.

Rules:
- If a sentence describes multiple findings, split them up into separate sentences.
- Exclude clinical speculation or interpretation (e.g. "... highly suggestive of pneumonia").
- Exclude recommendations (e.g. "Recommend a CT").
- Exclude comments on the technical quality of the X-ray (e.g. "there are low lung volumes").
- Include mentions of change (e.g. "Pleural effusion has increased") because change is visible when we compare two X-rays.
- If consecutive sentences are closely linked such that one sentence can't be understood without the other one, process them together.

The objective is to extract phrases which refer to things which can be located on a chest X-ray, or confirmed not to be present.
\end{lstlisting}

\begin{lstlisting}[float,caption={One of the examples provided to GPT-4 for extracting single-finding sentences from narrative reports. Speculative statements are removed and sentences with multiple findings are split into separate sentences. Here `orig' refers to the original sentence, and `new' is a list of derived sentences. The processed version of the report is the concatenation of the `new' sentences.},label={lst:phrasification_fewshot}]
User:
Chest x-rays were taken in the upright and lateral positions. Diffuse peribronchovascular opacities are observed throughout the lungs, which may indicate diffuse bronchopneumonia. No lobar consolidation, effusion, or pneumothorax is detected. The cardiomediastinal silhouette appears normal, and the bony structures are intact. No free air is observed below the right hemidiaphragm.

Assistant:
orig: Chest x-rays were taken in the upright and lateral positions.
new:
- 

orig: Diffuse peribronchovascular opacities are observed throughout the lungs, which may indicate diffuse bronchopneumonia.
new:
- Diffuse peribronchovascular opacities are observed throughout the lungs.

orig: No lobar consolidation, effusion, or pneumothorax is detected.
new:
- No lobar consolidation is detected.
- No effusion is detected.
- No pneumothorax is detected.

orig: The cardiomediastinal silhouette appears normal, and the bony structures are intact.
new:
- The cardiomediastinal silhouette appears normal.
- The bony structures are intact.

orig: No free air is observed below the right hemidiaphragm.
new:
- No free air is observed below the right hemidiaphragm.
\end{lstlisting}

This process changes the distribution of words and pathologies in the reports. To quantify this, in \cref{tab:compare_phrasification} we convert the MIMIC validation set to lists of sentences, and compare it to the original reports using standard report generation metrics.  For pathology-level CheXbert metrics, specificity is above 97\% for all classes, indicating the conversion into sentence lists does not produce \emph{additional} mentions of findings. For most findings, the recall is very high, indicating little loss. The notable exception is pneumonia, with recall of $\approx 3.4\%$, indicating that over 96\% of mentions of pneumonia in the original reports have been removed by this processing. This is expected because pneumonia is a clinical interpretation of other findings, often described with speculative language such as `... opacity suggesting pneumonia', and the prompt directs the \ac{LLM} to remove clinical speculation and interpretation.
\begin{table}
    \centering
    \caption{Conversion of reports into lists of sentences alters the distribution of words and pathologies. We use typical report generation metrics to compare the modified reports with the originals, using the \mimiccxr validation set.}
    \label{tab:compare_phrasification}
    \small
    \tabspace
    \begin{tabular}{@{}l|c@{}}
    \toprule
       \textbf{Metric} & \textbf{Modified} \\
       \midrule
        ROUGE-L & 82.1 \ci{[81.8, 82.5]} \\
        RG\textsubscript{ER} & 91.2 \ci{[90.9, 91.5]} \\
        \multicolumn{2}{@{}l}{\it CheXpert, uncertain as negative:}\\
        \hspace{\tabcolsep}\macrofourteen & 87.0 \ci{[86.3, 87.7]} \\
        \hspace{\tabcolsep}\macrofive & 93.6 \ci{[92.7, 94.3]}\\
        \hspace{\tabcolsep}Recall - Atelectasis & 91.2 \ci{[89.8, 92.6]}\\
        \hspace{\tabcolsep}Recall - Cardiomegaly & 96.2 \ci{[95.2, 97.1]}\\
        \hspace{\tabcolsep}Recall - No Finding & 96.9 \ci{[94.6, 96.7]}\\
        \hspace{\tabcolsep}Recall - Pneumonia & 3.4 \ci{[1.2, 6.7]}\\
        \bottomrule
    \end{tabular}
\end{table}

\clearpage
\section{\evmetric metric}
\label{app:evmetric}

\subsection{Extended description}
Due to space limitations in the main text, we provide further explanation of \evmetric here to complement \Cref{fig:metric_diagram}.
\paragraph{Logical entailment}
Inspired by approaches such as FActScore~\citep{min2023factscore}, we leverage a model that can perform entailment verification \citep{sanyal2024ev} to classify whether a candidate sentence (`hypothesis') is logically true given a reference text (`premise'). A class of models suitable for entailment verification are \acp{LLM}~\citep{liu2023exploring}.

The task is illustrated in \cref{fig:metric_diagram}.
The generated and ground-truth reports are assumed to consist of lists of sentences, each describing a single finding. In a conventional findings-generation scenario, free-text reports can first be converted into this format as described in \cref{app:phrasification}.

\evmetric computes entailment in both directions, defining the following text-level metrics:
\begin{enumerate}
    \item \evmetric logical precision: the fraction of generated sentences that are entailed by the ground-truth report. This measures how truthful the model generations are, as it penalises hallucinations.
    \item \evmetric logical recall: the fraction of ground-truth sentences that are entailed by the generated report. This measures how complete the generated report is, as it penalises omissions.
\end{enumerate}
This bidirectional approach differs from traditional factual verification approaches such as FActScore that assume a `single' source of truth (e.g., Wikipedia), but has precedents in medical summarisation where both completeness and conciseness are important~\citep{xie2023doclens}.

We further require the entailment verification model to provide \emph{evidence} for its classification: this is the set of premise sentences from the reference report that support the determination of entailment (or not) for each hypothesis.
Evidence may be empty for logically neutral statements, which are considered not-entailed by definition.
Evidence enables us to match the grounding regions from generated sentences with their (supposed) ground-truth regions.
Note that \evmetric does not require a one-to-one mapping between generated and reference sentences, and there can be several pieces of evidence to support a logical inference. For example, the sentence `bilateral pleural effusions' implies both `left pleural effusion' and `right pleural effusion' simultaneously, hence it can be used as evidence for either. Conversely, \emph{both} `left pleural effusion' and `right pleural effusion' are required to support the conclusion of `bilateral pleural effusions'.

\paragraph{Spatial and grounding entailment}
We can then define a notion of \emph{spatial entailment} based on pixel overlap: a region is spatially entailed by its evidence region(s) if at least a given fraction of its pixel mask is contained in the evidence pixel mask.
Specifically, this pixel-precision threshold is set to 0.5 in our implementation with multiple boxes as the form of grounding, but could be adjusted, e.g., for finer-grained segmentation masks.

This definition interprets a larger region as \emph{more specific} than a smaller region contained within it, as the former makes stronger claims about where a finding is located.
This provides for metrics on the text-and-grounding quality, analogously defining precision based on sentences from the generated report, and recall based on sentences from the ground-truth report:
\begin{enumerate}
    \item \evmetric grounding \{precision, recall\}: the fraction of \emph{logically entailed} grounded sentences that are \emph{also} spatially entailed. This tells us: which of the correctly \emph{described} findings were also \emph{correctly grounded}?
    \item \evmetric spatial \{precision, recall\}: the fraction of \emph{all} grounded sentences that are \emph{logically and spatially} entailed. This metric additionally penalises grounding incorrect sentences.
\end{enumerate}

These fractions are calculated once in each direction: `precision' scores describing the correctness of generated findings with respect to the ground-truth report, and conversely `recall' scores indicating their completeness.

Note that by design, \evmetric handles scenarios where a finding can have multiple boxes, for example `Bilateral pleural effusion.' It can also handle any image annotation in the form of a pixel mask, such as a segmentation mask.

\subsection{Implementation details}
\Cref{lst:radfact_system,lst:radfact_fewshot,lst:radfact_query} show the system message, sample few-shot examples, and a sample query for \evmetric.
The \ac{LLM} is prompted to produce valid YAML outputs that can easily be parsed, which is enforced with Pydantic (\url{https://github.com/pydantic/pydantic}) via LangChain (\url{https://www.langchain.com/}). 
As in \cref{app:phrasification}, due to space limitations we show only one of the few-shot examples -- the rest can be found in the code repository: \radfactcode.
Following chain-of-thought style prompting~\citep{wei2022chainofthought}, we found that prompting the assistant to provide the evidence before the classification (``\texttt{status}'') improved performance.
\begin{lstlisting}[float,caption={System message used for \evmetric, instructing the \ac{LLM} to assess the correctness of a single sentence given a list of reference sentences.},label={lst:radfact_system}]
System: You are an AI radiology assistant. Your task is to assess whether a statement about a chest X-ray (the "hypothesis") is true or not, given a reference report about the chest X-ray. This task is known as entailment verification. If the statement is true ("entailed") according to the reference, provide the evidence to support it.
\end{lstlisting}

\begin{lstlisting}[float,caption={Two of the examples used in the entailment verification task in \evmetric. The model is tasked with assigning a logical status (either \texttt{entailment} or \texttt{not\_entailment}) to the hypothesis sentence, given the list of reference sentences. The `evidence' field is a list of reference sentences supporting the logical state. For `Degenerative changes are seen throughout the spine', nothing in the reference sentences indicates this is true, so it is labelled with \texttt{not\_entailment}. `There is persistent consolidation in the left lung base' implies `Left basilar consolidation is present', so it is labelled with \texttt{entailment}. Note that the reverse does not hold, due to the additional detail of persistence.},label={lst:radfact_fewshot}]
User:
reference:
- The lungs are clear.
- The cardiomediastinal silhouette is unremarkable.
- There are no pleural effusions.

hypothesis: Degenerative changes are seen throughout the spine.

Assistant:
phrase: Degenerative changes are seen throughout the spine.
evidence: []
status: not_entailment
----------------------
User:
reference:
- A moderate size left pleural effusion slightly larger in size.
- Pacemaker is unchanged.
- Right lung is clear.
- There is persistent consolidation in the left lung base.

hypothesis: Left basilar consolidation is present.

Assistant:
phrase: Left basilar consolidation is present.
evidence:
- There is persistent consolidation in the left lung base.
status: entailment
\end{lstlisting}

\begin{lstlisting}[float,caption={An example query to \evmetric. Based on the reference sentences, the model must determine the logical state of the hypothesis.},label={lst:radfact_query}]
User:
reference:
- The heart is borderline in size.
- There is no evidence of CHF.
- No infiltrate.
- The diaphragm is well-visualized.

hypothesis: There is a new abnormal density filling most of the right hemithorax.
\end{lstlisting}

Using Llama3-70B as a backbone instead of GPT-4 -- as in \cite{chaves2024towards} -- provides multiple advantages: It is open-source and faster, making it more accessible to the research community and advantageous when evaluating large volumes. In \cref{tab:radfact_llama3_vs_gpt4}, we compare the performance and throughput of \evmetric using Llama3-70B and GPT-4. We measure performance on the binary task of entailment verification: classifying a given hypothesis sentence as entailed or not, given a list of references. In practice, to compute \evmetric we need to process one such query per sentence in the report, in each direction. This results in,  on average, six to seven queries per report. In this light, the performance drop observed in \cref{tab:radfact_llama3_vs_gpt4} seems neglegible relative to the gain in throughput.

\begin{table}[htb]
\centering
\caption{Accuracy and speed of \evmetric using Llama3-70B and GPT-4 as backbones. Llama3 runs on a single compute node with four A100 GPUs. GPT-4 is hosted on Microsoft Azure.}
\small
\tabspace
\begin{tabular}{lcc}
\toprule
& \textbf{Accuracy (\%)} & \textbf{Inference speed ($s/$report)} \\
\midrule
Llama3 & 92.0 & 17.35 \\
GPT-4 & 93.2 & 27.06 \\
\bottomrule
\end{tabular}
\label{tab:radfact_llama3_vs_gpt4}
\end{table}

\evmetricllama shows high alignment with the errors spotted by radiologists in the ReXVal dataset \citep{rexval_dataset}. The Kendall rank correlation coefficient between the error counts in ReXVal and the logical F1-score of \evmetric (computed as the harmonic mean between the logical precision and the logical recall) is 0.59 \ci{[0.51, 0.66]} (0.62 \ci{[0.55, 0.68]} for clinically significant errors). Confidence intervals were computed using bootstrapping with $n=1000$ in concordance with \citet{yu2023evaluating}. While the correlation of \evmetric is smaller than of the recently proposed CheXprompt~\citep{chaves2024towards}, the latter presents an attempt to directly count the different errors using a LLM. In contrast, \evmetric is not restricted to the six error types defined in ReXVal, and can perform entailment verification for any sentence that can potentially occur in a report, naturally leading to a lower alignment with ReXVal. We found, for example, mentions of lateral images in reports from all datasets used for training \mairatwo. Hallucinations or omissions of such mentions would not be detected by CheXprompt.

\clearpage
\section{Extended results}
\label{app:extended_results}
\subsection{Description of additional metrics}
Owing to the complexity of evaluating natural language generation, and the specific requirements of radiology report generation, a variety of metrics are used and have been developed. We supplement the results presented in \Cref{fig:_figure3_report_generation} with additional text metrics, and a metric to evaluate the quality of grounding alone.

\paragraph{Text-only evaluation.}
We employ a combination of traditional \ac{NLG} (`lexical') metrics and radiology-specific (`clinical') metrics. For lexical metrics, we use ROUGE-L~\citep{lin2004rouge}, BLEU-\{1,4\}~\citep{papineni2002bleu}, and METEOR~\citep{banerjee2005meteor}. For clinical metrics, we use RadGraph-F1~\citep{jain2021radgraph}, RG\textsubscript{ER}~\citep{delbrouck2022rewards}, RadCliQ version 0~\citep{yu2022evaluating-preprint}, and CheXbert vector similarity \citep{smit2020chexbert,yu2023evaluating}, as well as macro- and micro-averaged F1 scores for CheXpert classes~\citep{irvin2019chexpert} based on the CheXbert classifier~\citep{smit2020chexbert}. RG\textsubscript{ER} is implemented as \texttt{F1RadGraph} with \texttt{reward=partial} by \url{https://pypi.org/project/radgraph/}, and for RadGraph-F1, RadCliQ, and CheXbert vector similarity, we use \url{https://github.com/rajpurkarlab/CXR-Report-Metric}. For BLEU and Radgraph, which are case-sensitive metrics, we lowercase the text prior to computing the metric. We further report CheXprompt scores, which uses GPT-4 to estimate the number of errors in a generated report. Following \cite{chaves2024towards}, we report the mean errors per report, as well as the percentage of error-free reports, distinguishing between any errors, and significant errors.

\paragraph{Grounding-only evaluation.}
To evaluate bounding boxes independently of text generation, we employ a box-completion approach similar to \citet{peng2023kosmos2}.
The model is conditioned on the prompt and the grounded report up to and including the target phrase and the first \token{box} token, and is allowed to generate boxes until a closing \token{/obj} token is produced.
We do this for every grounded phrase over all reports in the dataset, then compute spatial overlap metrics between the pixel masks of the completed boxes and of the respective ground-truth boxes.
Note that \evmetric quantifies grounding on the sentence level in a binary fashion, whereas this complementary pixel-level evaluation measures the quality of the boxes in isolation.

\subsection{Findings generation -- additional results}
\label{app:fg_mimic_extended}

\begin{table}[h!]
\centering
\caption{\textbf{Findings generation performance on the official MIMIC-CXR test split}. $^\dagger$ means numbers were taken from prior work, except for \evmetric and CheXprompt for MAIRA-1 \citep{hyland2024maira1}.
\maintablesdescription
This figure complements \Cref{fig:_figure3_report_generation} with additional metrics and more precise numbers.
}
\small
\tabspace
\setlength{\tabcolsep}{3pt}
\centerline{%
\begin{tabular}{@{} l l|c c c c|c @{}}\toprule
    \multicolumn{2}{@{}l|}{\textbf{Metric}}  &    \textbf{\maira} & 
    \textbf{Med-PaLM M\citep{tu2024medpalmm}$^\dagger$} &
    \textbf{LLaVA-Rad\citep{chaves2024towards}$^\dagger$} &
    \textbf{MedVersa\citep{zhou2024generalist}$^\dagger$} & 
    \textbf{\mairatwo}
    \\
    \midrule
    \multicolumn{2}{@{}l|}{Lexical:} &&&&& \\
    & ROUGE-L & 
        28.9 {\scriptsize\textcolor{gray}{[28.4, 29.4]}} & %
        27.29 & %
        30.6 & %
        -- & %
        \bfseries 38.4 \ci{[37.9, 39.0]} %
        \\
    & BLEU-1 & 
        39.2 {\scriptsize\textcolor{gray}{[38.7, 39.8]}} & %
        32.41 & %
        38.1 & %
        -- & %
        \bfseries 46.0 \ci{[45.3, 46.7]} %
        \\
    & BLEU-4 & 
        14.2 {\scriptsize\textcolor{gray}{[13.7, 14.7]}} & %
        11.31 & %
        15.4 & %
        17.8 \ci{[17.2, 18.4]} & %
        \bfseries 23.1 \ci{[22.6, 23.7]} %
        \\
    & METEOR & 
       33.3 {\scriptsize\textcolor{gray}{[32.8, 33.8]}} & %
        -- & %
        -- & %
        -- & %
        \bfseries 41.7 \ci{[41.1, 42.4]} %
        \\
    \midrule
    \multicolumn{2}{@{}l|}{\evmetric:} &&&&& \\
    & Logical precision &
        48.3 \ci{[47.3, 49.4]} & %
        -- & %
        -- & %
        -- & %
        \bfseries 52.9 \ci{[51.8, 54.2]} %
        \\
    & Logical recall & 
       47.2 \ci{[46.3, 48.2]} & %
        -- & %
        -- & %
        -- & %
        48.2 \ci{[47.3, 49.4]} %
        \\
        \midrule
    \multicolumn{2}{@{}l|}{Clinical:} &&&&& \\
    & RadGraph-F1 & 
        24.3 {\scriptsize\textcolor{gray}{[23.7, 24.8]}} & %
        26.71 & %
        -- & %
        28.0 \ci{[27.3, 28.7]} & %
        \bfseries 34.6 \ci{[33.9, 35.3]}  %
        \\
    & RG\textsubscript{ER} & 
        29.6 {\scriptsize\textcolor{gray}{[29.0, 30.2]}} & %
        -- & %
        29.4 & %
        -- & %
        \bfseries 39.6 \ci{[39.0, 40.3]} %
        \\
    & RadCliQ ($\downarrow$) & 
        3.10 {\scriptsize\textcolor{gray}{[3.07, 3.14]}}& %
        -- & %
        -- & %
        2.71 \ci{[2.66, 2.75]} & %
        \bfseries 2.64 \ci{[2.61, 2.67]}  %
        \\
    & CheXbert vector &
        44.0 \ci{[43.1, 44.9]} & %
        -- & %
        -- & %
        46.4 \ci{[45.5, 47.4]} & %
        \bfseries 50.7 \ci{[49.9, 51.5]} %
        \\
    & {\it CheXprompt:} & & & & & \\
    & \hspace{\tabcolsep} Mean significant errors ($\downarrow$) &
        2.41 \ci{[2.35, 2.46]} & %
        -- & %
        \bfseries 2.25 & %
        -- & %
        \bfseries 2.21 \ci{[2.16, 2.26]} %
        \\
     & \hspace{\tabcolsep} Mean errors ($\downarrow$) &
        2.49 \ci{[2.44, 2.54]} & %
        -- & %
        2.95 & %
        -- & %
        \bfseries 2.29 \ci{[2.24, 2.34]} %
        \\
    & \hspace{\tabcolsep} \% Significant error free &
        4.65 \ci{[3.88, 5.55]} & %
        -- & %
        \bfseries 6.79 & %
        -- & %
        \bfseries 6.50 \ci{[5.53, 7.52]} %
        \\
     & \hspace{\tabcolsep} \% Error free &
        3.13 \ci{[2.43, 3.86]} & %
        -- & %
        2.58 & %
        -- & %
         \bfseries 4.79 \ci{[3.96, 5.73]} %
        \\
    & \multicolumn{2}{l}{\it CheXpert F1, uncertain as negative:} &&&& \\
    & \hspace{\tabcolsep} Macro-F1-14 & 
        38.6 {\scriptsize\textcolor{gray}{[37.1, 40.1]}} & %
        39.83 & %
        39.5 & %
        -- & %
        \bfseries 41.6 \ci{[40.1, 43.5]}  %
        \\
    & \hspace{\tabcolsep} Micro-F1-14 & 
        55.7 {\scriptsize\textcolor{gray}{[54.7, 56.8]}} & %
        53.56 & %
        \textbf{57.3} & %
        -- & %
        \bfseries 58.1 \ci{[57.0, 59.1]} %
        \\
    & \hspace{\tabcolsep} Macro-F1-5 & 
        47.7 {\scriptsize\textcolor{gray}{[45.6, 49.5]}} & %
        \bfseries 51.60& %
        47.7 & %
        -- & %
        \bfseries 50.4 \ci{[48.6, 52.5]} %
        \\
    & \hspace{\tabcolsep} Micro-F1-5 & 
        56.0 {\scriptsize\textcolor{gray}{[54.5, 57.5]}} & %
        \bfseries 57.88& %
        57.4 & %
        -- & %
        \bfseries 59.1 \ci{[57.6, 60.5]} %
        \\
    \bottomrule
\end{tabular}%
}
\label{tab:findings_generation_mimic_extended}
\end{table}

\begin{table}[h!]
\centering
\caption{\textbf{Findings generation performance on \padchest}.
\maintablesdescription This table complements \Cref{fig:_figure3_report_generation}.
}
\small
\tabspace
\centerline{%
\begin{tabular}{@{} l l c @{}}\toprule
    \multicolumn{2}{@{}l}{\textbf{Metric}}  &    
    \textbf{\mairatwo}
    \\
    \midrule
    \multicolumn{2}{@{}l}{Lexical:} \\
    & ROUGE-L & 
        27.7 \ci{[26.8, 28.7]}
        \\        
    & BLEU-1 & 
        25.1 \ci{[23.9, 26.2]}
        \\        
    & BLEU-4 & 
        10.4 \ci{[9.7, 11.2]}
        \\
    & METEOR & 
        29.2 \ci{[28.2, 30.2]}
        \\
    \midrule
    \multicolumn{2}{@{}l}{\evmetric:} \\
    & Logical precision &
        57.3 \ci{[55.6, 59.2]}
        \\        
    & Logical recall & 
        49.2 \ci{[47.4, 50.8]}
        \\       
    \midrule
    \multicolumn{2}{@{}l}{Clinical:} \\
    & RadGraph-F1 & 
        17.1 \ci{[16.3, 17.9]}
        \\
    & RG\textsubscript{ER} & 
        21.9 \ci{[20.8, 22.9]}
        \\
    & RadCliQ ($\downarrow$) & 
        2.74 \ci{[2.69, 2.77]}
        \\
    & CheXbert vector &
        70.6 \ci{[69.8, 71.4]}
        \\
    & \multicolumn{2}{l}{\it CheXpert F1, uncertain as negative:} \\
    & \hspace{\tabcolsep} Macro-F1-14 & 
        37.2 \ci{[34.8, 40.0]}
        \\
    & \hspace{\tabcolsep} Micro-F1-14 & 
        60.1 \ci{[58.4, 61.7]}
        \\
    & \hspace{\tabcolsep} Macro-F1-5 & 
        38.9 \ci{[35.6, 42.7]}
        \\
    & \hspace{\tabcolsep} Micro-F1-5 & 
        49.7 \ci{[46.3, 52.9]}
        \\ 
    \bottomrule
\end{tabular}%
}
\label{tab:findings_generation_padchest_extended}
\end{table}

\begin{table}[h!]
    \centering
\caption{\textbf{Findings generation performance on \openi.} We use \openi as a held-out dataset, hence evaluate the generalisation ability of \mairatwo here. \commondescription 
This table complements \Cref{fig:_figure3_report_generation} and provides a comparison to LLaVA-Rad~\citep{chaves2024towards}.}
\small
\tabspace
\setlength{\tabcolsep}{3pt}
\centerline{%

\begin{tabular}{@{} l c c c c c c c c @{}}\toprule
    & 
    \textbf{ROUGE-L} & 
    \textbf{BLEU-4} &
    \multicolumn{2}{c}{\textbf{CheXbert}} &
    \textbf{RadCliQ} &
    \textbf{RadGraph-F1} &
    \multicolumn{2}{c}{\textbf{\evmetric Logical}}
    \\
    \textbf{Model} & %
    & %
    & %
    \textbf{\macrofourteen} & %
    \textbf{\microfourteen} & %
    & %
    & %
    \textbf{Precision} &
    \textbf{Recall} \\
\midrule

\mairatwo &
    \bfseries 27.4 \ci{[27.0, 27.7]} & %
    11.7 \ci{[11.4, 12.0]} & %
    31.9 \ci{[29.0, 34.7]} & %
    52.5 \ci{[50.8, 54.2]} & %
    2.68 \ci{[2.66, 2.70]} & %
    27.1 \ci{[26.6, 27.6]} & %
    71.4 \ci{[70.5, 72.3]} & %
    67.6 \ci{[66.7, 68.3]}   %
    \\
LLaVA-Rad &
    25.3 \ci{[25.0, 25.7]} &
    -- &
    -- &
    53.5 \ci{[51.6, 55.8]} & %
    -- & %
    -- & %
    -- &
    -- \\
    \bottomrule
\end{tabular}}
\label{tab:findings_generation_openi}
\end{table}

\Cref{tab:findings_generation_mimic_extended,tab:findings_generation_padchest_extended,tab:findings_generation_openi} show extended metrics for findings generation performance on \mimiccxr, \padchest, and \openi to complement \Cref{fig:_figure3_report_generation}. For \openi (\Cref{tab:findings_generation_openi}, we additionally report the performance of LLaVA-Rad~\citep{chaves2024towards} as a comparison for the \emph{held-out} performance on the findings generation task, since most prior work uses a portion of \openi for training, unlike this work.
\mairatwo produces higher ROUGE-L scores and statistically equivalent CheXbert Micro F$_1$-14 scores. One risk associated with using additional inputs (such as the \technique and \comparison sections, which LLaVA-Rad does not use) is that \mairatwo would over-rely spurious, dataset-level associations between these inputs and the \findings section. However, our findings on \openi suggest this has not occurred to a significant degree. In particular, the high \evmetric scores suggest that \mairatwo may be producing higher-quality reports than it does on \mimiccxr, however this may also reflect that \openi is an `easier' dataset than \mimiccxr.

\FloatBarrier  %
\subsection{Grounded report generation -- additional results}
\Cref{tab:grounded_reporting_extended,tab:grounded_reporting_padchest_extended} show further metrics for the grounded reporting task on \eightk and \padchesttwo respectively.

\begin{table}[h!]
\centering
\caption{\textbf{Grounded reporting performance on the test fold of \eightk.}
\commondescription
This table complements \Cref{fig:_figure3_report_generation}.}
\small
\tabspace
\setlength{\tabcolsep}{5pt}
\centerline{%
\begin{tabular}{@{} ll c@{\hspace{5pt}}c @{}}
\toprule
    \multicolumn{2}{@{}l}{\textbf{Metric}}  &
    \multicolumn{2}{c}{\textbf{\mairatwo}}
    \\%\eightk  \\ 
\midrule
    \multicolumn{2}{@{}l}{Lexical: ROUGE-L} &
        \multicolumn{2}{c}{59.2 \ci{[57.8, 60.7]}} %
        \\
\midrule
    \multicolumn{2}{@{}l}{\evmetric:} & Precision & Recall\\
    & Logical  & 
        74.1 \ci{[72.9, 75.6]} & %
        72.8 \ci{[71.4, 74.0]} %
 \\ %
    & Spatial & 
        33.5 \ci{[30.9, 36.4]} & %
        34.2 \ci{[31.6, 36.7]} %
    \\
    & Grounding & 
        68.8 \ci{[65.5, 72.2]} & %
        90.6 \ci{[88.1, 93.0]} %
        \\
\midrule
    \multicolumn{2}{@{}l}{Clinical:} & \\
    & RadGraph-F$_1$ & 
        \multicolumn{2}{c}{55.3 \ci{[53.7, 56.9]}} %
 \\ %
    & \rger & 
        \multicolumn{2}{c}{57.8 \ci{[56.2, 59.3]}} %
 \\ %
    & RadCliQ ($\downarrow$) & 
        \multicolumn{2}{c}{1.59 \ci{[1.52, 1.66]}} %
 \\ %
    & CheXbert \macrofourteen &
        \multicolumn{2}{c}{43.6 \ci{[38.1, 50.2]}} %
 \\ %
     & CheXbert \microfourteen &
        \multicolumn{2}{c}{61.1 \ci{[58.5, 63.3]}} %
 \\ %
\midrule
    \multicolumn{2}{@{}l}{Phrase grounding:} & Precision & Recall \\   
    & Box-completion &
    68.7 \ci{[67.3, 70.1]} & %
    84.1 \ci{[83.2, 85.0]} %
    \\
    \bottomrule
\end{tabular}%
}
\label{tab:grounded_reporting_extended}
\end{table}

\begin{table}[h!]
\centering
\caption{\textbf{Grounded reporting performance on the test fold of \padchesttwo.}
\commondescription
This table complements \Cref{fig:_figure3_report_generation}.}
\small
\tabspace
\setlength{\tabcolsep}{5pt}
\centerline{%
\begin{tabular}{@{} ll cc @{}}
\toprule
    \multicolumn{2}{@{}l}{\textbf{Metric}}  & 
    \multicolumn{2}{c}{\textbf{\mairatwo}}
    \\%\eightk  \\ 
\midrule
    \multicolumn{2}{@{}l}{Lexical: ROUGE-L} &
        \multicolumn{2}{c}{30.8 \ci{[29.0, 32.8]}} %
        \\
\midrule
    \multicolumn{2}{@{}l}{\evmetric:} & Precision & Recall \\
    & Logical  & 
        56.0 \ci{[53.1, 58.8]} & %
        51.4 \ci{[48.7, 53.8]} %
 \\
    & Spatial & 
        37.1 \ci{[33.4, 40.7]} & %
        23.8 \ci{[21.3, 26.6]} %
    \\
    & Grounding & 
        80.2 \ci{[75.9, 83.5]} & %
        76.6 \ci{[72.4, 80.8]} %
    \\
\midrule
    \multicolumn{2}{@{}l}{Clinical:} & \\
    & RadGraph-F$_1$ &  %
        \multicolumn{2}{c}{18.0 \ci{[16.5, 19.5]}} %
 \\
    & \rger & 
        \multicolumn{2}{c}{23.3 \ci{[21.4, 25.2]}} %
 \\
    & RadCliQ ($\downarrow$) & 
        \multicolumn{2}{c}{2.68 \ci{[2.60, 2.76]}} %
 \\
    & CheXbert \macrofourteen &
        \multicolumn{2}{c}{31.9 \ci{[28.0, 36.7]}} %
 \\
     & CheXbert \microfourteen &
        \multicolumn{2}{c}{60.3 \ci{[57.6, 63.3]}} %
 \\
\midrule
    \multicolumn{2}{@{}l}{Phrase grounding:} & Precision & Recall \\   
    & Box-completion &
    63.5 \ci{[61.8, 65.2]} &%
    66.9 \ci{[65.3, 68.4]} %
    \\
    \bottomrule
\end{tabular}%
}
\label{tab:grounded_reporting_padchest_extended}
\end{table}

\FloatBarrier
\subsection{Phrase grounding on MS-CXR}
\label{app:mscxr}
Because there are no previously published results for grounded reporting, \mairatwo was also evaluated on the related task of phrase grounding, for which public baselines exist. Phrase grounding here means generating a set of bounding boxes given an image and an input phrase, such as `left retrocardiac opacity'.
We compare against MedRPG \citep{chen2023grounding}, ChEX \citep{muller2024chex}, and TransVG \citep{deng2021transvg}.
Compared to MAIRA-2, these baselines directly regress bounding box coordinates using \ac{MLP} heads.
MedRPG additionally employs a combination of contrastive and attention losses to better align image- and text-features. Similarly, the phrase grounding in ChEX benefits from the synergies of multitask training, combining report generation and localisation tasks.

\Cref{tab:phrase_grounding_mscxr} presents the \ac{mIoU} of pixel masks from generated vs ground-truth boxes on our test split of the MS-CXR dataset \citep{boecking2022mscxr}.
Note that \citet{chen2023grounding} and \citet{muller2024chex} used different custom splits of MS-CXR.
To enable fair comparison, we therefore report comparative results on the intersections of our test set with their respective test subsets.
The 95\% confidence intervals for ChEX and TransVG were approximated assuming a normal distribution based on the bootstrapped standard deviation reported by \citet{muller2024chex}.

On the phrase grounding task, \mairatwo achieves competitive performance against baselines developed specifically for phrase grounding (MedRPG and TransVG) and appears to strongly outperform the multi-task ChEX model.

\begin{table}[h!]
\centering
\caption{\textbf{Phrase grounding performance (mIoU) on MS-CXR}. MedRPG \citep{chen2023grounding} reports performance on 20\% of the single-box cases from MS-CXR (approx.\ 178 phrases, 162 images), whereas ChEX \citep{muller2024chex} included only samples in the official MIMIC-CXR validation and test splits (196 phrases, 169 images).
Because the final MAIRA-2 model was trained with a part of MS-CXR, we report results on the intersections of our new held-out test split (176 phrases, 155 images) and each of the splits from MedRPG (138 phrases, 124 images) and ChEX (30 samples, 24 images), respectively.
Results for TransVG \citep{deng2021transvg} are quoted here from the comparisons originally reported for MedRPG and ChEX. 
}
\label{tab:phrase_grounding_mscxr}
\small
\tabspace
\sisetup{table-format=2.2{ \ci{[00.00, 00.00]}}}
\begin{tabular}{@{} lSSccc @{}}
    \toprule
    \textbf{Model}     & {\textbf{Single-box only}}           & {\textbf{In MIMIC-CXR val./test}}  & \textbf{Test split} \\
    \midrule
              & {($n\approx 178$)}          & {($n=196$)}               & -- \\
    MedRPG    & 59.37                       & {--}                      & -- \\
    ChEX      & {--}                        & 46.51~\ci{[44.68, 50.36]} & -- \\
    TransVG   & 58.91                       & 53.51~\ci{[50.51, 56.51]} & -- \\
    \midrule
              & {($n=138$)}                 & {($n=30$)}                & {($n=176$)} \\
    \mairatwo & 57.21~\ci{[53.32, 60.98]}   & 56.86~\ci{[48.28, 64.35]} & 54.68~\ci{[51.26, 58.25]} \\
    \bottomrule
\end{tabular}%
\end{table}

\subsection{Synergy between findings generation and grounded reporting training}
\label{app:synergy}

\mairatwo is a multitask model optimised for both findings generation (\findgen) and grounded report generation (\groundrep). Since \groundrep is based on \findgen, we might expect positive transfer between these tasks. Here we compare the performance of \mairatwo (7B) to models trained \emph{only} on the task of interest, dropping either \findgen and evaluating on \groundrep (\Cref{tab:gr_ab_fg_impact_txt}), or dropping \groundrep (as well as \phraseground, to remove all grounding information during training) and evaluating on \findgen (\Cref{tab:fg_ab_gr_impact_txt}).
Note this analysis was conducted on an earlier variant of \mairatwo as described in \Cref{app:maira_variant}.

\begin{table}[hbt!]
\caption{Impact of dropping the \findgen task during training on \eightk grounded reporting performance. The top table shows text-based metrics, while the bottom table shows box and grounding-based metrics. \commondescription}
\label{tab:gr_ab_fg_impact_txt}
\small
\tabspace
\centerline{
    \begin{tabular}{@{}lcccccc@{}}
    \toprule
     & \textbf{ROUGE-L} & \textbf{CheXbert} & \textbf{\rger} & \textbf{RadCliQ ($\downarrow$)}  & \multicolumn{2}{c}{\textbf{\evmetric Logical}} \\
    \textbf{Experiment} &  & \textbf{\macrofourteen} &  & & \textbf{Precision} & \textbf{Recall} \\
    \midrule
    \mairatwo & 58.2 \ci{[56.7, 59.8]} & 40.9 \ci{[35.9, 47.1]} & 56.9 \ci{[55.3, 58.5]} & 1.63 \ci{[1.55, 1.70]} & 73.5 \ci{[72.2, 74.9]} & 72.4 \ci{[71.0, 73.8]} \\
    \abGROnly & 55.6 \ci{[53.9, 57.0]} & 19.6 \ci{[16.7, 23.4]} & 53.1 \ci{[51.5, 54.7]} & 1.86 \ci{[1.79, 1.93]} & 68.9 \ci{[67.5, 70.4]} & 64.9 \ci{[63.4, 66.4]} \\
    \bottomrule
    \end{tabular}}
    \vspace{0.25cm}
    \centerline{\begin{tabular}{@{}lccccc@{}}
    \toprule
     & \multicolumn{2}{c}{\textbf{\evmetric Grounding}} & \multicolumn{3}{c}{\textbf{Box-completion}} \\
    \textbf{Experiment} & \textbf{Precision} & \textbf{Recall}  & \textbf{Precision} & \textbf{Recall} & \textbf{IoU} \\
    \midrule
    \mairatwo & 68.2 \ci{[64.7, 71.7]} & 92.2 \ci{[89.8, 94.4]} & 68.4 \ci{[67.2, 69.7]} & 84.6 \ci{[83.7, 85.5]} & 60.7 \ci{[59.4, 61.9]} \\
    \abGROnly & 74.3 \ci{[70.2, 78.5]} & 92.5 \ci{[89.6, 95.1]} & 66.3 \ci{[64.9, 67.6]} & 82.7 \ci{[81.8, 83.6]} & 58.4 \ci{[57.1, 59.5]} \\
    \bottomrule
    \end{tabular}
}
\end{table}

\Cref{tab:gr_ab_fg_impact_txt} shows the impact of omitting \findgen task from \mairatwo training in terms of text (top row) and box (bottom row) metrics. We find that dropping \findgen task results in a significant drop in all text metrics, suggesting a positive transfer from \findgen to \groundrep on the quality and clinical factuality of the generated grounded report phrases. In particular, we notice a very large decrease (-52.07\%) in \macrofourteen when dropping \findgen, indicating that \mairatwo grounded reports identify the presence or absence of the 14 CheXpert findings more accurately when the model is trained jointly on \findgen. Additionally, we see a substantial decrease in \evmetric logical precision (-6.25\%) and recall (-10.36\%), indicating that the model trained without \findgen is generating more hallucinations and omissions.
This may also explain the \emph{increase} in \evmetric \emph{grounding} precision (+8.94\%) when we drop \findgen -- the model generates fewer logically entailed sentences, but those which it generates are grounded correctly more often.

\begin{table}[hbt!]
\caption{Impact of dropping the \groundrep task during training on MIMIC \findings generation performance. \commondescription}
\label{tab:fg_ab_gr_impact_txt}
\small
\tabspace
\centerline{
\begin{tabular}{@{}lcccccc@{}}
\toprule
 & \textbf{ROUGE-L} & \textbf{CheXbert} & \textbf{\rger} & \textbf{RadCliQ ($\downarrow$)}  & \multicolumn{2}{c}{\textbf{\evmetric Logical}}  \\
\textbf{Experiment} &  & \textbf{\macrofourteen} & & & \textbf{Precision} & \textbf{Recall}   \\
\midrule
\mairatwo & 38.4 \ci{[37.8, 39.1]} & 42.7 \ci{[40.9, 44.4]} & 51.5 \ci{[49.3, 53.5]} & 39.7 \ci{[38.9, 40.4]} & 2.64 \ci{[2.61, 2.68]} & 50.5 \ci{[49.7, 51.3]} \\
\abFGOnly & 38.3 \ci{[37.7, 38.9]} & 41.8 \ci{[40.2, 43.8]} & 49.9 \ci{[47.7, 51.7]} & 39.6 \ci{[39.0, 40.3]} & 2.65 \ci{[2.61, 2.68]} & 51.2 \ci{[50.4, 52.1]} \\
\bottomrule
\end{tabular}
}
\end{table}

While training on \findgen seems to improve \groundrep performance, we do not observe the reverse: training with \groundrep does not appear to benefit the \findgen task. \Cref{tab:fg_ab_gr_impact_txt} indicates limited impact with most metrics showing overlapping confidence intervals.

\subsection{Further ablations on additional inputs}
\label{app:input_ablations}
In \Cref{fig:_figure4_additional_inputs} we demonstrated the impact of removing additional inputs provided to \mairatwo, either during training, or at inference time alone.
We categorise these inputs along two dimensions: (i)~inputs that are related to the temporal nature of reporting, namely the prior image and report (collectively referred to as the prior study) and the \comparison section; and (ii)~inputs relating to multiple view types collected in a single imaging study, namely the lateral image and \technique section. We do not explore dropping the \indication section here as its importance is already well-established~\citep{nguyen2023pragmatic,hyland2024maira1}.

In this section, we provide further details on these experiments, and provide additional ablations demonstrating the impact of removing each input individually.  Note that as in \Cref{fig:_figure4_additional_inputs,app:synergy}, we conducted these ablations on a slightly earlier version of \mairatwo.

\subsubsection{Description of the `\%Comparison mentions' and `\%Lateral mentions' metrics}
We used a language model (\texttt{Llama3-70B-Instruct}) to detect if a findings-section mentions a comparison to a prior report. We evaluated the prior detection algorithm on 100 samples from the training sets of each \mimiccxr, \padchest, and \privatedata, and found it very robust with 98\%, 96\%, and 97\% accuracy, respectively. The evaluation sets were balanced w.r.t.\ the prevalence of prior mentions.
Since mentions of lateral images in the findings are usually explicit, we resorted to a simple regular expression shown in Listing~\ref{lst:lateral_regex} and refrained from creating evaluation sets for this task.
Applying these algorithms to the generated and reference findings allows us to estimate in how many cases the model should have, and in how many cases it has mentioned a prior report or lateral image. Logical precision or recall values as in \evmetric can not be computed from these numbers, as the detected mentions in the prediction and the reference do not have to be related.

\begin{lstlisting}[caption={Case-insensitive regular expression used to detect mentions of lateral images including explicit (e.g., ``AP and \emph{lateral} views of the chest'') and implicit (e.g., ``Chest \emph{two} views'') mentions of the lateral view.},label={lst:lateral_regex}]
(pa|ap|frontal) and lateral|
\blateral and (pa|ap|frontal)|
\blateral (projection|view)|
(two|2) views
\end{lstlisting}

\begin{table}[hbt!]
\centering
\caption{Prior and comparison ablation experiments on \mimiccxr, on the set of test cases with a prior study (n=2181). \abDropComp\xspace means we drop the comparison section, \abDropPrior\xspace means we drop the prior frontal image and the prior report. \abInf means we drop the inputs only at inference time (we evaluate a model trained using these inputs), otherwise we both train and evaluate without the inputs. This table complements \cref{fig:_figure4_additional_inputs}. \commondescription}
\label{tab:fg_prior_and_comparison_ablation}
\small
\tabspace
\begin{tabular}{@{}lcccccc@{}}
    \toprule
     & \textbf{ROUGE-L} & \textbf{CheXbert} & \textbf{RadCliQ ($\downarrow$)} & \multicolumn{2}{c}{\textbf{\evmetric Logical}} & \textbf{\% Mentions}  \\
    \textbf{Experiment} &  & \textbf{\macrofourteen} &  & \textbf{Precision} & \textbf{Recall} & \textbf{comparison} \\
    \midrule
    \mairatwo & 38.4 \ci{[37.7, 39.0]} & 43.7 \ci{[41.9, 45.6]} & 2.64 \ci{[2.61, 2.68]} & 52.6 \ci{[51.4, 53.6]} & 48.6 \ci{[47.4, 49.7]} & 85.6 \ci{[84.2, 87.0]} \\
    \abInf\abDropComp & 29.8 \ci{[29.2, 30.4]} & 39.9 \ci{[38.0, 41.6]} & 3.07 \ci{[3.03, 3.10]} & 47.9 \ci{[46.7, 49.0]} & 42.6 \ci{[41.7, 43.8]} & 71.2 \ci{[69.2, 73.2]} \\
    \abTrain\abDropComp & 34.9 \ci{[34.3, 35.5]} & 41.9 \ci{[39.9, 43.7]} & 2.81 \ci{[2.78, 2.85]} & 52.7 \ci{[51.6, 53.7]} & 46.4 \ci{[45.3, 47.4]} & 86.4 \ci{[84.9, 87.9]} \\
    \abInf\abDropPrior & 37.9 \ci{[37.3, 38.6]} & 39.2 \ci{[37.6, 41.2]} & 2.69 \ci{[2.66, 2.73]} & 51.5 \ci{[50.5, 52.5]} & 47.1 \ci{[46.2, 48.2]} & 72.9 \ci{[71.1, 74.8]} \\
    \abTrain\abDropPrior & 38.2 \ci{[37.5, 38.9]} & 40.0 \ci{[38.2, 41.8]} & 2.67 \ci{[2.63, 2.71]} & 52.5 \ci{[51.6, 53.6]} & 47.4 \ci{[46.4, 48.4]} & 82.8 \ci{[81.2, 84.3]} \\
    \abInf\abDropPriorComp & 27.3 \ci{[26.7, 28.0]} & 35.8 \ci{[34.2, 37.5]} & 3.18 \ci{[3.15, 3.22]} & 45.5 \ci{[44.4, 46.5]} & 40.5 \ci{[39.6, 41.4]} & 38.6 \ci{[36.7, 40.5]} \\
    \abTrain\abDropPriorComp & 33.9 \ci{[33.2, 34.5]} & 39.3 \ci{[37.5, 41.1]} & 2.89 \ci{[2.86, 2.93]} & 50.6 \ci{[49.5, 51.5]} & 44.7 \ci{[43.7, 45.7]} & 75.8 \ci{[73.9, 77.4]} \\
    \bottomrule
\end{tabular}
\end{table}

\subsubsection{Inputs containing temporal information}
In \Cref{tab:fg_prior_and_comparison_ablation}, we show training and inference-time ablations demonstrating the independent effect of including the prior study and comparison section. As in \cref{fig:_figure4_additional_inputs}, this analysis is performed on the subset of the MIMIC test set that has prior images. When we train without the prior study `\abTrain\abDropPrior', we observe a significant drop in \macrofourteen (-8.5\%). We also see a similar but larger drop in \macrofourteen (-10.3\%) when we train with the prior study but drop it during inference `\abInf\abDropPrior', indicating that \mairatwo uses the prior study to produce more factually correct reports. We also note that using a model trained with prior studies and running inference without prior studies will cause fewer hallucinations of comparisons (72.9\%) as compared to a model that was not trained with prior studies (82.8\%). When we train a model without the comparison section `\abTrain\abDropComp', we observe a significant drop in lexical metrics (-9.1\% drop in ROUGE-L) as well as an increase in RadCliQ (+6.4), but no significant drop in \macrofourteen. When we train with the comparison section but drop it at inference time `\abInf \abDropComp', we note an even larger drop in ROUGE-L (-22.4) in addition to an overall decrease in performance across all other metrics. Based on the large drop in lexical metrics when not using the comparison section, and the reduction in hallucinations when we train a model with comparison sections and run inference without them `\abInf\abDropPriorComp' as compared to training without these sections entirely `\abDropPriorComp', we hypothesise that the model uses the comparison section as an indicator of whether or not temporal change mentions should be generated in the text, and that the prior image is necessary to ensure the change words generated are correct.

\begin{table}[hbt!]
\centering
\caption{Lateral and technique ablations on \mimiccxr for the subset of the test set with a lateral view (n = 1,116). \abDropLat\xspace means we drop the lateral view, \abDropTech\xspace means we drop the \technique section. `Inf' means we drop the inputs only at inference time, evaluating a model trained using those inputs. Otherwise, we both train and evaluate without the inputs. This table complements \cref{fig:_figure4_additional_inputs}. \commondescription}
\label{tab:fg_lateral_and_technique_ablation}
\small
\tabspace
\begin{tabular}{@{}lcccccc@{}}
    \toprule
     & \textbf{ROUGE-L} & \textbf{CheXbert} & \textbf{RadCliQ ($\downarrow$)} & \multicolumn{2}{c}{\textbf{\evmetric Logical}} & \textbf{\% Mentions} \\
     \textbf{Experiment} & 
     & %
     \textbf{\macrofourteen} & %
     & %
     \textbf{Precision} & %
     \textbf{Recall}  & %
     \textbf{lateral} \\% mentions\\
    \midrule
    \mairatwo & 40.4 \ci{[39.5, 41.4]} & 38.8 \ci{[35.9, 41.5]} & 2.50 \ci{[2.44, 2.56]} & 60.2 \ci{[58.7, 61.7]} & 54.7 \ci{[53.2, 56.0]} & 39.6 \ci{[36.6, 42.5]} \\
    \abInf\abDropLat & 38.9 \ci{[38.0, 39.8]} & 36.8 \ci{[34.5, 39.6]} & 2.54 \ci{[2.49, 2.60]} & 60.5 \ci{[59.0, 61.9]} & 53.2 \ci{[51.7, 54.6]} & 13.2 \ci{[11.3, 15.3]} \\
    \abTrain\abDropLat & 40.4 \ci{[39.4, 41.4]} & 39.1 \ci{[36.5, 42.5]} & 2.51 \ci{[2.46, 2.56]} & 60.4 \ci{[58.9, 62.0]} & 54.1 \ci{[52.8, 55.6]} & 38.2 \ci{[35.4, 41.3]} \\
    \abInf\abDropTech & 34.1 \ci{[33.2, 34.9]} & 36.6 \ci{[33.7, 39.5]} & 2.81 \ci{[2.76, 2.86]} & 56.6 \ci{[55.2, 58.2]} & 51.1 \ci{[49.8, 52.3]} & 61.2 \ci{[58.4, 64.0]} \\
    \abTrain\abDropTech & 37.2 \ci{[36.3, 38.2]} & 36.1 \ci{[33.3, 38.6]} & 2.64 \ci{[2.59, 2.69]} & 58.4 \ci{[57.0, 59.7]} & 53.1 \ci{[51.8, 54.4]} & 36.3 \ci{[33.2, 39.4]} \\
    \abInf\abDropLatTech & 31.5 \ci{[30.8, 32.3]} & 35.2 \ci{[32.8, 37.7]} & 2.87 \ci{[2.82, 2.92]} & 57.7 \ci{[56.3, 59.0]} & 49.1 \ci{[47.8, 50.7]} & 5.1 \ci{[3.8, 6.4]} \\
    \abTrain\abDropLatTech & 36.8 \ci{[36.0, 37.9]} & 39.0 \ci{[36.2, 41.9]} & 2.66 \ci{[2.61, 2.71]} & 58.5 \ci{[57.2, 59.9]} & 53.7 \ci{[52.3, 55.0]} & 36.1 \ci{[33.0, 39.2]} \\
    \bottomrule
\end{tabular}
\end{table}

\subsubsection{Inputs related to multi-view studies} \Cref{tab:fg_lateral_and_technique_ablation} shows training and inference-time ablations to evaluate the impact of including the lateral view and the technique section independently, as a complement to the analysis in \cref{fig:_figure4_additional_inputs} demonstrating their joint effect. Similarly, we restrict the analysis to the test studies that include a lateral view (n=1,116, 30.6\%). 
When we drop the lateral view at inference-time `\abInf\abDropLat', we notice that \mairatwo generates less lateral mentions (13.23\% vs 39.57\%) and therefore limited ``lateral hallucinations''.
We also observe a drop in almost all metrics including \macrofourteen (-5.15\%) highlighting the importance of the lateral view in making accurate diagnosis. On the other hand, a model trained without the lateral view continues to hallucinate lateral mentions (38.16\%) since it can use the technique section as a proxy to make simple lateral predictions.
Even though this ablated model is able to generate simple lateral references using the technique section as a shortcut, there is no guarantee that it has improved it's clinical accuracy when a pathology can only be seen on the lateral. 
Moreover, when we drop the technique in the presence of the lateral view `\abInf\abDropTech', we see a large drop in ROUGE-L (-15.59\%) and a substantial increase of the \lateralmentions, exceeding 35.57\% (percentage of lateral mentions in the ground truth) by a very large margin. This suggests that the technique section is a strong indicator for generating lateral mentions. However, when this information is omitted during training `\abTrain\abDropTech', the model can still figure out when to mention the lateral view (36.33\%) but not as accurately as in \mairatwo. Finally, when we drop both the lateral and the technique at the same time during inference `\abInf\abDropLatTech', the percentage of lateral mentions drops down to 5.1\% (getting closer to 0) indicating that both the lateral view and the technique section are essential to reduce hallucinations related to lateral mentions. 
This further becomes clearer when compared to a model that is trained without this information `\abTrain\abDropLatTech' but still hallucinates lateral mentions (36.12\%) as discussed in Results.

\clearpage
\section{Additional qualitative examples}
\label{app:qualitative}

\subsection{Successful grounded reporting examples from \eightk}
 We showcase additional sample generations from \mairatwo with comments from radiologist review in  \Cref{fig:grounded_reporting_main_example_flipped,fig:grounded_reporting_main_example_2_flipped,fig:grounded_reporting_main_example_3_flipped,fig:grounded_reporting_main_example_4_flipped}.

\afterpage{
\thispagestyle{empty}
\begin{figure*}
    \centering
    \vspace{-1cm}\centerline{\includegraphics[scale=0.40]{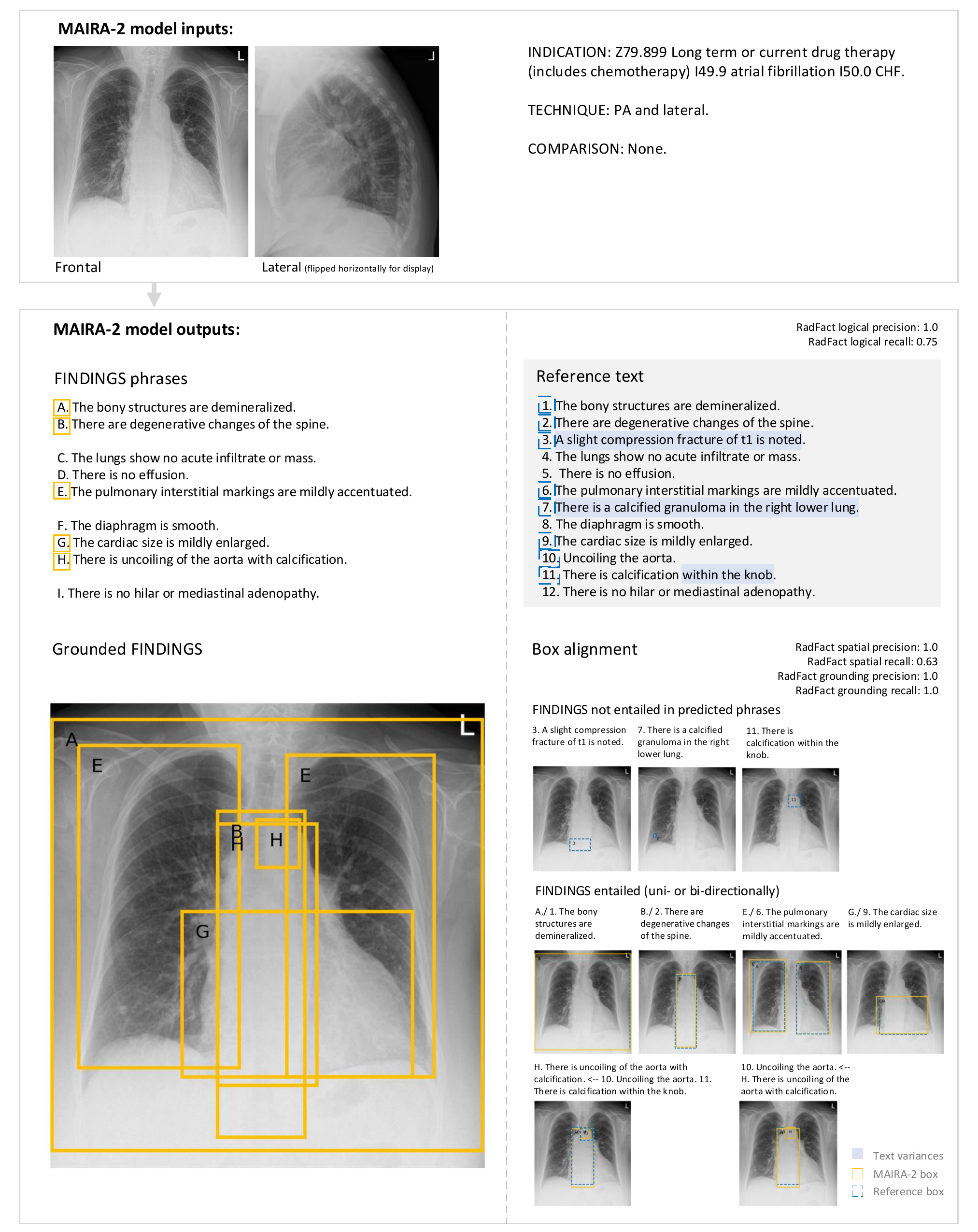}}
\caption{\textbf{A manually-selected qualitative example of MAIRA-2 output on \eightk}. This 3-part figure shows MAIRA-2 model inputs (top); the MAIRA-2 phrase outputs vis-à-vis the reference text (middle); and grounding boxes for the MAIRA-2 phrases on the current frontal image alongside their alignment with reference boxes (bottom). In this example, all generated MAIRA-2 phrases were evidenced by the reference text (\evmetric logical precision: 1.0). In a radiologist review, we find two missed findings: ``There is a calcified granuloma in the right lower lung.'' and ``A slight compression fracture of t1 is noted.'', which can only be seen on the lateral view. \evmetric further counts finding 11 as missed, bringing logical recall to 0.75. Reviewing the reference findings, radiologists pointed out that the compression fracture is on L1 vertebra in the image, suggesting a potential typo in the reference text. Concerns were also raised that small fracture cases may not always be reported and could be missed in training data. Although the compression fracture was not detected, MAIRA-2 correctly outputs the ``degenerative changes of the spine'' that are always better seen on the lateral view. %
For image grounding, no boxes were generated for missed findings 2 and 7. While finding 11 (``There is calcification within the knob'') was also not logically entailed according to \evmetric, the model did correctly generate a separate box around the aortic knob when grounding finding H (``There is uncoiling of the aorta with calcification'').}
    \label{fig:grounded_reporting_main_example_flipped}
\end{figure*}
\clearpage
}

\afterpage{
\thispagestyle{empty}
\begin{figure}
    \centering
    \centerline{\includegraphics[scale=0.4]{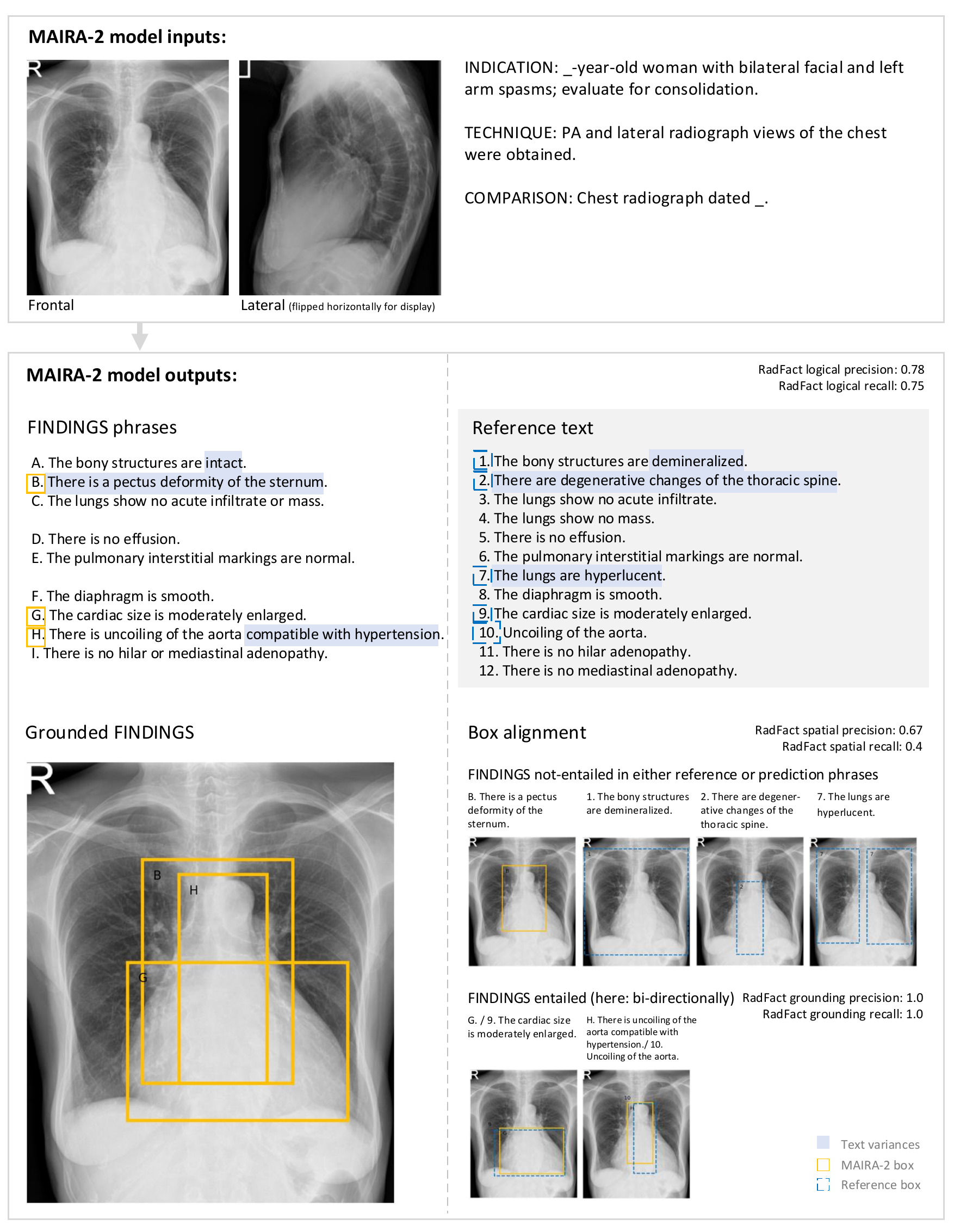}}
\caption{\textbf{A manually selected qualitative example of \mairatwo output on \eightk.} This 3-part figure shows \mairatwo model inputs (top); the \mairatwo phrase outputs vis-à-vis the reference text (middle); and grounding boxes for the \mairatwo phrases on the current frontal image alongside their alignment with reference boxes (bottom). The selected example has moderate \mairatwo \evmetric logical precision (0.78) and recall (0.75). Qualitative comparison with the reference text suggests that \mairatwo misclassified the patient's bony structures as ``intact''; added that the uncoiling of the aorta is ``compatible with hypertension''; and missed detecting the ``degenerative changes of the thoracic spine'' and that the ``lungs are hyperlucent''. In individual reviews with two consultant radiologists, it was suggested that the demineralisation of the bony structures is difficult to see on the images and therefore considered a borderline finding to call out. Similarly, the degenerative changes of the spine were assessed as only mild. Furthermore, the addition of hypertension was regarded as `acceptable' since the aorta is slightly torturous. Lastly, it was noted how the \mairatwo findings also included that ``There is a pectus deformity of the sternum'', which was not reported in the reference and can only be clearly seen on the lateral view. %
For image grounding, there was no overlap between four abnormal findings that were reported in either the \mairatwo candidate or the reference text, resulting in non-corresponding bounding box as is reflected in lower spatial precision (0.67) and recall (0.4) scores. For the two abnormal findings that were reported and entailed in both findings texts, however, there is high grounding precision and recall (1.0).}
    \label{fig:grounded_reporting_main_example_2_flipped}
\end{figure}
\clearpage
}

\afterpage{
\thispagestyle{empty}
\begin{figure}
    \centering
    \centerline{\includegraphics[scale=0.4]{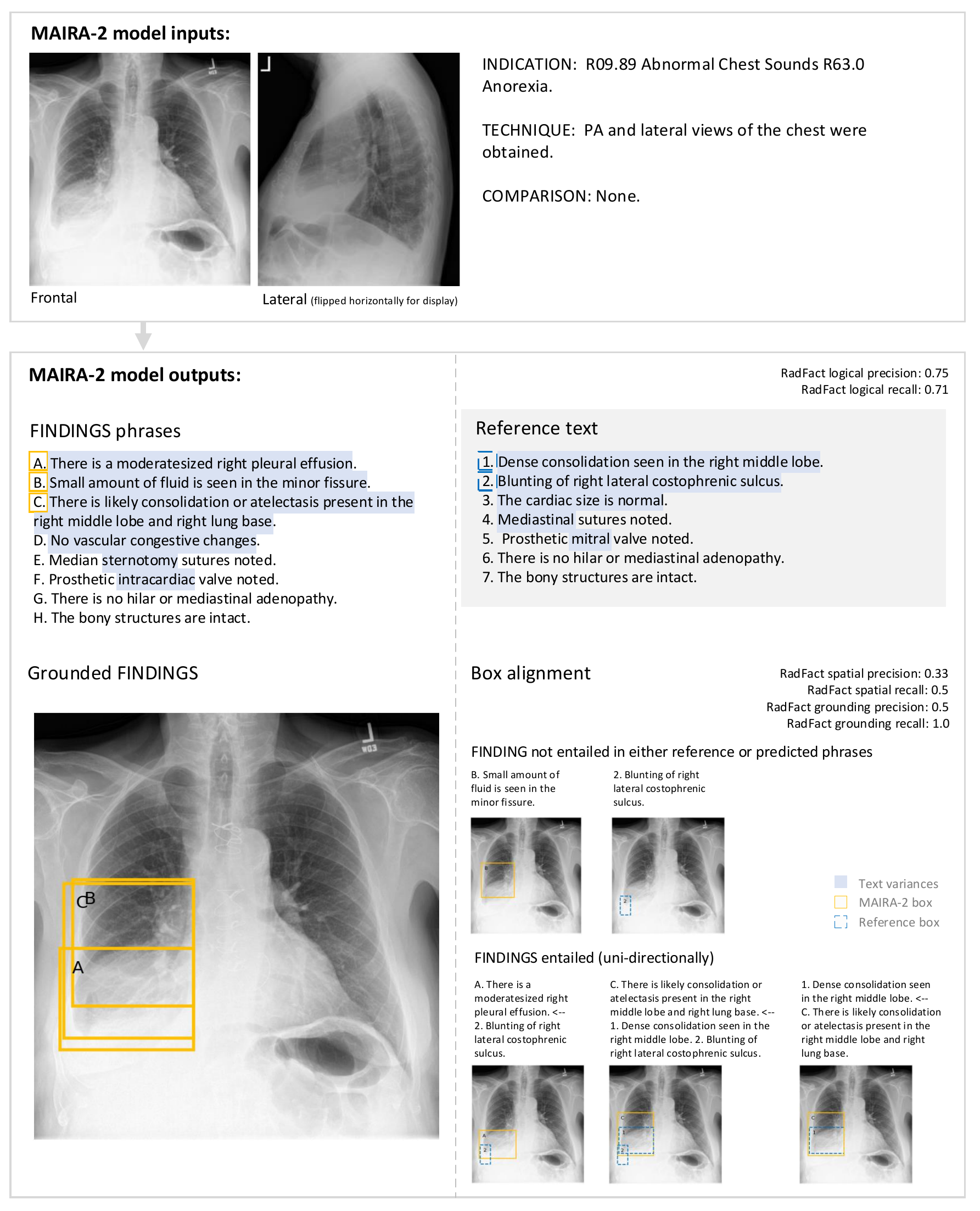}}
\caption{\textbf{A manually selected qualitative example of \mairatwo output on \eightk.} This 3-part figure shows \mairatwo model inputs (top); the \mairatwo phrase outputs vis-à-vis the reference text (middle); and grounding boxes for the \mairatwo phrases on the current frontal image alongside their alignment with reference boxes (bottom). The selected example has moderate \evmetric logical precision (0.75) and recall (0.71). In this example study, \mairatwo model outputs state moderate right pleural effusion, small amount of fluid in the minor fissure; as well as the presence of consolidation or atelectasis in the right middle lobe and right lung base. In review with a consultant radiologist, they agreed with these findings, however, they found that the corresponding \mairatwo bounding boxes for findings B and C were too big. For example, a small amount of fluid in the minor fissure is only visible as a small single line in the middle of the much larger box for finding C. As such, this study presents an example of good logical precision, however, with lower spatial performance. Both the reference text and the \mairatwo outputs also state different normals (e.g., normal cardiac size, no vascular congestive changes). Furthermore, reviewing finding E and reference finding 4, the consultant radiologist preferred the \mairatwo phrase of  ``Median sternotomy sutures noted.'', since it is more accurate in its indication of the sutures: sternotomy rather than the mediastinal. Regarding finding 5, the term ``mitral'' simply presents a type of ``intracardiac valve'', and therefore finding F was considered acceptable.}
    \label{fig:grounded_reporting_main_example_3_flipped}
\end{figure}
\clearpage
}

\afterpage{
\thispagestyle{empty}
\begin{figure}
    \centering
    \centerline{\includegraphics[scale=0.4]{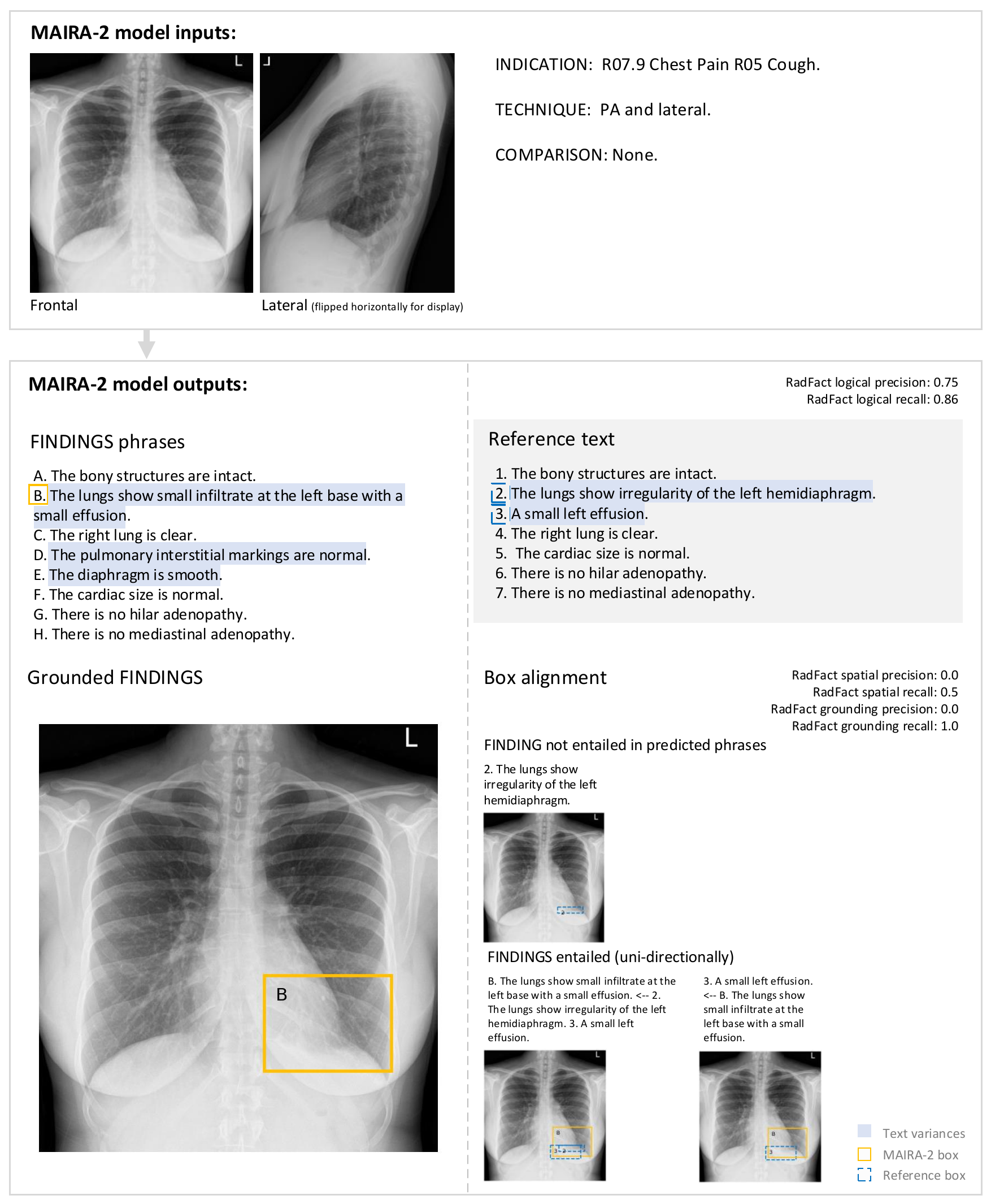}}
\caption{\textbf{A manually selected qualitative example of \mairatwo output on \eightk.} This 3-part figure shows \mairatwo model inputs (top); the \mairatwo phrase outputs vis-à-vis the reference text (middle); and grounding boxes for the \mairatwo phrases on the current frontal image alongside their alignment with reference boxes (bottom). The selected example has moderate \mairatwo \evmetric logical precision (0.75) and recall (0.86). Both the reference text and \mairatwo phrase output suggest the existence of a small left effusion, which can be clearly seen on the lateral view. On the frontal image, the irregularity of the diaphragm suggests that there is small infiltrate at the left base. The identified infiltrate and effusion are considered to explain well the symptoms of chest pain and cough that are given in the indication; and the grounding box for finding B is evaluated to be appropriate for the finding. Nonetheless, \mairatwo findings erroneously state that the diaphragm is smooth, when it has irregularities. Whilst not mentioned in the reference text, \mairatwo outputs also include ``The pulmonary interstitial markings are normal.'', which is correct. In this instance, the reference boxes for findings 2 and 3, which were drawn by human annotators, are very small. Consequently, even though there was good logical entailment for the key abnormal findings, their corresponding boxes did not overlap enough (given the set 50\% threshold), explaining the low grounding precision scores.%
}
    \label{fig:grounded_reporting_main_example_4_flipped}
\end{figure}
\clearpage
}

\subsection{High and low-scoring examples from \eightk according to \evmetric}
 \Cref{fig:grounded_reporting_example_phrase_precision_top,fig:grounded_reporting_example_phrase_precision_middle,fig:grounded_reporting_example_phrase_precision_bottom} present manually selected qualitative example of \mairatwo output on \eightk with varying \evmetric logical precision: 1.0, 0.78 and 0.0 respectively.  \Cref{fig:grounded_reporting_example_box_entail_precision_top,fig:grounded_reporting_box_entail_precision_middle,fig:grounded_reporting_box_entail_precision_bottom_1} present additional examples selected based on varying \evmetric grounding precision: 1.0, 0.5 and 0.0 respectively.

\afterpage{
\begin{figure}
    \centerline{\includegraphics[scale=0.5]{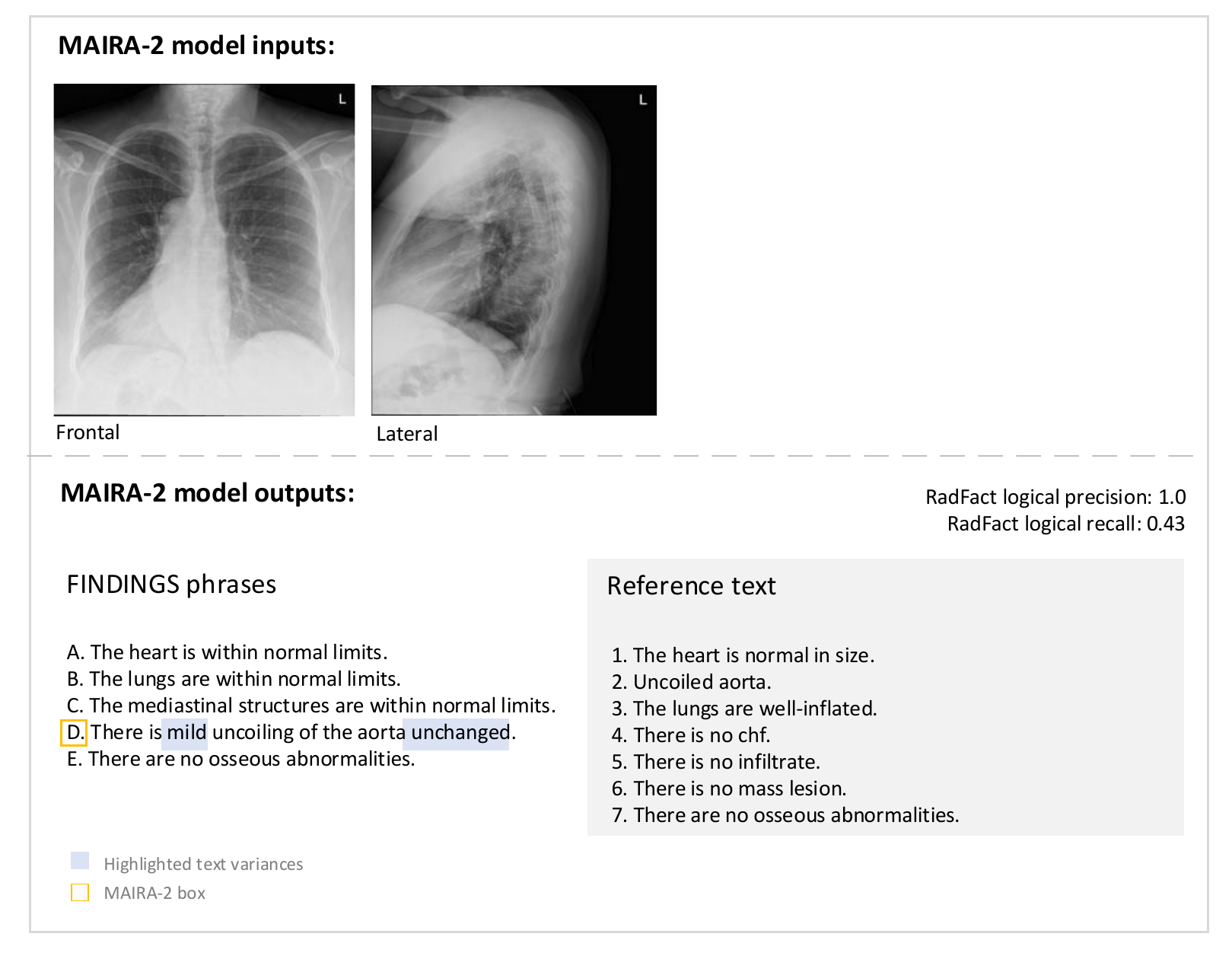}}
\caption{This example has high \evmetric logical precision (1.0) and presents an interesting case since the patient has ``situs inversus'', meaning all organs are mirrored in the body (e.g., the heart, aorta and stomach are on the right rather than the left side). Neither the reference text nor \mairatwo outputs state ``situs inversus'', an observation that is, of course, within normal limits. The study is mostly normal and findings well entailed. However, while the study information state no comparison, the \mairatwo output hallucinated ``unchanged'' about the uncoiled aorta. The reference text, relating to a normal study, also does not have any box annotations, meaning that the uncoiled aorta is only grounded within the \mairatwo findings. 
}
    \label{fig:grounded_reporting_example_phrase_precision_top}
\end{figure}
\clearpage
}

\afterpage{
\begin{figure}
    \centerline{\includegraphics[scale=0.5]{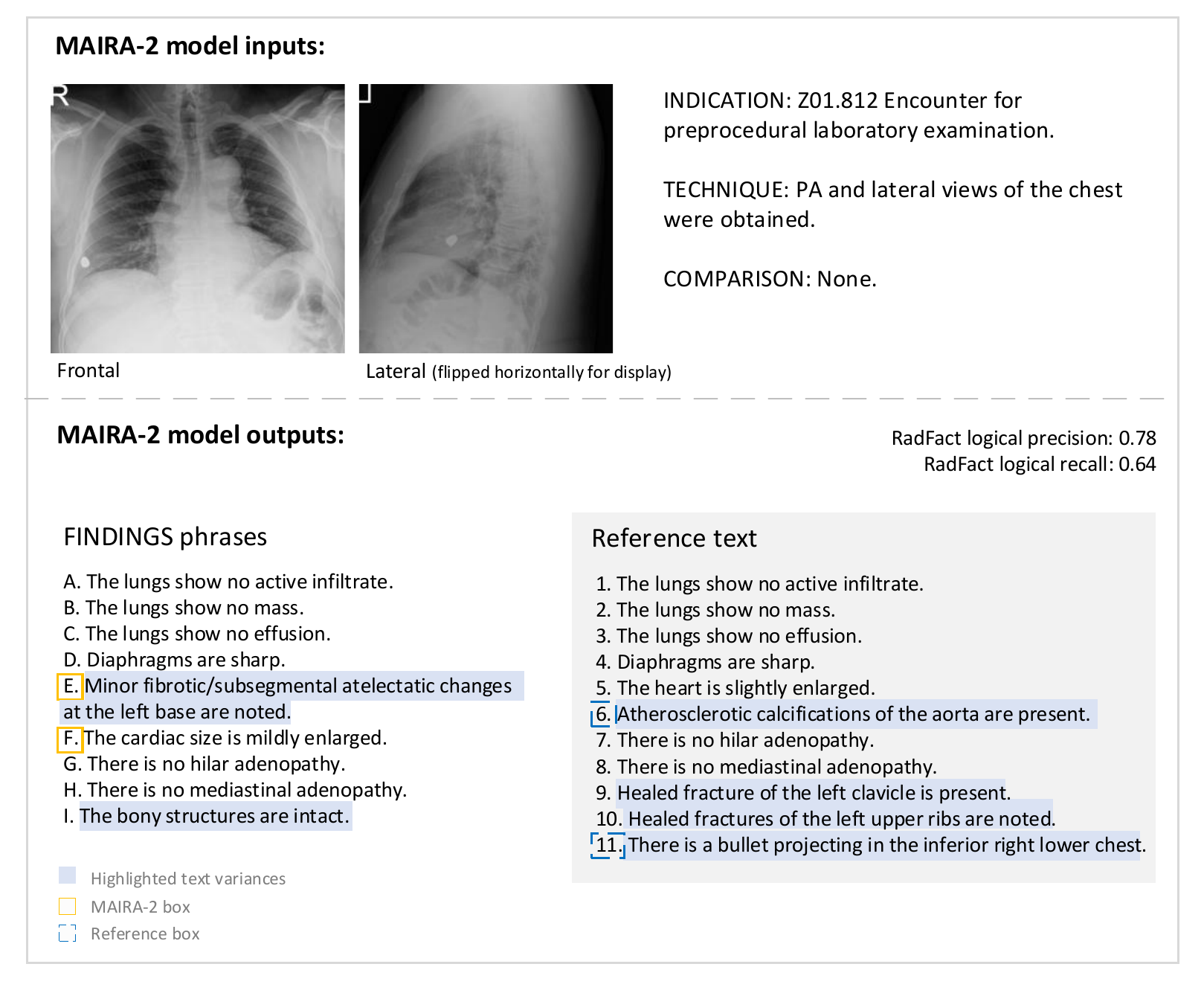}}
\caption{This example has moderate \evmetric logic precision (0.78) and recall (0.64). Many of the phrases are well-matched. \mairatwo output however missed the ``bullet'' that is projecting in the right lower chest, and it does not include the calcification of the aorta, which was described in review with a consultant radiologist as a very difficult to see finding and therefore a borderline observation. Where the reference states healed fractures, \mairatwo outputted that the bony structures are intact. \mairatwo outputs further include ``Minor fibrotic/subsegmental atelectatic changes at the left base are noted''; which is evidenced by the elevated left hemidiaphragm pushing in the lung with resulting atelectasis -- a finding that was not reported in the reference text. 
}
    \label{fig:grounded_reporting_example_phrase_precision_middle}
\end{figure}
\clearpage
}

\afterpage{
\begin{figure}
    \centerline{\includegraphics[scale=0.5]{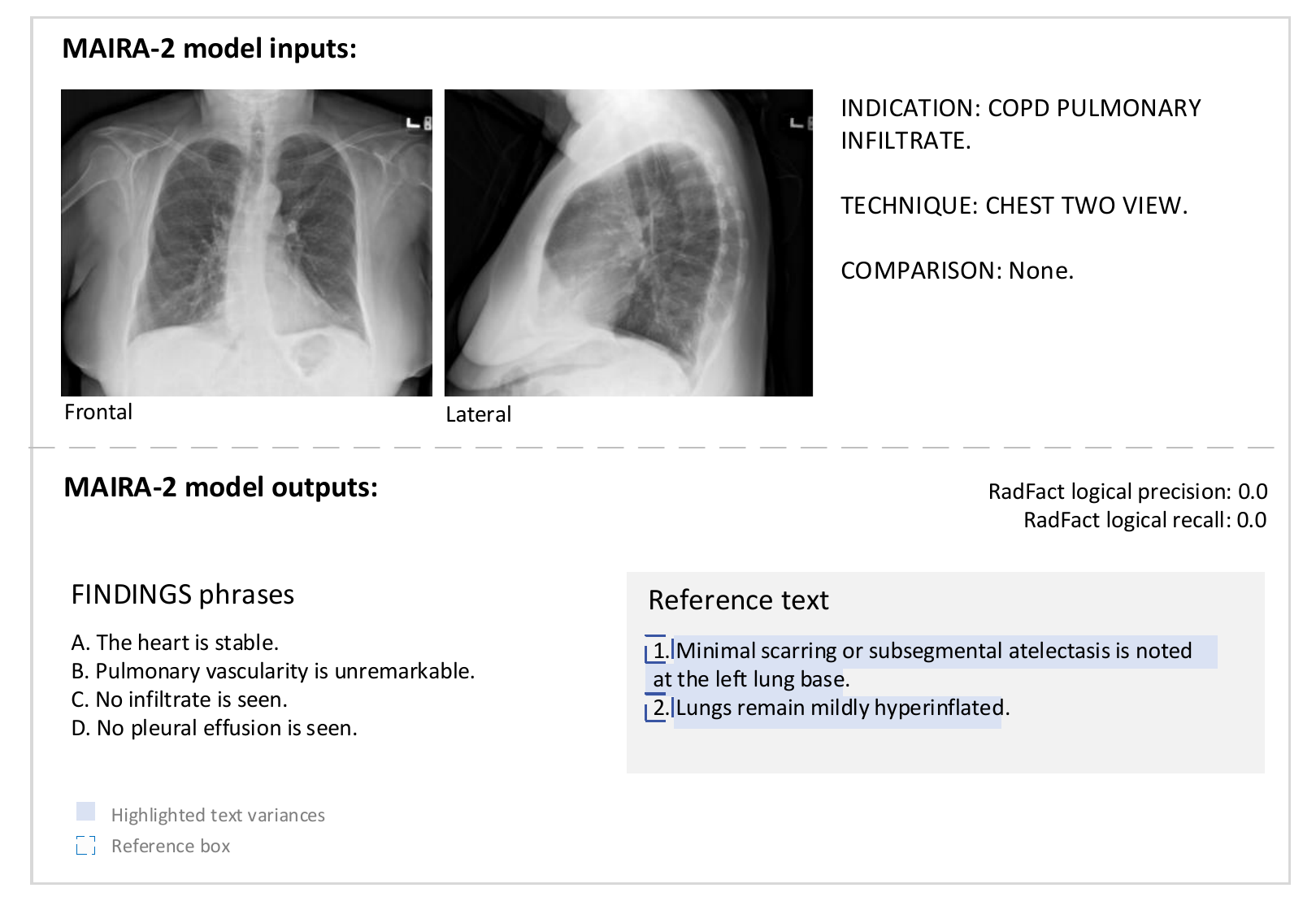}}
\caption{This example has low \evmetric logical precision and recall (0.0). \mairatwo phrases suggest this is a normal study, whilst the reference states ``Minimal scarring or subsegmental atelectasis is noted at the left lung base.'' and ``Lungs remain mildly hyperinflated.''. Both present minimal or mild findings that were however missed. In reviews with a consultant radiologists it was pointed out that the study indication states COPD, which – where it is a known condition – would mean hyperinflation is to be expected. Furthermore, the review surfaced that both text candidates missed the ``scoliosis'' -- a sideways curvature of the spine -- that is visible in the frontal image.   
}
    \label{fig:grounded_reporting_example_phrase_precision_bottom}
\end{figure}
\clearpage
}

\afterpage{
\thispagestyle{empty}
\begin{figure}
    \centerline{\includegraphics[scale=0.5]{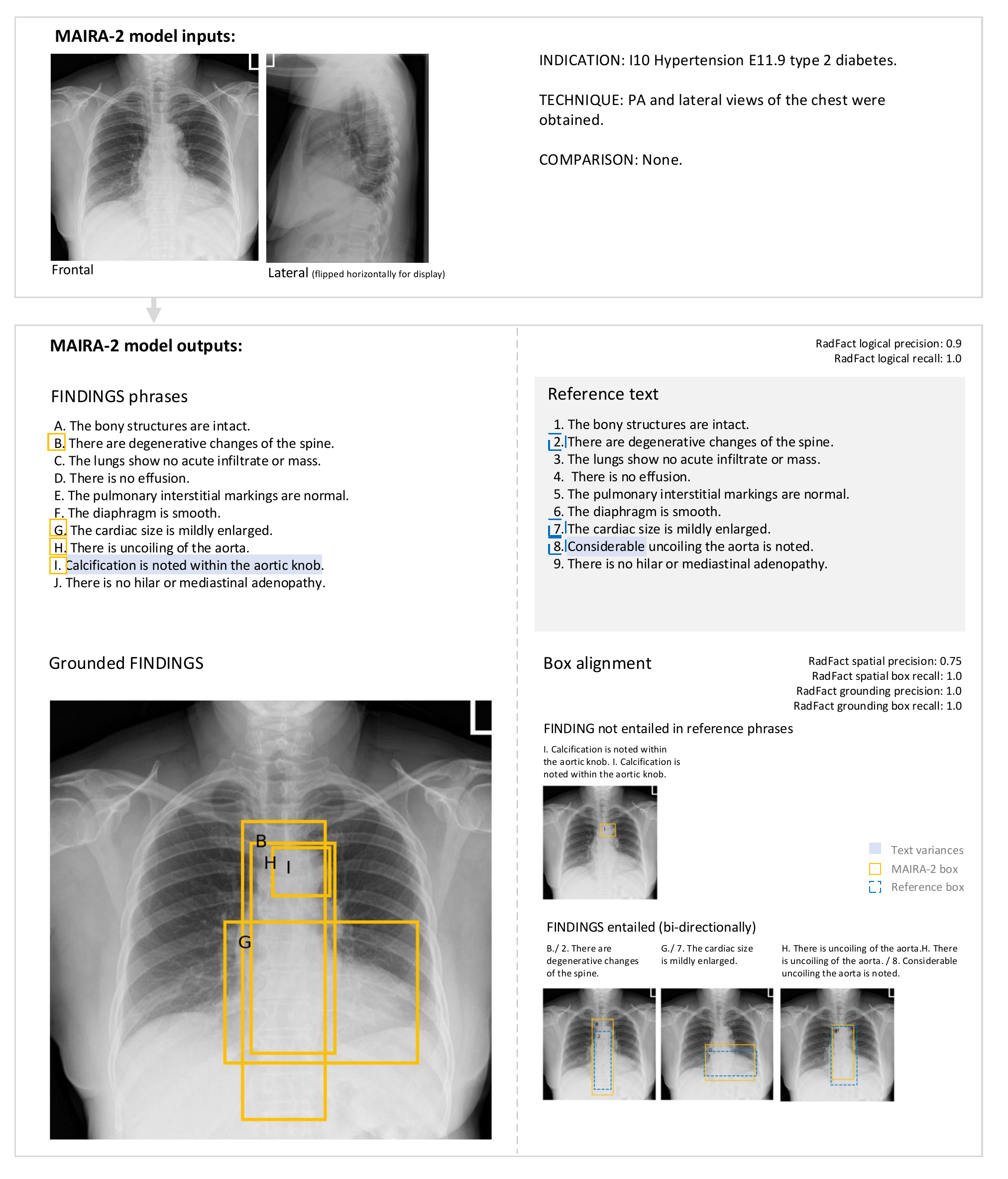}}
\caption{This example has high grounding precision (1.0). There is generally high overlap between both findings texts. \mairatwo output includes the finding of a ``Calcification is noted within the aortic knob.'', which is described in radiologist review as a plausible, borderline findings that was however not included in the reference text. Where generated \mairatwo findings and boxes are matching the reference, resulting grounding precision is high.}
    \label{fig:grounded_reporting_example_box_entail_precision_top}
\end{figure}
\clearpage
}

\afterpage{
\thispagestyle{empty}
\begin{figure}
    \centerline{
     \includegraphics[scale=0.4]{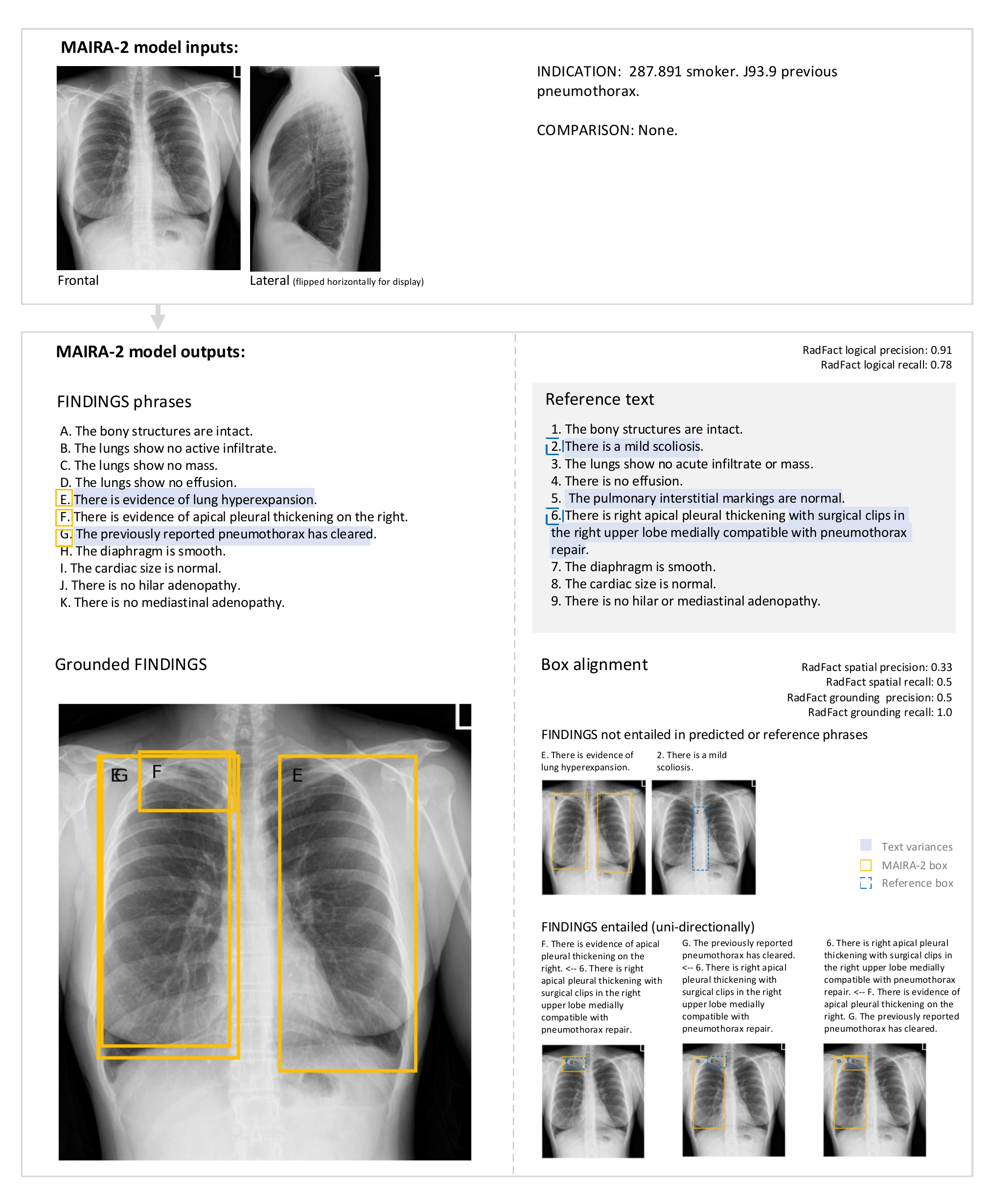}}
\caption{This example has moderate grounding precision (0.5). The \mairatwo outputs include a finding stating ``There is evidence of lung hyperexpansion'', which in radiologist review was verified to be correct. Both findings texts correctly identified the apical pleural thickening on the right upper lobe. However, the reference text expands this finding to also include commentary about surgical clips and pneumothorax repair, whereas the \mairatwo model outputs a separate phrase stating the ``previously noted pneumothorax has cleared''. Although there is good alignment with the reference boxes where the findings specify the apical pleural thickening (findings F and 6), \mairatwo falsely generated a whole right lung box for a cleared pneumothorax (finding G), which would not match to the much narrower reference bounding box that is centered on the pleural thickening; thereby explaining the lower entailed box performance metrics. Nonetheless, it is interesting to point out that change information about the pneumothorax was generated even though no prior image was available to this study, likely as a consequence of the study indication that states ``previous pneumothorax''.}
    \label{fig:grounded_reporting_box_entail_precision_middle}
\end{figure}
\clearpage
}

\afterpage{
\thispagestyle{empty}
\begin{figure}
    \centerline{
     \includegraphics[scale=0.5]{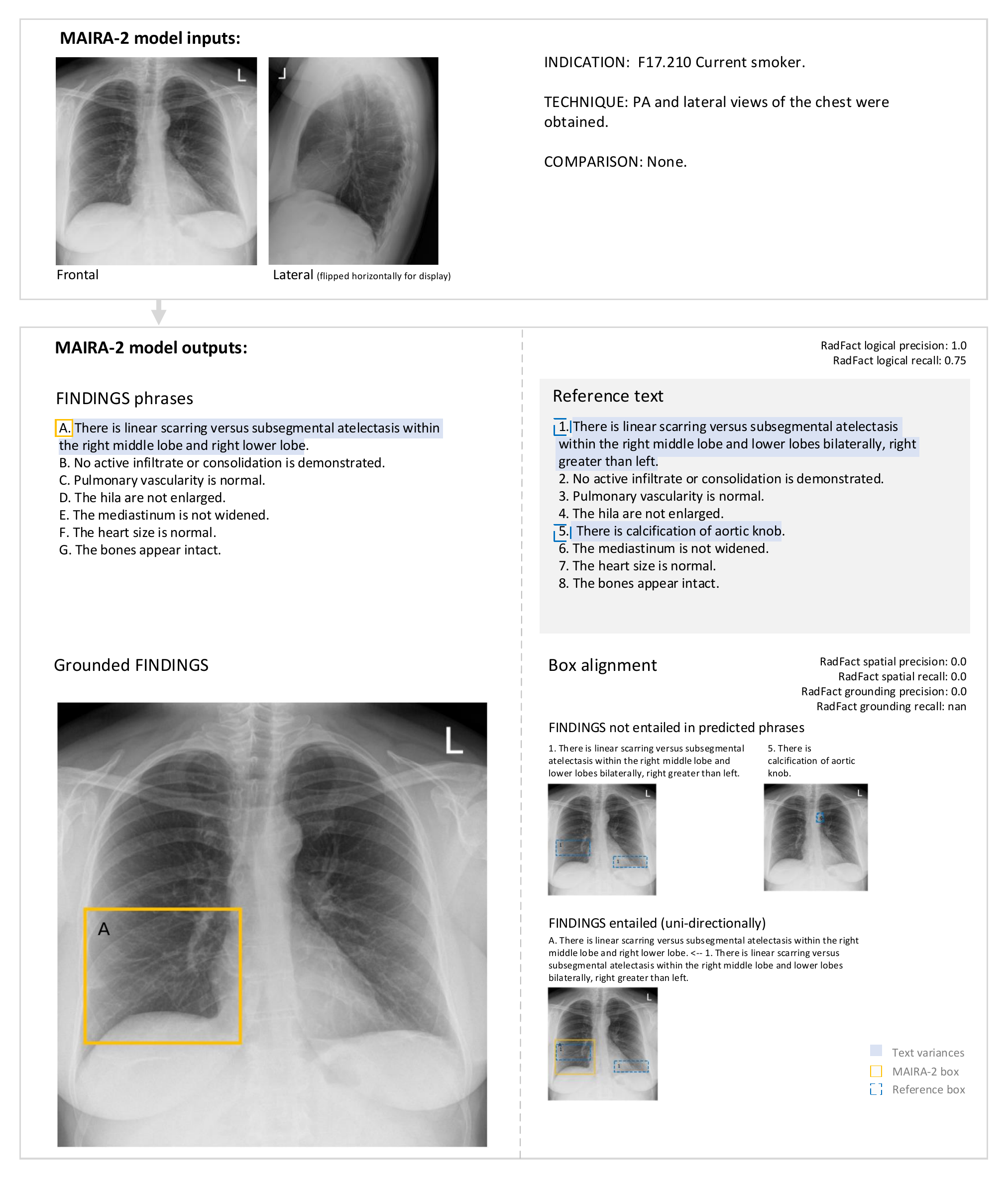}}
\caption{This example has low grounding precision (0.0). The reference text states that the finding of ``linear scarring versus subsegmental atelectasis'' exists bilaterally, whereas the \mairatwo outputs describe these as right-sided only. In radiologist review, the left lower lobe omission was indeed considered a missed subtle finding. \mairatwo outputs also did not include the report of the ``calcification of aortic knob''. In terms of bounding box placement and size, the \mairatwo box for finding A was considered a bit too big, but acceptable; whereas the corresponding bounding boxes for the reference text (Finding 1) were described as a bit too narrow; especially on the right side. This low overlap in bounding boxes explains the lower box precision scores in this instance. }
    \label{fig:grounded_reporting_box_entail_precision_bottom_1}
\end{figure}
\clearpage
}

\subsection{Findings generation examples from MIMIC-CXR}

\label{app:qualitative_mimic}

\begin{figure}
       \centerline{\includegraphics[scale=0.5]{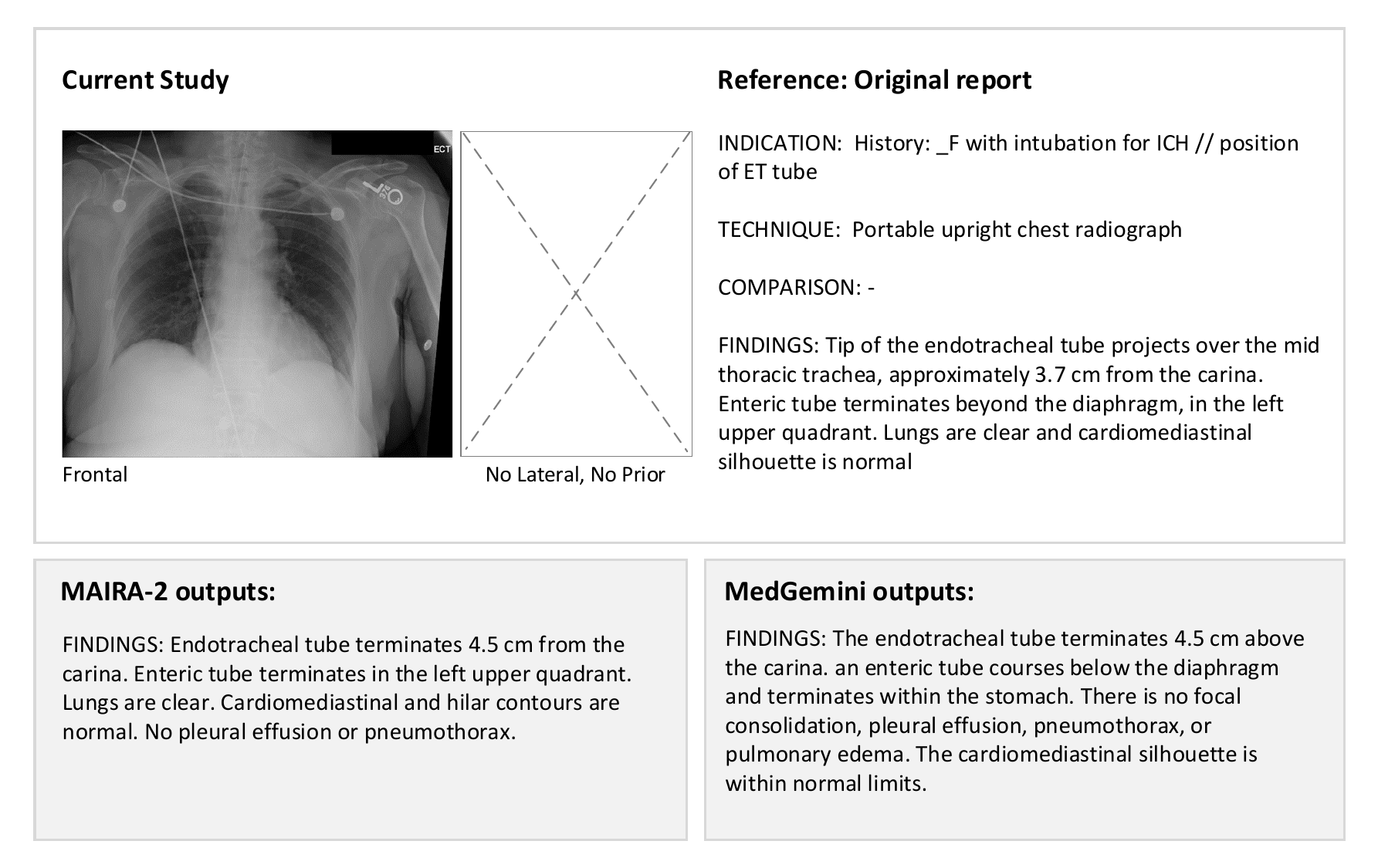}}
    \caption{One of the examples from \cite{yang2024advancing} to enable qualitative comparison to \medgemini. Apart from the specific lines and tubes findings, this study is mostly normal. Neither generated report findings missed any observations and both were assessed in radiologist reviews as equivalent from a clinical perspective. Interestingly, both candidate texts hallucinated the 4.5 cm measure of the endotracheal tube terminating above the carina, as neither model can plausibly predict the correct measurement from a chest X-ray as they have no information on scale. %
While the \mairatwo output produces the exact same location description for the enteric tube as the reference text stating it to terminate ``in the left upper quadrant''; the \medgemini location of ``terminates within the stomach'' is considered as more precise. On the other hand, preferences were expressed for \mairatwo stating ``Lungs are clear'', which is more concise than its counterpart; and for \mairatwo’s ordering of the findings from lungs to the cardiomedistinal/ hilar structures and then the pleura (similar to a structured report generation), compared to the \medgemini findings that move between those structures. 
}
    \label{fig:MedGemini_example_1}
\end{figure}

\begin{figure}
       \centerline{\includegraphics[scale=0.5]{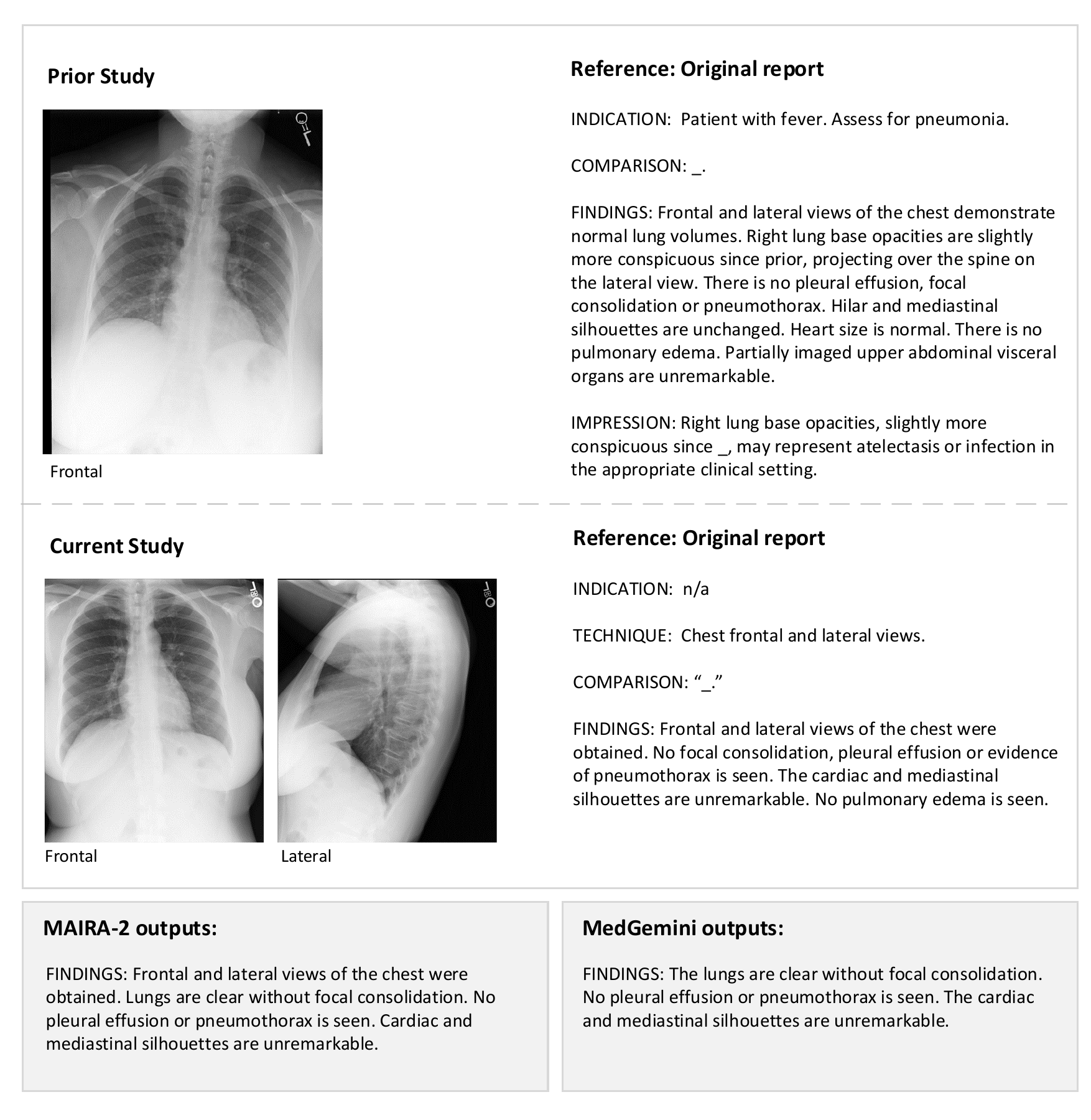}}
    \caption{One of the examples from \cite{yang2024advancing} to enable qualitative comparison to \medgemini. This example study reports a ``normal''. Both generated candidates are near identical and match the reference text findings. Like the reference text, \mairatwo outputs the phrase ``Frontal and lateral views of the chest were obtained.'' Whilst learned from such input instances, technically, this information does not present an image finding and it is already included in the \technique description. Neither the reference text, nor \mairatwo and \medgemini phrases include any comparison information with the prior study. }
    \label{fig:MedGemini_example_2}
\end{figure}

It is not possible to quantitatively compare to models trained to generate other sections, such as \reportsection{Impression}~\citep{bannur2023biovilt} or both \reportsection{Findings} and \reportsection{Impression} together~\citep{tanno2023flamingocxr,yang2024advancing}. In \Cref{fig:MedGemini_example_1,fig:MedGemini_example_2,fig:MedGemini_example_3,fig:MedGemini_example_4}, we qualitatively compare on the examples shown in \cite{yang2024advancing}, which were sourced from the MIMIC-CXR validation set. We find that all four study examples represent mostly ``normal'' patient cases that make little or no references to prior or lateral images. As illustrated in the model outputs, there’s little difference between \mairatwo and \medgemini phrases, and the original reference text. In independent reviews with two radiologists, minor variances were surfaced in terms of findings missed or hallucinated, and preferences for their conciseness or ordering that are described in the Figure captions. Overall, for this very limited set of examples, which predominantly report negative rather than more clinically relevant (positive) findings, it is difficult to surface any more clinically significant differences between the outputs of either model.  

\clearpage
\clearpage

\clearpage

\afterpage{
\thispagestyle{empty}
\begin{figure}
       \centerline{\includegraphics[scale=0.5]{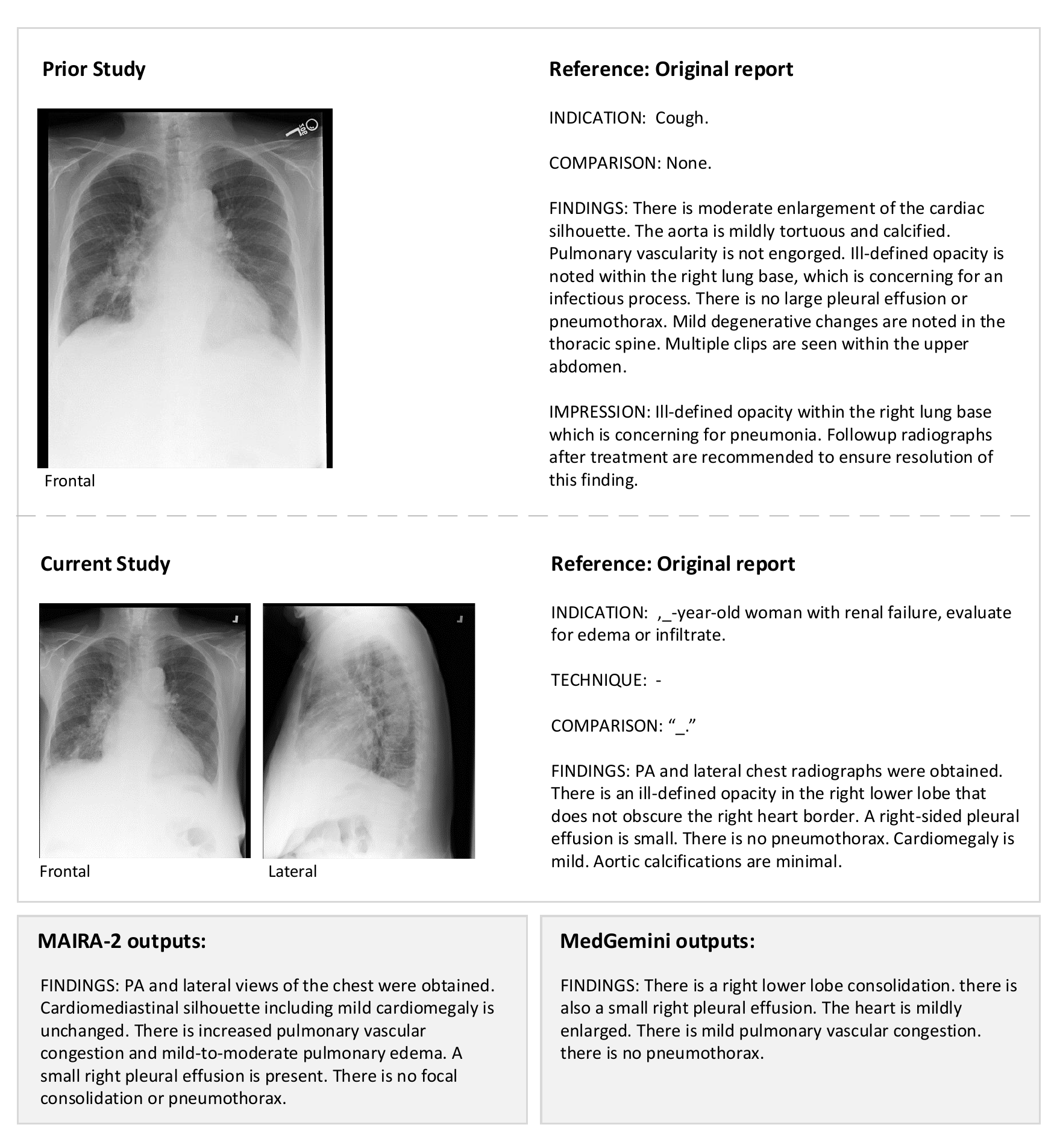}}
    \caption{One of the examples from \cite{yang2024advancing} to enable qualitative comparison to \medgemini. In this example study, like \Cref{fig:MedGemini_example_2}, \mairatwo outputs the \technique, which is in keeping with the reference report. The MAIRA-2 outputs also correctly describe the ``unchanged'' mild cardiomegaly. However, neither candidate findings say anything about the progression of the previously reported right lower lobe opacity, nor does the reference text explicitly describe any changes from the prior. Both generated report candidates state the existence of ``pulmonary vascular congestion'', which is most apparent via comparison with the prior study, and yet this finding was not included in the reference text. %
In the reference report, it is implied that the ill-defined opacity in the right lower lobe is a consolidation. Whilst the \medgemini findings indeed include ``There is a right lower lobe consolidation'', the \mairatwo outputs falsely state ``There is no focal consolidation''. %
Lastly, neither generated findings texts report the minimal ``aortic calcifications'', which – as a chronic finding – was reported previously, and thus, our radiologists did not consider this as a significant omission.
}
    \label{fig:MedGemini_example_3}
\end{figure}
\clearpage
}

\afterpage{
\thispagestyle{empty}
\begin{figure}
       \centerline{\includegraphics[scale=0.4]{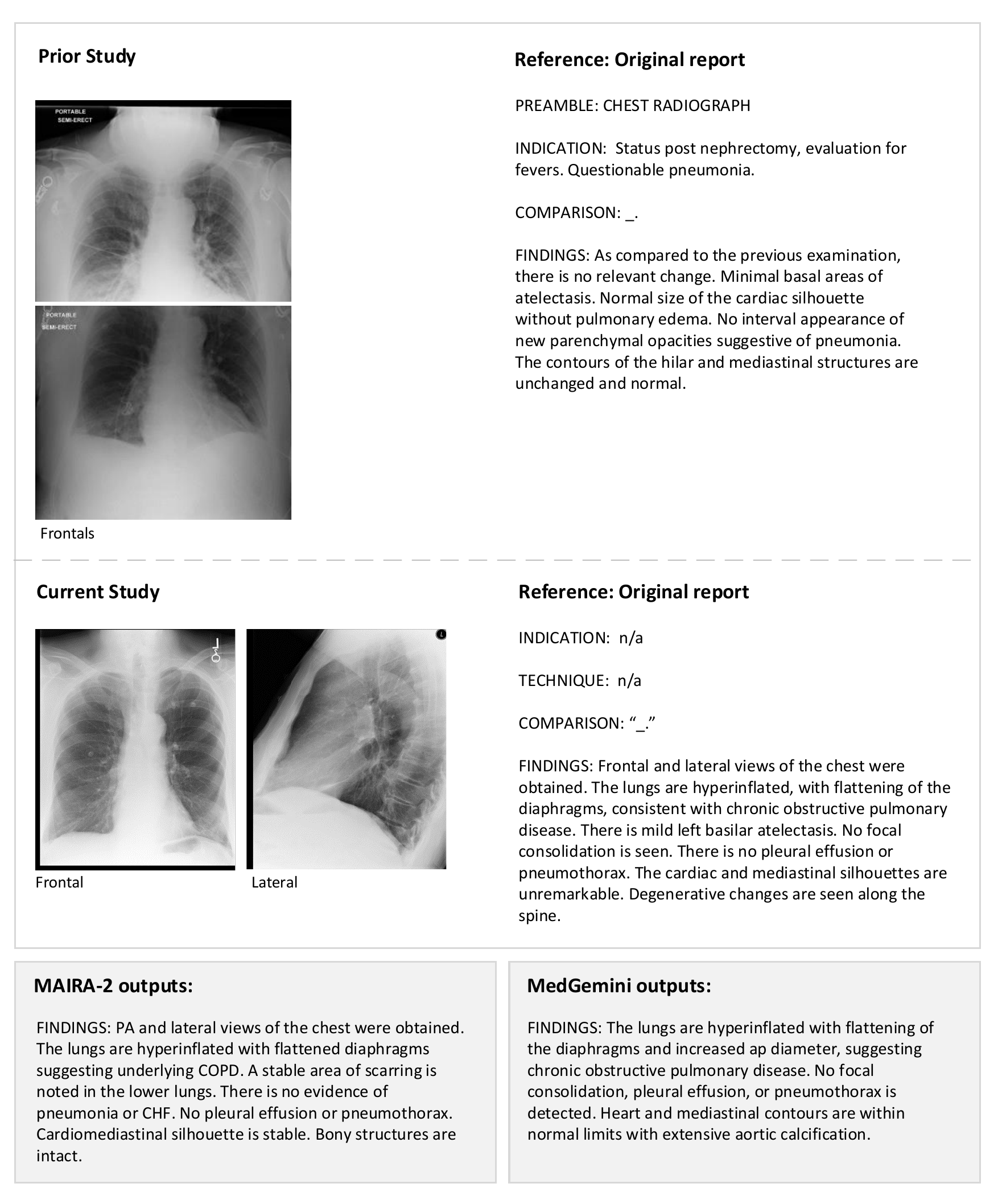}}
    \caption{One of the examples from \cite{yang2024advancing} to enable qualitative comparison to \medgemini. Again, \mairatwo outputs technical details of image views as part of the \findings as is reflective of the reference text. Both candidate reports include the suggestion of an underlying ``COPD'', which presents a clinical diagnosis rather than an image finding. %
    \medgemini outputs further state ``increased ap diameter''. Whilst this finding is not false, it likely presents a hallucination since the AP dimension can only be seen on the lateral view, which was not part of the \medgemini model training. The \mairatwo findings of ``stable area of scarring is noted in the lower lungs'' relates to the mild left basilar atelectasis in the reference text – a finding that was not reported by \medgemini. While reporting of the area of scarring and its progression from the prior (``stable'') are correct in the \mairatwo outputs, its location description is imprecise and should state in which lower lung (singular, left) it is present. %
The reference text further states ``Degenerative changes are seen along the spine''. \mairatwo outputs instead state that the ``Bony structures are intact''. There is no commentary made about the bones in the \medgemini output, which – similar to \mairatwo – may suggest an assumed normal. In general, degenerative changes to the spine, especially with the existence of prior studies, are not considered a new finding and are therefore less important to mention. Lastly, our radiologists could not see the ``extensive aortic calcification'' that was described in the \medgemini findings and that were also not remarked on by the reference text.\textit{  Please note, for \mairatwo, only the upper frontal image of the prior study was included into the analysis. }
}
    \label{fig:MedGemini_example_4}
\end{figure}
\clearpage
}

\clearpage
\section{Qualitative evaluation of twenty random MAIRA-2 generated reports}
\label{app:qual_review}
\subsection{Method}

\begin{figure}
    \centerline{
     \includegraphics[scale=0.8]{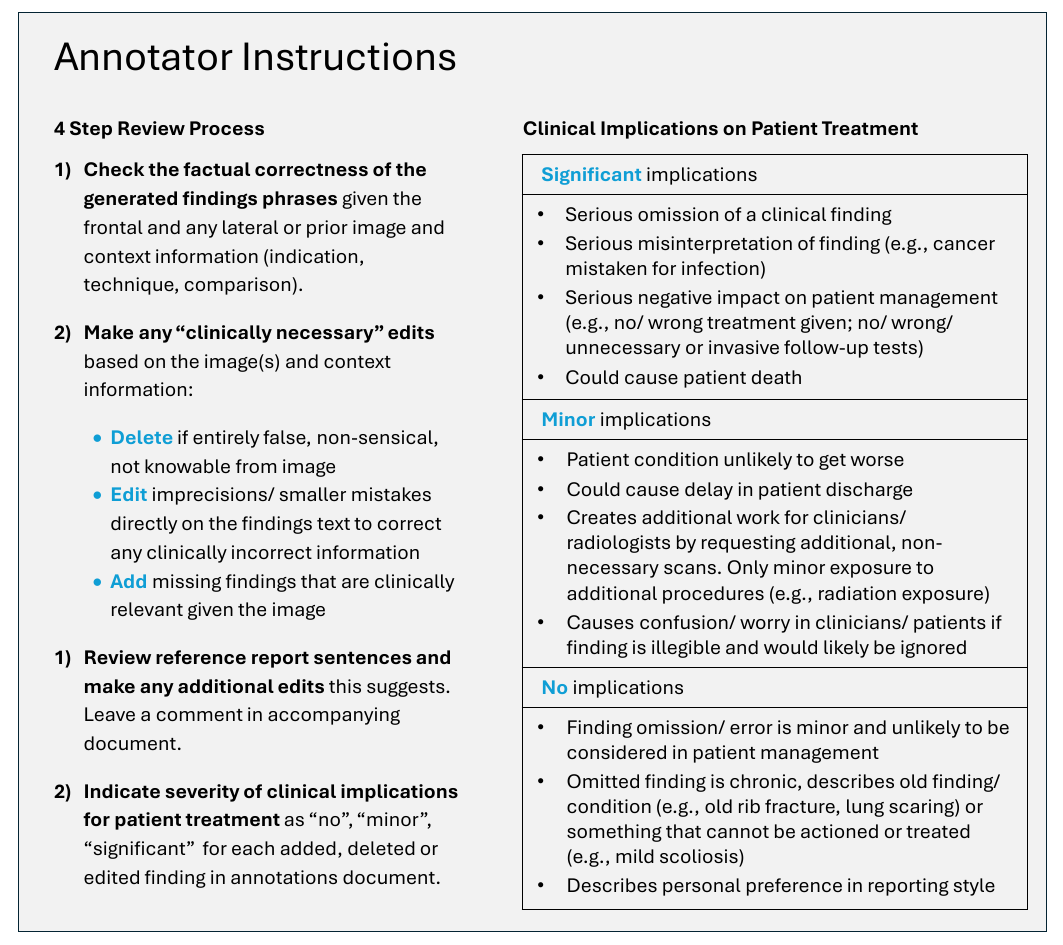}}
\caption{Outline of the 4-step review process the radiologists was asked to follow in their analysis. The radiologist was asked to: (1) check the factual correctness of the generated findings phrases; (2) make any “clinically necessary” edits given the model inputs; (3) review the reference findings as additional insight to the study; and (4) for each of the false, incomplete or omitted findings, they had to indicate if it would have “no”, “minor” or “significant” clinical implications on patient treatment (see definitions in Table to the right). We chose to include step 3, the review of the reference findings sentences, as additional context to the image interpretation – alike a peer reviewer perspective – since the web-UI did not provide the same image review resolution or functionality that is provided by DICOM viewers, nor was any broader patient context available.}
    \label{fig:Annotator_Instructions}
\end{figure}

We conducted a systematic, in-depth qualitative review of generated MAIRA-2 output for twenty randomly selected examples of the US Mix  dataset with a thoracic radiologist. To scaffold the process, we utilized a custom-built web-UI that illustrates each study with its corresponding model inputs and outputs as shown in \Cref{fig:Demo_UI}. The UI provides expert reviewers with functionality to “edit”, “delete” or “add” any findings in the generated output, intended to emulate a closer-to-real review scenario that captures (i) the extent and (ii) type of corrections a radiologists might make in practice. Following a 4-step review process that is outlined in \Cref{fig:Annotator_Instructions}, our radiologist annotations were further accompanied by a (iii) rating of the clinical implications of each false, incomplete or omitted finding on patient treatment as “no”, “minor” or “significant”; a definition is provided in \Cref{fig:Annotator_Instructions}. Exported as a csv file, resulting annotations were then analysed to derive both a comprehensive, human-expert based overview of MAIRA-2 report generation quality for a specified data subset; as well as more detailed insights into error cases and their potential implications. While the Demo UI has additional capability to show (and draw additional) grounding boxes, this evaluative exploration focused solely on assessing the generated report text.

\afterpage{
\thispagestyle{empty}
\begin{figure}
    \centerline{
     \includegraphics[scale=0.8]{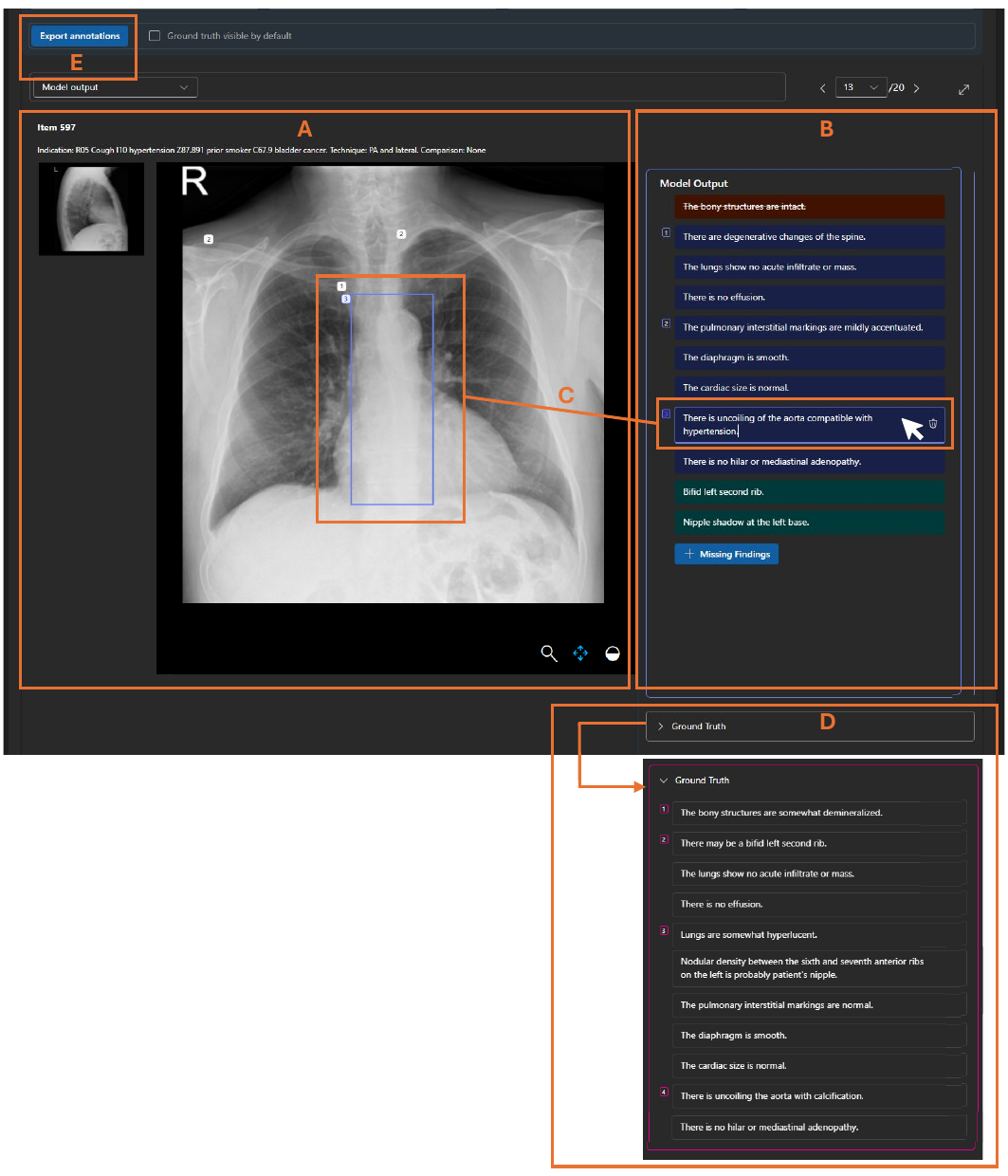}}
\caption{Illustration of the custom built web-UI that allows sequential review of multimodal image studies by expert annotators. Section A shows the model inputs including: the current frontal image; any lateral or prior images to the left side; and any text descriptions pertaining to the Indication, Technique or Comparison above the frontal image. Section B shows the MAIRA-2 generated phrase outputs.) Upon clicking on a finding with corresponding bounding box(es), this becomes interactively shown on the frontal image (C). Any finding phrases that were “deleted” are crossed-out (see first finding phrase); and those that “added” are show as green (see bottom two findings phrases); and where text “edits” are made become colour-graded as red (not shown here). A tap underneath the generated phrases unfolds the list of reference sentences and boxes for that study (D). Lastly, all annotations made can be exported as csv file for analysis (E).}
    \label{fig:Demo_UI}
\end{figure}
\clearpage
}

\clearpage

\subsection{Findings}

\textbf{Prevalence of corrections}: The 20 example reports included 5 normal studies\footnote{These are studies with no pathological findings being reported} and encompassed in total 135 generated phrases, equating to a Median of six phrases per report (\textit{min} = 2, \textit{max} = 13). The vast majority of these phrases (\textit{n} = 123) – and correspondingly 30\% of all reports (6 out of 20) – did not require any edits. The remaining reports received on average 1-2 corrections (\textit{Median} = 1, \textit{M} = 1.78, \textit{min} = 1, \textit{max} = 5). Amongst the total of 25 corrections, eleven presented “edits” to existing finding phrases, two findings were “deleted” and another 12 were “added”. 

\textbf{Error types \& insights: }Amongst the 25 corrections (see \Cref{tab:corrections_edits} to \Cref{tab:corrections_del} for the full list), “omissions” of findings  was the most prominent error type (\textit{n} = 15) evidenced through  either the addition of an entirely new finding phrase (\textit{n} = 12) or as an adjunct to an existing phrase (\textit{n} = 3). Apart from two significant findings that were missed (detailed below), added findings or text predominantly served: (i) to \textit{provide more detail or add a differential to a finding} (e.g., “There is a small left pleural effusion \underline{and/or thickening}” [underline added]); (ii) for \textit{completeness in reporting}, which pertained to mild, borderline or chronic findings (e.g., “Left-sided deviation of the trachea”); and (iii) to \textit{reduce ambiguity or need for clinical follow-up} by explicitly stating the absence of a pathology (e.g., “No pleural effusion”), or clarifying the appearance of an otherwise potentially misinterpreted anatomical structure (e.g., “Nipple shadows” that may resemble a lung nodule).  

Outside findings omissions and additional clarifications, we identified three instances of misclassification. Two of these cases stated about the image that “The bone structures are intact”, which however contradicted subsequently generated phrases describing “degenerative changes of the spine” and “demineralized” bones. To eliminate this contradiction, the false initial sentence was deleted by the radiologist. This foregrounds the need to assess the \textit{internal logic and consistency across generated phrases within a report}; and identifies it as an important avenue for future improvement to the \evmetric metric. 

Amongst the remaining edits, the generated report findings were twice identified as being \textit{overspecific} in their formulation (e.g., stating “There is evidence of cholecystectomy” when the presence of abdominal surgical clips is suggestive of abdominal surgery, but not necessarily a cholecystectomy). We further observed one instance of \textit{incorrect progression} information whereby the phrase stated “Bilateral infiltrates \underline{have improved}.” [underline added] although no comparison study was available; as well as two instances of \textit{incorrect location} information. One of these cases pertained to the generated description of a pacemaker: “Pacemaker\underline{s} are noted in the right \underline{atrium and} ventricle.” [underline added]  that was corrected to “Pacemaker with lead noted in the right ventricle”. Notably, not only is the pacemaker singular, it has also only 1 lead and consequently cannot end in two locations. The presence of only 1 lead means that the text output should be restricted to only include one location and suggest that the MAIRA-2 model has little embedded knowledge of the device characteristics. 

\textbf{Clinical implications:} Most crucially, two of the added findings were rated as having potentially “significant” clinical implications on patient care. These related to a missed pathology (an infiltrate in the lingula) that requires clinical action – for example treatment with antibiotics; as well as missed acute rib fractures in a patient case, which explained the “chest pain” mentioned in the study indication, therefore presenting the most relevant finding for explaining that patients’ symptoms. Apart from these two significant finding omissions, the majority of missed, misclassified or insufficiently described cases reflected either mild or more borderline cases – and were rated as potentially having “minor” (\textit{n} = 15) or “no” (\textit{n} = 8) implications on patient treatment. For instance, as “minor” were rated the misclassification of a heart as “normal” in size, when it was “mildly enlarged”; or the missing of “minor atelectasis” at a lung base. Instances with “no” implications pertained to text corrections or additions that served to improve precision and completion in text descriptions, to disambiguate non-pathological findings, or describe non-actionable, chronic conditions (e.g., mild scoliosis). 

Overall, the step-by-step review of all phrases of the twenty reports led the radiologist to conclude that the MARIA-2 outputs were “acceptable as a draft” alike a junior-to-mid level resident performance that requires however a more senior radiologist or consultant to double-check before sign-off. However, they also acknowledged that the example cases were not very complex in that on average, they included a Median of 2 pathologies per report (\textit{Mean} = 1.8, \textit{min} = 0, \textit{max} = 7) with no overlap or interaction between pathologies as would be more common, for example, in ICU or post-operative care settings.  

\subsection{Conclusions}
While  this qualitative investigation of twenty random MAIRA-2 output instances by one thoracic radiologists does not yield any generalizable results, it makes three important contributions:
\begin{enumerate}
    \item \textit{\textbf{It provides an instance of a more fine-grained, phrase-level human assessment \& extended understanding of reporting quality criteria.}} By inviting a domain expert to review and adapt individual phrase outputs, our approach to qualitative evaluation of report generation differs from most existing works that primarily ask experts to rate overall report quality via a (Likert-) scale~\citep{huang2023generative}; provide total error counts/ categorisation~\citep{yu2023evaluating, ostmeier2024green}; or engage in comparison rankings across different report candidates~\citep{boag2021pilot}. Whilst evaluating at a phrase-level also enables higher-level aggregates and quantifications, it has the added advantage to provide a record of concrete corrections. These can expand existing definitions of possible “error types” and their differentiation; and deepens understanding of what constitutes “good quality” in reporting outputs beyond common goals to reduce the occurrence of falsely classified, imprecise, or missed pathological findings. In keeping with established radiology reporting guidelines~\citep{langlotz2016radiology, hartung2020create}, our findings showed that – under consideration of the study context (e.g., baseline scan) – data annotations were made to also serve goals of: “completeness”; to “disambiguate potentially inconclusive, non-pathological findings”; and to improve “language clarity”– suggesting their inclusion as important evaluation criteria in future benchmarking efforts.
    \item \textbf{\textit{It exemplifies a comprehensive, non-ML domain expert legible overview of model performance with detailed examples to aid clinical utility-risk assessments.}} Specifically, our findings representation \Cref{fig:qual_review} conveys the likely “effort” required to correct errors for a specific data subset and different error types that occur that clarify “model limitations and needed improvements”. In this instance identified requirements included the need: to improve sensitivity to minor findings detection; to consider internal logic and consistency; and requirements to expand on device knowledge. Combined with indications of “potential clinical implications” that could result from utilizing draft reports in practice; these insights allow developers to determine risk-benefit and what an acceptable level of risk looks like. 
    \item \textbf{\textit{It suggests leveraging user-interface design to aid human-expert annotations\& their prototyping into more scalable evaluation or model adaptation approaches.}}  The involvement of human domain experts as (qualitative) annotators is widely considered as effortful and costly~\citep{boag2021pilot}; and as less flexible or scalable for iterative model testing. However, we believe that model development processes benefit from balancing fast, quantitative, more easily scalable approaches with more detailed, qualitative reviews involving ideally multiple experts at key stages of the process (as illustrated above). For this, having a user interface that enables ML teams to flexibly set-up annotation projects to assist effective data collection in a format that supports subsequent analysis has multiple additional advantages. For example, concrete lists of “added” or “deleted” or “edited” findings can be used to test out the performance of error classification techniques; serve as examples in GPT-based prompting to expand error analysis across to a larger dataset; or may serve as useful input to RLHF methods. These avenues warrant further exploration in future research. 
\end{enumerate}

\begin{table}[hbt!]
  \caption{List of the 11 edits made for the twenty example studies. For each phrase, it shows the report item number (as reference only), error type categorisation, and clinical implications rating (0 = no implications, 1 = minor implications, 2 = significant implications).}
  \label{tab:corrections_edits}
  \small 
  \tabspace
    \centerline{
    \begin{tabular}{@{}p{0.05\linewidth}p{0.3\linewidth}p{0.3\linewidth}p{0.2\linewidth}p{0.1\linewidth}@{}}
    \toprule
    \mr{\textbf{Item}} & \mr{\textbf{Generated Phrase}} & \mr{\textbf{Edited Phrase}} & \mr{\textbf{Error type}} & \textbf{Clinical}\\
    & & & & \textbf{implications}\\
    \midrule
    333 & The heart size is normal.  & The heart size is \textbf{mildly enlarged}. & MISCLASSIFICATION & 1 \\
    \midrule
    49 & There are degenerative changes of the spine.  & There are degenerative changes of the spine \textbf{with mid thoracic spine compression fractures}. & OMISSION & 1 \\
    \midrule
    155 & The lungs are well expanded.  & The lungs are well expanded \textbf{with upper zone emphysema noted}. & OMISSION & 1 \\
    \midrule
    411 & There is a small left pleural effusion.  & There is a small left pleural effusion \textbf{and/or thickening}. & OMISSION & 1 \\
    \midrule
    333 & No pleural effusion is identified \textbf{on the left}.  & No pleural effusion is identified. & OVERSPECIFIC & 1 \\
    \midrule
    415 & There is evidence of \textbf{cholecystectomy}.  & There is evidence of \textbf{prior abdominal surgery}. & OVERSPECIFIC & 1 \\
    \midrule
    781 & Bilateral infiltrates \textbf{have improved}.  & Bilateral infiltrates. & INCORRECT PROGRESSION & 1 \\
    \midrule
    168 & Focal infiltrate \textbf{left lower lobe}.  & Focal infiltrate \textbf{in the lingula}. & INCORRECT LOCATION & 1 \\
    \midrule
    49 & Pacemaker\textbf{s} are noted in the right\textbf{ atrium and} ventricle.  & Pacemaker with lead noted in the right ventricle. & INCORRECT LOCATION & 1 \\
    \midrule
    \mr{240} & \mr{\textbf{The lungs show }no effusion.} & \mr{No \textbf{pleural} effusion.} & OTHER: Anatomically & \mr{0} \\
    & & & imprecise description & \\
    \midrule
    \mr{467} & \mr{\textbf{The lungs show }no effusion.} & \mr{No \textbf{pleural} effusion.} & OTHER: Anatomically & \mr{0} \\
    & & & imprecise description & \\
    \bottomrule
    \end{tabular}
    }
\end{table}

\begin{table}[hbt!]
  \caption{List of the 12 phrase additions to the twenty example studies. For each phrase, it shows the report item number (as reference only), error type categorisation, and clinical implications rating (0 = no implications, 1 = minor implications, 2 = significant implications).}
  \label{tab:corrections_add}
  \small 
  \tabspace
    \centerline{
    \begin{tabular}{@{}p{0.075\linewidth}p{0.525\linewidth}p{0.25\linewidth}p{0.1\linewidth}@{}}
    \toprule
    \mr{\textbf{Item}} & \mr{\textbf{Added Phrase}} & \mr{\textbf{Error type}} & \textbf{Clinical}\\
    & & & \textbf{implications}\\
    \midrule
    155 & Infiltrate noted in the lingula. & OMISSION & 2 \\
    \midrule
    292 & Right 3rd and 4th posterior rib fractures. & OMISSION & 2 \\
    \midrule
    49 & Left sided deviation of the trachea. & OMISSION & 1 \\
    \midrule
    49 & Air-fluid level in the upper esophagus. & OMISSION & 1 \\
    \midrule
    155 & Mild scoliosis and degenerative changes of the spine. & OMISSION & 1 \\
    \midrule
    363 & Minor atelectasis at the left base. & OMISSION & 1 \\
    \midrule
    363 & Mild scoliosis. & OMISSION & 0 \\
    \midrule
    \mr{597} & \mr{Bifid left second rib.} & OMISSION: & \mr{0} \\
    & & Anatomical Observation & \\
    \midrule
    \mr{597} & \mr{Nipple shadow at the left base.} & OMISSION: & \mr{0} \\
    & & Anatomical Observation & \\
    \midrule
    \mr{921} & \mr{Nipple shadows noted.} & OMISSION: & \mr{0} \\
    & & Anatomical Observation & \\
    \midrule
    1033 & No pleural effusion. & OMISSION: Normal Finding & 0 \\
    \midrule
    921 & No pleural effusion or pneumothorax. & OMISSION: Normal Finding & 0 \\
    \bottomrule
    \end{tabular}
    }
\end{table}

\begin{table}[hbt!]
  \caption{List of the 2 deletions made across the twenty example studies. For each phrase, it shows the report item number (as reference only), error type categorisation, and clinical implications rating (0 = no implications, 1 = minor implications, 2 = significant implications).}
  \label{tab:corrections_del}
  \small 
  \tabspace
    \centerline{
    \begin{tabular}{@{}p{0.075\linewidth}p{0.525\linewidth}p{0.25\linewidth}p{0.1\linewidth}@{}}
    \toprule
    \mr{\textbf{Item}} & \mr{\textbf{Deleted Phrase}} & \mr{\textbf{Error type}} & \textbf{Clinical}\\
    & & & \textbf{implications}\\    
    \midrule
    \mr{49} & \mr{The bony structures are intact.} & MISCLASSIFICATION: & \mr{1} \\
    & & Inconsistency (internal logic) & \\
    \midrule
    \mr{597} & \mr{The bony structures are intact.} & MISCLASSIFICATION: & \mr{1} \\
    & & Inconsistency (internal logic) & \\
    \bottomrule
    \end{tabular}
    }
\end{table}

\clearpage
\nolinenumbers

\includepdf[
  addtotoc={1,section,1,Grounded reporting annotation protocol,app:annotation_protocol},
  pages=-,
]{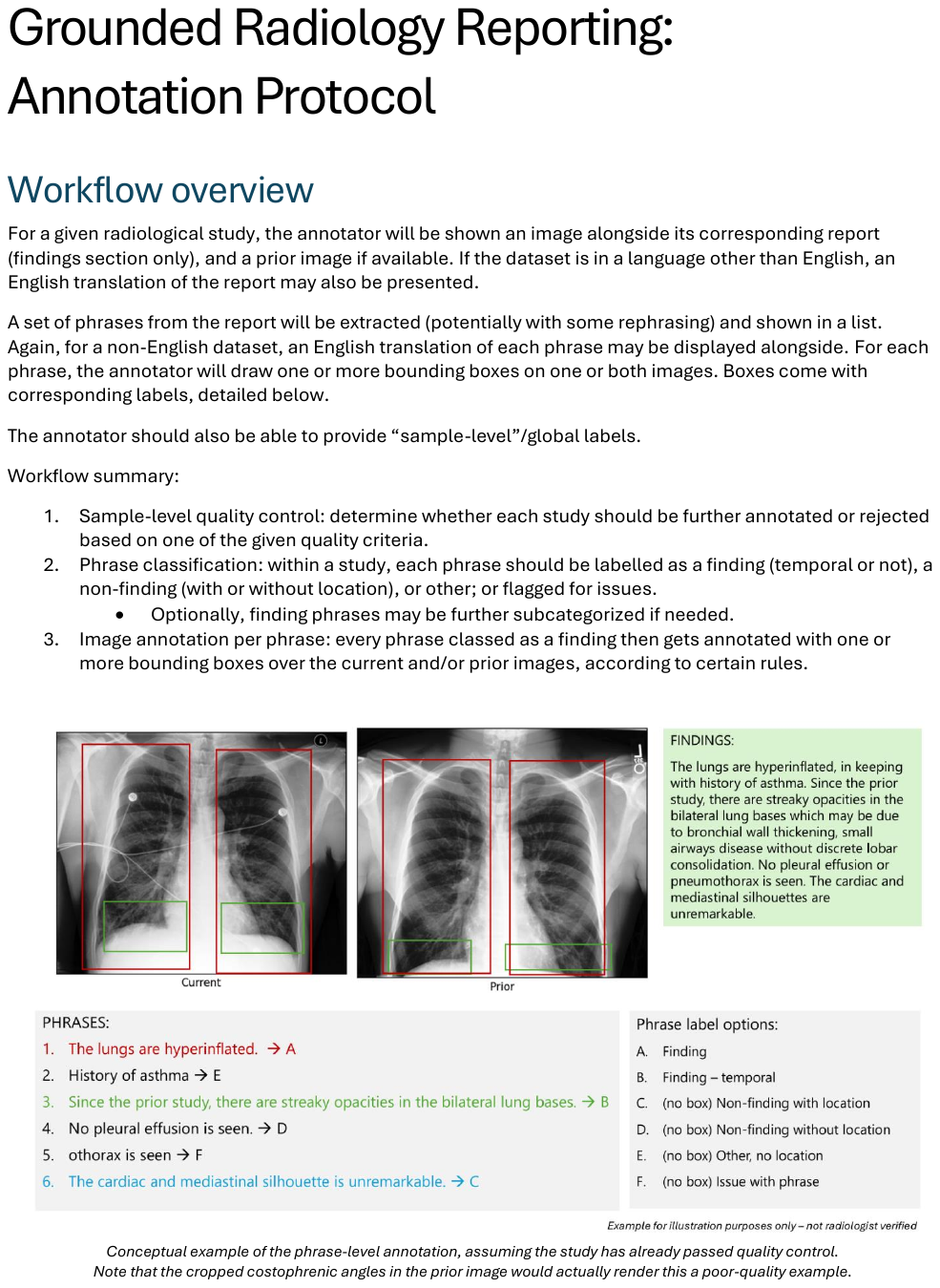}

\clearpage

\printbibliography[heading=bibintoc]
\end{refsection}
\end{document}